\documentclass[ROB,biber]{nowfnt} % creates the journal version, needs biber version
\usepackage{amsmath}
\usepackage{amssymb,amsfonts,amscd,amsthm}
\usepackage[printonlyused,withpage]{acronym}
\usepackage{listings}
\usepackage{subcaption}
\usepackage{xcolor}
\usepackage{hyperref}
\usepackage{enumitem}

 \graphicspath{{figs/}} 

\newcommand{\Real}{\mathbb{R}}
\newcommand{\diag}{\textup{diag}}
\newcommand{\bbm}{\begin{bmatrix}}
\newcommand{\ebm}{\end{bmatrix}}
\newcommand{\mbf}{\mathbf}
\newcommand{\mbs}[1]{{\boldsymbol{#1}}}
\def\ep{\epsilon}

\newcommand{\beq}{\begin{equation}}
\newcommand{\eeq}{\end{equation}}
\newcommand{\bdis}{\begin{displaymath}}
\newcommand{\edis}{\end{displaymath}}
\newcommand{\beqn}[1]{\begin{subequations}\label{eq:#1}\begin{eqnarray}}
\newcommand{\eeqn}{\end{eqnarray}\end{subequations}}
\newcommand{\est}[1]{\hat{#1}}
\newcommand{\pri}[1]{\check{#1}}
\newcommand{\wdg}{\wedge}

\newcommand{\Tsmall}{\mbf{T}}

\newcommand{\Jbig}{\mbs{\mathcal{J}}}
\lstdefinestyle{C++Style}{
  language=C++,
  basicstyle=\scriptsize\ttfamily, % Smaller monospaced font
  keywordstyle=\color{blue}, % Keywords in blue
  commentstyle=\color{green!50!black}, % Comments in dark green
  stringstyle=\color{red}, % Strings in red
  showstringspaces=false, % Don't show spaces in strings
  numbers=left, % Line numbers on the left
  numberstyle=\tiny\color{gray}, % Style for line numbers
  frame=single, % Add a frame around the listing
  breaklines=true, % Allow line breaking
  backgroundcolor=\color{black!5}, % Light gray background
  upquote=true,  % Don't convert ' into curly smart quotes
}

\newcommand{\eg}{\emph{e.g.}}

\title{Smoothing Out the Edges: Continuous-Time Estimation with Gaussian Process Motion Priors on Factor Graphs}

\subtitle{}

\maintitleauthorlist{
{\sffamily\bfseries Connor Holmes$^*$} \\
{\sffamily\bfseries Sven Lilge$^*$} \\
{\sffamily\bfseries Zi Cong Guo} \\
University of Toronto \\
\and
Frank Dellaert \\
Georgia Institute of Technology \\
\and
Timothy D. Barfoot \\
University of Toronto
}

\issuesetup
{%
 copyrightowner={C.~Holmes, Z.~C.~Guo, S.~Lilge, F.~Dellaert, and T.~D.~Barfoot},
 volume        = xx,
 issue         = xx,
 pubyear       = 2025,
 isbn          = xxx-x-xxxxx-xxx-x,
 eisbn         = xxx-x-xxxxx-xxx-x,
 doi           = 10.1561/XXXXXXXXX,
 firstpage     = 1, %Explain
 lastpage      = 18 
 }

\usepackage{mwe}

\author[1*]{Holmes, Connor}
\author[1*]{Lilge, Sven}
\author[1]{Guo, Zi Cong}
\author[2]{Dellaert, Frank}
\author[1]{Barfoot, Timothy D.}

\affil[1]{University of Toronto; {connor.holmes, zc.guo}@mail.utoronto.ca, {sven.lilge, tim.barfoot}@utoronto.ca}
\affil[2]{Georgia Institute of Technology; frank.dellaert@cc.gatech.edu}
\affil[*]{Denotes equal contribution}

\articledatabox{\nowfntstandardcitation}

\acrodef{BA}{Bundle Adjustment}
\acrodef{SLAM}{Simultaneous Localization and Mapping}
\acrodef{STEAM}{simultaneous trajectory estimation and mapping}
\acrodef{SDP}{Semidefinite Program}
\acrodef{IRLS}{Iteratively Reweighted Least-Squares}
\acrodef{PSD}{positive-semidefinite}
\acrodef{GP}{Gaussian Process}
\acrodef{WNOV}{White Noise on Velocity}
\acrodef{WNOA}{White Noise on Acceleration}
\acrodef{WNOJ}{White Noise on Jerk}
\acrodef{RTS}{Rauch-Tung-Striebel}
\acrodef{KF}{Kalman filter}
\acrodef{SDE}{stochastic differential equation}
\acrodef{GN}{Gauss-Newton}
\acrodef{MAP}{maximum a posteriori}
\acrodef{BCH}{Baker-Campbell-Hausdorff}
\acrodef{LTV}{linear time-varying}
\acrodef{LTI}{linear time-invariant}
\acrodef{GTSAM}{Georgia Tech Smoothing and Mapping}
\acrodef{lidar}[LiDAR]{light detection and ranging}
\acrodef{radar}[RADAR]{radio detection and ranging}
\acrodef{imu}[IMU]{inertial measurement unit}
\acrodef{API}[API]{Application Programming Interface}

\begin{document}

\makeabstracttitle

\begin{abstract}
Continuous-time state estimation is gaining in popularity due to its abilities to provide smooth solutions, handle asynchronous sensors, and interpolate between data points.  While there are two main paradigms, {\em parametric} (e.g., temporal basis functions, splines) and {\em nonparametric} (Gaussian processes), the latter has seen less adoption despite its technical advantages and relative ease of implementation.  In this article, we seek to rectify this situation by providing a new simplified explanation of GP continuous-time estimation rooted in the language of {\em factor graphs}, which have become the de facto estimation paradigm in much of robotics.  To simplify onboarding, we also provide three working examples implemented in the popular GTSAM estimation framework. 

\end{abstract}

%!TEX root =  FnTarticle.tex

\chapter{Introduction}

In this article, we consider the problem of estimating a continuous-time trajectory from discrete measurements.  This is a common problem in robotics, computer vision, and many other fields where we have sensors that provide measurements at discrete time intervals, but we would like to query the state of a system at any point in time.  

Specifically, we will discuss using continuous-time estimation based on \ac{GP} regression.  This topic is not itself new and therefore this article should be considered mostly of tutorial value.  Its contributions beyond the existing literature are:
\begin{itemize}
\item {\em We provide new insights on the connections between continuous-time \ac{GP} regression and factor graphs.}  In particular, we show that querying the trajectory at any time of interest can be accomplished by an application of the factor-graph elimination algorithm.  This leads to a novel way to do covariance interpolation and new insights into interpolating measurement states.

\item {\em We implement \ac{GP}-based continuous-time estimation in the popular \ac{GTSAM} framework and apply it on three tutorial examples.}  This allows us to demonstrate the use of continuous-time estimation in a practical setting, including how to handle high-rate and asynchronous measurements.

\item {\em We release the code and the datasets used in the examples open source.}  This will hopefully make adoption by others easier and allow for further experimentation and development in this area.
\end{itemize}
In the rest of this chapter, we motivate the continuous-time estimation problem, discuss the specific type of continuous-time estimation we consider, and provide an overview of related work.

\section{What is Continuous-Time Estimation?}

\begin{figure}[t]
     \includegraphics[width=\textwidth]{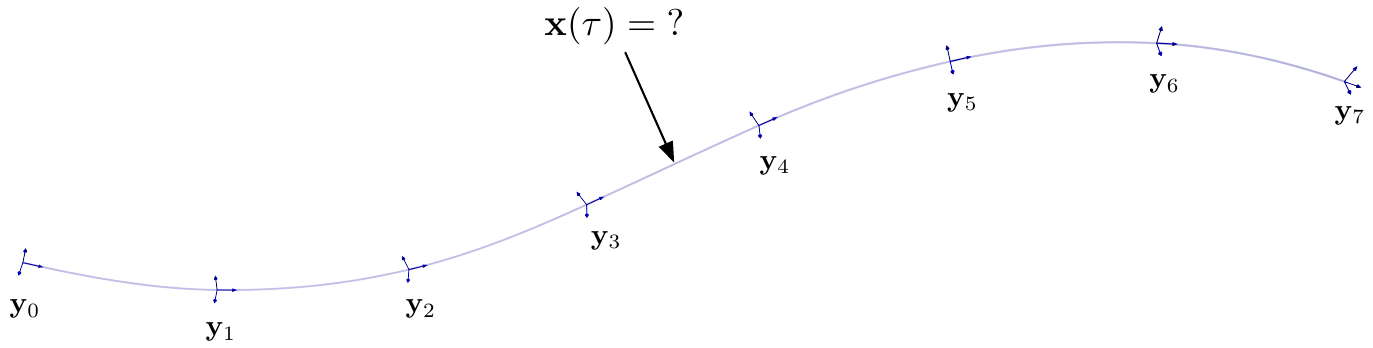}
     \centering
     \caption{We consider a trajectory to be a continuous function of time.  Measurements $\mbf{y}_k$ occur (possibly asynchronously) at discrete times $t_k$ yet we would like to be able to query (estimate) the trajectory state $\mbf{x}(\tau)$ at any time of interest, $\tau$.}
     \label{fig:ct}
\end{figure}

Figure~\ref{fig:ct} illustrates the general problem in which we are interested.  We have a trajectory $\mbf{x}(t)$ that we would like to estimate, but we only have measurements $\mbf{y}_k$ at discrete times $t_k$.  We would like to be able to query the trajectory at any time of interest, $\tau$. 

There are four main advantages to continuous-time estimation over its discrete-time counterpart:
\begin{itemize}
\item {\em We can easily handle high-rate and asynchronous measurements.}  This is accomplished by assigning each measurement to the trajectory at its time of occurrence, $t_k$.  

\item {\em We can query the trajectory at any time of interest, $\tau$, not just at the measurement times, $t_k$.}  This is useful for applications where we might use one sensor to estimate the trajectory but another to build a map, for example.  It is also useful in control where we might want to query the trajectory at a high rate to compute control inputs. 

\item {\em The continuous-time representation can impose smoothness on the estimated trajectory.}  Often the object to which our sensors are attached is moving smoothly through the world because it is governed by physics, and we can use this knowledge to regularize the solution through the choice of motion prior.

\item {\em The continuous-time representation can be used to overcome observability issues.}  For sweeping-while-moving sensors such as rolling-shutter cameras or spinning lidars, the trajectory is often under-constrained by the measurements alone since each pixel or point is acquired from a unique pose.  Our motion prior serves to regularize the solution and can be physically motivated, such as a constant-velocity prior or a constant-acceleration prior.
\end{itemize}

The ability to separate the roles of {\em measurement times}, {\em estimation times}, and {\em query times} is an important property of continuous-time estimation.  We will next discuss the two main paradigms for continuous-time estimation, parametric and nonparametric, and why we focus on the latter in this article.

\section{Parametric vs. Nonparametric}

A common way to perform continuous-time estimation is to use a {\em parametric} representation of the trajectory, such as a spline or temporal basis functions.  In this case, we assume that the trajectory can be represented by a finite number of parameters, and we estimate these parameters from the measurements.  For example, we might use a representation such as 
\begin{equation}\label{eq:spline}
\mbf{x}(t) = \sum_{n=0}^N \mbf{B}_n(t) \mbs{\theta}_n, 
\end{equation}
where $\mbs{\theta}_n$ are the coefficient parameters to be estimated and $\mbf{B}_n(t)$ are known temporal basis functions.  We receive some discrete-time measurements $\mbf{y}_k$ at times $t_k$, given by the linear measurement function,
\begin{equation}
\mbf{y}_k = \mbf{C}_k \mbf{x}(t_k) + \mbf{n}_k,
\end{equation}
with $\mbf{n}_k \sim \mathcal{N}(\mbf{0}, \mbf{R}_k)$ measurement noise.  We can then estimate the parameters $\mbs{\theta}_n$ by minimizing the residuals between the measurements and the predicted values from the trajectory,
\begin{multline}
\min_{\mbs{\theta}_n} \left( \sum_{k=0}^K \|\mbf{y}_k - \mbf{C}_k\mbf{x}(t_k)\|^2_{\mbf{R}_k} + \sum_{n=0}^N \|\mbs{\theta}_n\|^2 \right) \\ = \min_{\mbs{\theta}_n} \left( \sum_{k=0}^K \|\mbf{y}_k - \mbf{C}_k \sum_{n=0}^N \mbf{B}_n(t) \mbs{\theta}_n \|^2_{\mbf{R}_k} + \sum_{n=0}^N \|\mbs{\theta}_n\|^2 \right),
\end{multline}
where we have substituted in the trajectory from~\eqref{eq:spline}.  
The second term $\|\mbs{\theta}_n\|^2$ regularizes the problem by penalizing the magnitude of the parameters, $\mbs{\theta}_n$ (i.e., minimum description length prior).
Stacking all the quantities up into vectors/matrices, we can write this problem cleanly as
\begin{equation}\label{eq:least-sqr-intro}
\min_{\mbs{\theta}} \left( \|\mbf{y} - \mbf{C}\mbf{B}\mbs{\theta}\|^2_{\mbf{R}} + \|\mbs{\theta}\|^2 \right),
\end{equation}
The solution to this least-squares problem is the solution to the following linear system of equations:
\begin{equation}
\left( \mbf{B}^T \mbf{C}^T \mbf{R}^{-1} \mbf{C} \mbf{B} + \mbf{I} \right) \mbs{\theta} = \mbf{B}^T\mbf{C}^T \mbf{R}^{-1} \mbf{y}.
\end{equation}
One challenge with parametric methods is to choose the right basis functions $\mbf{B}_n(t)$, which can be difficult in practice; too few and the trajectory does not have enough flexibility to fit the data, too many and the solution becomes overfit, although the regularization term can help with this.  

The state at the measurement times is $\mbf{x} = \mbf{B} \mbs{\theta}$ and so we can manipulate the linear system of equations above to become
\begin{equation}\label{eq:nonparametric_system}
\left( \mbf{C}^T \mbf{R}^{-1} \mbf{C} + \mbf{K}^{-1} \right) \mbf{x} = \mbf{C}^T \mbf{R}^{-1} \mbf{y},
\end{equation}
where $\mbf{K} = \mbf{B}\mbf{B}^T$ can now be interpreted as a prior or regularizer term on the state $\mbf{x}$.\footnote{Note that the $\mbf{K}^{-1}\mbf{x} $ term in~\eqref{eq:nonparametric_system} corresponds exactly to the regularization term on $\mbs{\theta}$ in~\eqref{eq:least-sqr-intro}.}

The leap from parametric to {\em nonparametric} comes from the observation that we do not actually need the individual basis functions, only their inner products in order to construct $\mbf{K}$. As such, we can instead select a {\em kernel function} $\mbf{K}(t_k, t_\ell)$ that defines the inner product between two points in time, and we can write the {\em kernel matrix} as
\begin{equation}
\mbf{K} = \bbm \mbf{K}(t_k, t_\ell)\ebm,
\end{equation}
which is block-wise populated with evaluations of the kernel function at all pairs of measurement times.  This substitution is known as the {\em kernel trick}.  

There are a number of technical advantages of the nonparametric approach over the parametric one.  Selecting a kernel function rather than the individual basis functions provides more representational power and is inherently regularized without having to introduce and tune a minimum-description-length prior.  The kernel function essentially builds in smoothness to the system and we can therefore also use it to determine what is going on between the measurements; this is the essence of \ac{GP} regression \citep{rasmussen06}.  In general, the inverse kernel matrix $\mbf{K}^{-1}$ may be dense.  However, with the right choice of kernel function, we will show that it can be {\em block-tridiagonal}, which is a key advantage of our specific \ac{GP} approach as it means large computational savings.  Moreover, the kernel function can be chosen intuitively to represent a physically motivated motion prior, such as a constant-velocity or constant-acceleration prior, which is useful for many applications in robotics and computer vision.  Finally, there is also a well-known connection between the sparsity pattern of the left-hand side of~\eqref{eq:nonparametric_system} and the structure of the equivalent factor graph that we will discuss in more detail in Chapter~\ref{chap:fg-ctls}.  

\section{Related Work}

Much has been written about continuous-time estimation.  For an overview, we refer the reader to \citet{talbot_tro25}, which is a recent comprehensive survey of both parametric and nonparametric methods.  We also point to \citet{johnson_arxiv24} for a recent detailed comparison between parametric and nonparametric methods.  

In this article, we focus on the nonparametric approach, which has been less widely adopted in practice despite its technical advantages.  Here we provide a non-exhaustive timeline of some of the key publications that have specifically used \ac{GP} regression for continuous-time estimation up to this point:

\begin{itemize}
\item \citet{rasmussen06} - Introduced the use of \ac{GP} regression for continuous-time estimation in the context of dynamic systems.
\item \citet{sarkka06} - Showed the equivalence between discrete-time estimation and continuous-time estimation at the measurement times in the context of filtering and smoothing (i.e., chain-like graph structures).
\item \citet{tong_crv12}, \citet{tong_ijrr13b} - Made the initial connection between \ac{GP} regression and continuous-time estimation for general batch estimation problems (i.e., arbitrarily connected graph structures such as \ac{SLAM}).
\item \citet{barfoot_rss14}, \citet{anderson_ar15} - Showed how to construct a kernel function from a physically motivated motion prior that resulted in a block-tridiagonal inverse kernel matrix, which is key to efficient computation; made the initial connection between the \ac{GP} approach and sparse factor graphs.  Coined the term \ac{STEAM}.
\item \citet{anderson_iros15}, \citet{anderson_phd16} - Showed how to apply the \ac{GP} continuous-time approach (specifically a \ac{WNOA} motion model) on Lie groups using the `local GP' approach.
\item \citet{yanIncrementalSparseGP2015}, \citet{yanIncrementalSparseGP2017} - Showed that once the \ac{GP} approach was formulated as a factor graph it could be incrementalized using the Bayes-tree ideas of \citet{kaess08, kaessISAM2IncrementalSmoothing2012}.
\item \citet{dongSparseGaussianProcesses2018} - Generalized the \ac{GP} formulation to different Lie groups and provided examples.
\item \citet{mukadamContinuoustimeGaussianProcess2018} - Extended the approach beyond estimation to include motion-planning problems.
\item \citet{tang_ral19}, \citet{nguyen_tro25} - Showed how to construct a \ac{WNOJ} prior for Lie groups that worked well for vehicles experiencing acceleration.
\item \citet{wong_ral20} - Showed how to construct a Singer prior for Lie groups that was trained on data to better fit a real-world motion profile.
\item \citet{legentilGaussianProcessPreintegration2020}, \citet{legentilContinuousLatentState2023} - Showed how to use \acp{GP} to preintegrate inertial measurements in continuous time; this work is not directly connected to the \ac{GP} approach discussed in this article but is an important related topic.
\item \citet{lilge_ijrr22}, \citet{lilge_tro25} - Showed that the \ac{GP} approach could be used to also estimate the shape of continuum robots.
\item \citet{duembgen_ral23}, \citet{barfoot_ijrr25} - Showed how to incorporate \ac{GP} motion priors within a certifiably optimal estimation framework.
\item \citet{johnson_arxiv24} - Provided a detailed experimental comparison of parametric and nonparametric methods for continuous-time estimation.  Also showed how to incorporate motion priors into parametric methods to make them essentially equivalent to nonparametric methods.
\item \citet{barfoot_ser24} - Provided a new way to carry out covariance interpolation during querying that was simpler than \citet{anderson_phd16}.  An even better method is proposed in the current article.
\item \citet{lilge_rob25} - Extended the Lie-group \ac{GP} motion priors to include non-zero mean functions, allowing more specific prior motion to be captured.
\item \citet{burnett_tro25} - Showed state-of-the-art performance on radar- and lidar-inertial estimation using the \ac{GP} approach.
\end{itemize}
Additional related works and applications of the \ac{GP} approach are listed by \citet{talbot_tro25} in Table VI.

\section{How to Read This Article}

The rest of the article is structured in the following way:

\begin{itemize}
\item Chapter 2: We review the details of \ac{GP} regression for continuous-time linear state estimation, highlighting key mathematical derivations and practical results. Though this chapter sets the stage for the next chapter, it can be safely skipped by readers that already have a good understanding of this topic.
\item Chapter 3: We present a novel reframing of \ac{GP} continuous-time estimation from the perspective of the {\em factor graph} paradigm. This framing translates several key concepts from continuous-time estimation into the {\em lingua franca} of state-estimation and represents our core \emph{theoretical} contribution. We also motivate continuous-time estimation with practical strengths throughout the chapter.
\item Chapter 4:  We extend \ac{GP} state estimation to the case of factor graphs on Lie groups. Here, we assume that the reader is already familiar with Lie theory for robotics and refer unfamiliar readers to the expositions in~\citet{solaMicroLieTheory2021} and~\citet{barfoot_ser24}. A novel covariance interpolation method is introduced that is simpler than the one in \citet{anderson_phd16} and more efficient than the one in \citet{barfoot_ser24}.
\item Chapter 5:  We elaborate on our main \emph{technical} contribution: the integration of \ac{GP}-based continuous-time techniques into \ac{GTSAM}, a popular library for robotic state estimation. We provide detailed examples of application of these techniques on three state-estimation datasets of our own creation and demonstrate the practical advantages of the continuous-time approach.
\item Chapter 6:  We wrap up and provide an outlook for the future.
\end{itemize}
We hope that this article provides a useful introduction to continuous-time estimation using \ac{GP} regression and that the examples provided will help practitioners to adopt this approach in their own work. 
%!TEX root =  main.tex

\chapter{Gaussian-Process Regression}\label{sec:gp-regression}

An inherent limitation of classical discrete-time optimization problems in robotics is that each state of interest must be  represented \emph{explicitly} as a variable. This can lead to complications when downstream robotics tasks, such as motion planning or control, require trajectory and uncertainty information at \emph{arbitrary} times that cannot be known prior to solving the estimation problem~\citep{mukadamContinuoustimeGaussianProcess2018}. Similar failings occur when performing estimation of \emph{smooth shapes} (\eg, for continuum robotics~\citep{lilge_tro25}). These issues can be readily handled by {\em continuous-time} estimation, which processes measurements at given times to estimate a trajectory that can be queried at \emph{any} time by a downstream process. 

In this chapter, we show how \ac{GP} regression -- a popular method for function approximation that fits with the Bayesian estimation paradigm \citep{rasmussen06} -- can be used to carry out {\em continuous-time} estimation. Here, we primarily follow \citet{barfoot_ser24} to provide a background on \ac{GP} regression and set the stage for the factor graph interpretation that will be the subject of the next chapter. Readers who are already familiar with the details of \ac{GP} regression can safely skip this chapter.

\section{Brief Derivation of GP Regression}

\begin{figure}[h]
    \centering
     \includegraphics[width=0.8\textwidth]{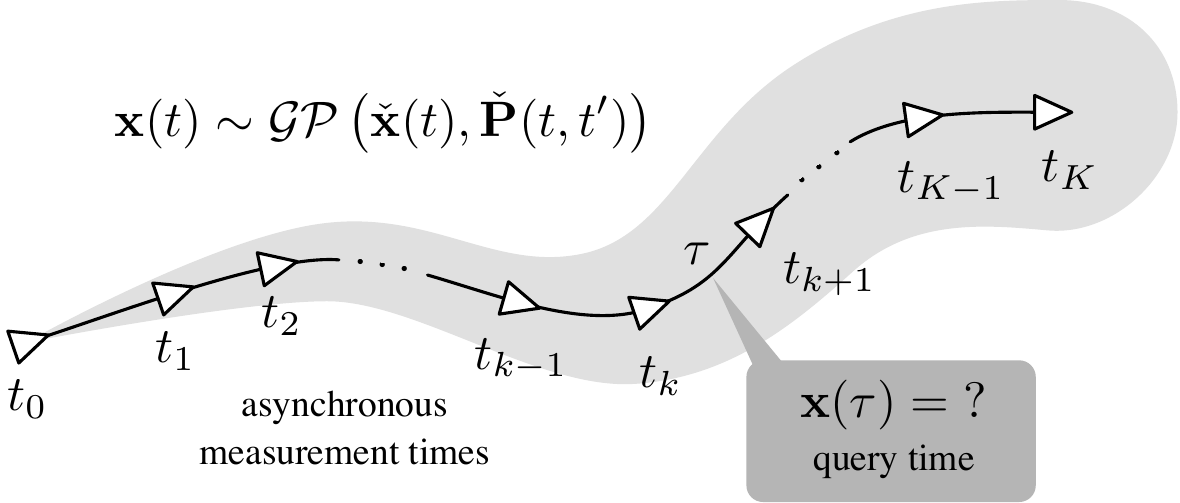}
     \caption{In \ac{GP} regression, we begin with some prior over trajectories, including mean and covariance functions.  After incorporating measurements, we obtain a posterior distribution over trajectories that can be queried at any time.}
     \label{fig:gpprior}
\end{figure}

We begin by considering systems with a continuous-time \ac{GP} prior and a discrete-time, linear measurement model:
\beqn{}
\mbf{x}(t) & \sim & \mathcal{GP}( \pri{\mbf{x}}(t), \pri{\mbf{P}}(t,t^\prime) ), \qquad t_0 < t,t^\prime, \\
\mbf{y}_k & = & \mbf{C}_k\mbf{x}(t_k) + \mbf{n}_k, \qquad t_0 < t_1 < \cdots < t_K,
\eeqn 
where $\mbf{x}(t)$ is the state, $\pri{\mbf{x}}(t)$ is the mean function, $\pri{\mbf{P}}(t,t^\prime)$ is the covariance function, $\mbf{y}_k$ are measurements, $\mbf{n}_k \sim \mathcal{N}(\mbf{0}, \mbf{R}_k )$ is Gaussian measurement noise, and $\mbf{C}_k$ is the measurement model coefficient matrix. The state at $t=0$ is defined by a prior, $\mbf{x}(0) \sim \mathcal{N}( \pri{\mbf{x}}_0, \pri{\mbf{P}}_0 )$. The first equation defines a \emph{Gaussian process} that captures our prior belief about the likelihood of all possible trajectories that the state could take. Accordingly, $\pri{\mbf{P}}(t,t^\prime)$ is a {\em kernel function} that describes the covariance between the state at two different times, $t$ and $t^\prime$. Figure~\ref{fig:gpprior} illustrates this conceptually.

We consider that we want to query the state at a number of times ($\tau_0 < \tau_1< \ldots < \tau_J$) that may or may not be different from the measurement times ($t_0 < t_1 < \ldots < t_K$). 
The joint density between the state (at the query times) and the measurements (at the measurement times) is written as
\begin{equation}\label{eq:gaussjoint}
p\left( \bbm   \mbf{x}_\tau \\ \mbf{y} \ebm \right) = \mathcal{N} \left( \bbm   \pri{\mbf{x}}_\tau  \\ \mbf{C} \pri{\mbf{x}} \ebm, \bbm  \pri{\mbf{P}}_{\tau\tau} & \pri{\mbf{P}}_\tau \mbf{C}^T \\ \mbf{C} \pri{\mbf{P}}_\tau^T &   \mbf{R} + \mbf{C} \pri{\mbf{P}}\mbf{C}^T \ebm  \right),
\end{equation}
where
\begin{gather*}
\mbf{x} = \bbm \mbf{x}(t_0) \\ \vdots \\ \mbf{x}(t_K) \ebm,  \quad \pri{\mbf{x}} = \bbm \pri{\mbf{x}}(t_0)  \\ \vdots \\ \pri{\mbf{x}}(t_K) \ebm, \quad \mbf{x}_\tau = \bbm \mbf{x}(\tau_0) \\ \vdots \\ \mbf{x}(\tau_J) \ebm,  \quad \pri{\mbf{x}}_\tau = \bbm \pri{\mbf{x}}(\tau_0) \\ \vdots \\ \pri{\mbf{x}}(\tau_J) \ebm, \\
\mbf{y} = \bbm \mbf{y}_0 \\ \vdots \\ \mbf{y}_K \ebm,  \quad \mbf{C} = \mbox{diag}\left( \mbf{C}_0, \ldots, \mbf{C}_K \right), \quad \mbf{R} = \mbox{diag}\left( \mbf{R}_0, \ldots, \mbf{R}_K \right), \\
\pri{\mbf{P}} = \bbm \pri{\mbf{P}}(t_i, t_j) \ebm_{ij}, \quad \pri{\mbf{P}}_\tau = \bbm \pri{\mbf{P}}(\tau_i, t_j) \ebm_{ij}, \quad \pri{\mbf{P}}_{\tau\tau} = \bbm \pri{\mbf{P}}(\tau_i, \tau_j) \ebm_{ij}.
\end{gather*}
In \ac{GP} regression, the matrix, $\pri{\mbf{P}}$, is known as the \index{kernel matrix} {\em kernel matrix} and comprises evaluations of the kernel covariance function between all pairs of measurement times.  Factoring~\eqref{eq:gaussjoint} via the Schur complement,\footnote{This factorization is common in Gaussian state-estimation and is provided in more general form in \eqref{eq:gausselim} of Appendix~\ref{app:add-fg-bg}.} we can show that
\begin{multline}
\label{eq:gpquerytimes}
p(\mbf{x}_\tau | \mbf{y}) = \mathcal{N}\biggl( \underbrace{\pri{\mbf{x}}_\tau + \pri{\mbf{P}}_{\tau} \mbf{C}^T( \mbf{C} \pri{\mbf{P}} \mbf{C}^T + \mbf{R})^{-1} (\mbf{y} - \mbf{C}\pri{\mbf{x}} )}_{\est{\mbf{x}}_{\tau}, \;\mbox{\small mean}}, \biggr. \\ \biggl. 
\underbrace{\pri{\mbf{P}}_{\tau\tau} - \pri{\mbf{P}}_\tau \mbf{C}^T ( \mbf{C} \pri{\mbf{P}} \mbf{C}^T + \mbf{R}  )^{-1} \mbf{C} \pri{\mbf{P}}_\tau^T }_{\est{\mbf{P}}_{\tau\tau}, \;\mbox{\small covariance}}\biggr),
\end{multline}
for the density of the predicted state at the query times, given the measurements.  

The expression simplifies further if we take the query times to be exactly the same as the measurement times (i.e., $\tau_k = t_k, K=J$).  This implies that 
\begin{equation}
\pri{\mbf{P}} = \pri{\mbf{P}}_\tau = \pri{\mbf{P}}_{\tau\tau},
\end{equation}
and then we can write
\begin{multline}
\label{eq:gpmeastimes2}
p(\mbf{x} | \mbf{y}) = \mathcal{N}\biggl( \underbrace{\pri{\mbf{x}} + \pri{\mbf{P}} \mbf{C}^T( \mbf{C} \pri{\mbf{P}} \mbf{C}^T + \mbf{R})^{-1} (\mbf{y} - \mbf{C}\pri{\mbf{x}} )}_{\est{\mbf{x}}, \;\mbox{\small mean}}, \biggr. \\ \biggl.
\underbrace{\pri{\mbf{P}} - \pri{\mbf{P}} \mbf{C}^T ( \mbf{C} \pri{\mbf{P}} \mbf{C}^T + \mbf{R}  )^{-1} \mbf{C} \pri{\mbf{P}}^T }_{\est{\mbf{P}}, \;\mbox{\small covariance}}\biggr).
\end{multline}
We can rearrange this using the {\em matrix-inversion lemma} so that
\begin{multline}
\label{eq:gpmeastimes}
p(\mbf{x} | \mbf{y}) = \mathcal{N}\biggl(  \underbrace{\left( \pri{\mbf{P}}^{-1} + \mbf{C}^T \mbf{R}^{-1} \mbf{C} \right)^{-1} \left( \pri{\mbf{P}}^{-1} \pri{\mbf{x}}  + \mbf{C}^T \mbf{R}^{-1}\mbf{y} \right)}_{\est{\mbf{x}}, \;\mbox{\small mean}}, \biggr. \\ \biggl.\underbrace{\left( \pri{\mbf{P}}^{-1} + \mbf{C}^T \mbf{R}^{-1} \mbf{C} \right)^{-1} }_{\est{\mbf{P}}, \;\mbox{\small covariance}} \biggr).
\end{multline}
Further rearranging the mean expression, we have a linear system for $\est{\mbf{x}}$:
\begin{equation}
\label{eq:gpgnlin}
 \left( \pri{\mbf{P}}^{-1} + \mbf{C}^T \mbf{R}^{-1} \mbf{C} \right) \, \est{\mbf{x}} =  \pri{\mbf{P}}^{-1} \pri{\mbf{x}} + \mbf{C}^T \mbf{R}^{-1}\mbf{y}.
\end{equation}
The solution to this linear system gives us the estimate of the state at the measurement times.

If, after solving for the estimate at the measurement times, we later want to query the state at some other times of interest ($\tau_0 < \tau_1< \ldots < \tau_J$), we can use the classic \ac{GP}  interpolation equations \citep{rasmussen06} to do so:
\begin{subequations}
\label{eq:gpinterp}
\begin{eqnarray}
\est{\mbf{x}}_\tau & = & \pri{\mbf{x}}_\tau + \left( \pri{\mbf{P}}_\tau \pri{\mbf{P}}^{-1} \right) (\est{\mbf{x}} - \pri{\mbf{x}}), \\
\est{\mbf{P}}_{\tau\tau} & = & \pri{\mbf{P}}_{\tau\tau} + \left( \pri{\mbf{P}}_\tau \pri{\mbf{P}}^{-1}\right) \left( \est{\mbf{P}} - \pri{\mbf{P}} \right) \left( \pri{\mbf{P}}_\tau \pri{\mbf{P}}^{-1}\right)^T.
\end{eqnarray}
\end{subequations}
This is linear interpolation in the state variable (though not necessarily equivalent to linear interpolation in time).  

In general, this \index{Gaussian process} \ac{GP} approach has complexity $O(K^3 + K^2J)$ since the initial solve is $O(K^3)$ (for an arbitrary kernel function such as squared exponential) and the query is $O(K^2J)$; this is quite expensive, and next we will seek to improve the cost by designing a special kernel function.

\section{Exactly Sparse Kernels}

Looking at~\eqref{eq:gpgnlin}, we see that the {\em inverse kernel matrix}, $\pri{\mbf{P}}^{-1}$, appears on the left-hand side.  Naively choosing a \ac{GP} kernel function could result in this matrix being dense and therefore the solution expensive.  Instead, we will select our kernel function carefully so that the inverse kernel matrix is {\em exactly sparse}.

We base our kernel prior on a \ac{LTI}\footnote{We could also make this equation time-varying \citep{barfoot_ser24} but leave it time-invariant in the interest of simplicity.} \acf{SDE}:
\begin{equation}
\dot{\mbs{x}}(t) = \mbs{A} \mbs{x}(t) + \mbs{v}(t) + \mbs{L} \mbs{w}(t),
\end{equation}
with
\begin{equation}
\mbs{w}(t) \sim \mathcal{GP}(\mbf{0}, \mbs{Q} \, \delta(t-t^\prime)),
\end{equation}
a (stationary) zero-mean \index{Gaussian process} \ac{GP} with (symmetric, positive-definite) \index{power spectral density matrix} {\em power spectral density matrix}, $\mbs{Q}$.

Referring to \citet{barfoot_ser24}, we can show that the solution to this \ac{SDE} is a \ac{GP} with the following kernel function:
\begin{multline}
\mbs{x}(t) \sim \mathcal{GP} \biggl(  \underbrace{\mbs{\Phi}(t,t_0)\pri{\mbf{x}}_0 + \int_{t_0}^t \mbs{\Phi}(t,s) \mbs{v}(s) \, ds}_{\pri{\mbf{x}}(t)},  \\ \underbrace{\mbs{\Phi}(t,t_0) \pri{\mbf{P}}_0 \mbs{\Phi}(t^\prime,t_0)^T + \int_{t_0}^{\min(t,t^\prime)} \mbs{\Phi}(t,s) \mbs{L}\mbs{Q}\mbs{L}^T \mbs{\Phi}(t^\prime,s)^T  \, ds}_{\pri{\mbf{P}}(t,t^\prime)} \biggr),
\end{multline}
where $\mbs{\Phi}(t,s)$ is known as the {\em transition function}.  In our case, the system is \ac{LTI} and so the transition function is given by
\begin{equation}
\mbs{\Phi}(t,s) = \exp\left( \mbs{A}(t-s) \right).
\end{equation}
Evaluating the kernel function at the measurement times allows us to write our motion model prior as a joint Gaussian density:
\begin{equation}
\label{eq:gpprior}
p(\mbf{x}) = \mathcal{N}(\pri{\mbf{x}}, \pri{\mbf{P}}) = \mathcal{N} \left( \mbf{A} \mbf{v}, \mbf{A} \mbf{Q} \mbf{A}^T \right),
\end{equation}
where
\begin{gather}
\mbf{v} = \bbm \pri{\mbf{x}}_0 \\ \mbf{v}_{1,0} \\ \vdots \\ \pri{\mbf{x}}_k\\ \mbf{v}_{K,K-1} \ebm, \quad 
\mbf{Q} = \diag\left( \pri{\mbf{P}}_0, \mbf{Q}_{1,0}, \mbf{Q}_{2,1}, \ldots, \mbf{Q}_{K,K-1} \right), \\
\mbf{v}_{k,k-1} = \int_{t_{k-1}}^{t_k} \mbs{\Phi}(t_k,s) \mbs{v}(s) \, ds, \\
\mbf{Q}_{k,k-1} = \int_{t_{k-1}}^{t_k} \mbs{\Phi}(t_k,s) \mbs{L} \mbs{Q} \mbs{L}^T \mbs{\Phi}(t_k,s)^T \, ds, \nonumber \\
 \mbf{A} = \bbm \mbf{1} &  &  &  &   & \\
\mbf{A}_{1,0} & \mbf{1} &  &  &  &  \\
\mbf{A}_{2,0} & \mbf{A}_{2,1} & \mbf{1} & &  &  \\
\vdots & \vdots & \vdots & \ddots &  & \\
\mbf{A}_{K-1,0} & \mbf{A}_{K-1,1} & \mbf{A}_{K-1,2} & \cdots & \mbf{1} &  \\
\mbf{A}_{K,0} & \mbf{A}_{K,1} & \mbf{A}_{K,2} & \cdots & \mbf{A}_{K,K-1} & \mbf{1}
\ebm, \nonumber
\end{gather}
and we have simplified notation by writing $\mbf{A}_{k,\ell} = \mbs{\Phi}(t_k, t_\ell)$.

Substituting our specific prior into~\eqref{eq:gpgnlin}, the linear system that we need to solve becomes
\begin{equation}
\label{eq:gpgnlinsparse}
\left( \mbf{A}^{-T} \mbf{Q}^{-1} \mbf{A}^{-1} + \mbf{C}^T \mbf{R}^{-1} \mbf{C} \right) \, \est{\mbf{x}} =  \mbf{A}^{-T} \mbf{Q}^{-1} \mbf{v} + \mbf{C}^T \mbf{R}^{-1}\mbf{y},
\end{equation}
which as yet does not seem to offer any advantage over~\eqref{eq:gpgnlin}.  However, it turns out that $\mbf{A}^{-1}$ has the special form
\begin{equation}
\mbf{A}^{-1} = \bbm \mbf{1} &  & &  &  &  \\
-\mbf{A}_{1,0} & \mbf{1} &  &  &  &  \\
& -\mbf{A}_{2,1} & \mbf{1} & &  &  \\
 &  & -\mbf{A}_{3,2} & \ddots &  &  \\
 &  & & \ddots & \mbf{1} &  \\
 &  &  &  & -\mbf{A}_{K,K-1} & \mbf{1}
\ebm.
\end{equation}
Since $\mbf{A}^{-1}$ has only the main diagonal and the one below non-zero, and $\mbf{Q}^{-1}$ is block-diagonal, the inverse kernel matrix $\pri{\mbf{P}}^{-1} = (\mbf{A}\mbf{Q}\mbf{A}^T)^{-1} = \mbf{A}^{-T} \mbf{Q}^{-1} \mbf{A}^{-1}$ is exactly block-tridiagonal.  Since $\mbf{C}^T \mbf{R}^{-1} \mbf{C}$ is block-diagonal, it means the left-hand side of~\eqref{eq:gpgnlinsparse} is also block-tridiagonal.  This means that we can solve the linear system in $O(K)$ time instead of $O(K^3)$ for a dense inverse kernel matrix.  This has been made possible by our specific choice of kernel function, built from an \ac{LTI} \ac{SDE}.

We can also view~\eqref{eq:gpgnlinsparse} as the solution to the following least-squares optimization problem:
\begin{equation}
\est{\mbf{x}} = \mbox{arg}\min_{\mbf{x}} || \mbf{v} - \mbf{A}^{-1}\,\mbf{x} ||^2_{\mbf{Q}} + || \mbf{y} - \mbf{C}\mbf{x}||^2_{\mbf{R}}.
\end{equation}
Or, breaking it down by time steps,
\begin{multline}
\est{\mbf{x}} = \mbox{arg}\min_{\mbf{x}} || \pri{\mbf{x}}_0 - \mbf{x}_0 ||^2_{\pri{\mbf{P}}_0} + \sum_{k=1}^K || \mbf{x}_k - \mbf{A}_{k,k-1} \mbf{x}_{k-1} - \mbf{v}_{k,k-1} ||^2_{\mbf{Q}_{k,k-1}} \\ + \sum_{k=0}^K || \mbf{y}_k - \mbf{C}_k \mbf{x}_k||^2_{\mbf{R}_k}.
\end{multline}
It is clear from this expression that solving at the measurement times can be carried out using classical, discrete-time state estimation tools. Note that from the factor graph perspective presented in the next Chapter~\ref{chap:fg-ctls}, each of these terms accounts for exactly one factor.

\section{Querying is $O(1)$}

Finally, we show that querying the trajectory at any time $\tau$ is an $O(1)$ operation.  To do this, we use the interpolation equations~\eqref{eq:gpinterp}, where we notice that the product $\pri{\mbf{P}}_\tau \pri{\mbf{P}}^{-1}$ plays a critical role.  To simplify matters, let us assume that we have only a single query time, $t_{k-1} < \tau < t_k$.  Then we can show that
\begin{multline}
\pri{\mbf{P}}_\tau \pri{\mbf{P}}^{-1} = \Bigl[ \, \mbf{0} \;\;\; \cdots \;\;\;  \mbf{0} \;\;\; \underbrace{\mbf{A}_{\tau,k-1} - \mbf{Q}_{\tau,k-1} \mbf{A}_{k,\tau}^T \mbf{Q}_{k,k-1}^{-1} \mbf{A}_{k,k-1}}_{\mbs{\Lambda}_\tau\; \mbox{\small block column $k-1$}}  \\  \cdots \;\;\;  \underbrace{\mbf{Q}_{\tau,k} \mbf{A}_{k,\tau}^T \mbf{Q}_{k,k-1}^{-1}}_{\mbs{\Psi}_\tau,\;\mbox{\small block column $k$}}  \;\;\; \mbf{0} \;\;\; \cdots \;\;\; \mbf{0}  \, \Bigr],
\end{multline}
so that only block-columns $k-1$ and $k$ are non-zero.  This means that we only require the marginal associated with the joint density of $\mbf{x}_{k-1}$ and $\mbf{x}_k$ to compute the query at time $\tau$:
\beqn{}
\est{\mbf{x}}_\tau & = & \pri{\mbf{x}}_\tau + \bbm \mbs{\Lambda}_\tau & \mbs{\Psi}_\tau \ebm \left( \bbm \est{\mbf{x}}_{k-1} \\ \est{\mbf{x}}_{1} \ebm - \bbm \pri{\mbf{x}}_{k-1} \\  \pri{\mbf{x}}_{k} \ebm \right), \label{eq:meaninterp} \\
\est{\mbf{P}}_\tau & = & \pri{\mbf{P}}_\tau + \bbm \mbs{\Lambda}_\tau & \mbs{\Psi}_\tau \ebm \left( \bbm \est{\mbf{P}}_{k-1,k-1} & \est{\mbf{P}}_{k-1,k}^T \\ \est{\mbf{P}}_{k,k-1} & \est{\mbf{P}}_{k,k} \ebm \right.  \label{eq:covinterp} \\ & & \qquad\qquad\qquad\qquad \quad\left. - \bbm \pri{\mbf{P}}_{k-1,k-1} & \pri{\mbf{P}}_{k,k-1}^T \\ \pri{\mbf{P}}_{k,k-1} & \pri{\mbf{P}}_{k,k} \ebm \right) \bbm \mbs{\Lambda}_\tau^T \\ \mbs{\Psi}_\tau^T \ebm. \nonumber
\eeqn

The next chapter will reframe the \ac{GP} regression problem using factor graphs, which provides a new perspective on solving at the measurement times as well as interpolation.

%!TEX root =  main.tex
\chapter{Factor Graph Point of View}
\label{chap:fg-ctls}

As noted above, the previous chapter presents \ac{GP}-based continuous-time estimation from the perspective of classical \ac{GP} regression. In this chapter, we will consider this problem through the lens of \emph{factor graphs} -- a common paradigm in modern state estimation for robotics~\citep{dellaert17}. Crucially, this framework will allow us to build key insights into the inner workings of continuous-time estimation. 

We first provide a brief introduction to factor graphs in the next section. This section can be safely skipped by readers that are already familiar with this concept. For readers that are new to factor graphs, we recommend reading this section as well as the additional background in Appendix \ref{app:add-fg-bg} before continuing to the rest of this chapter. The rest of the chapter will reframe \ac{GP}-based continuous-time estimation using factor graphs and present insights that result from this reframing.

\section{A Brief Introduction to Factor Graphs}
\label{sec:brief-fg}

\begin{figure}[t]
     \includegraphics[width=\textwidth]{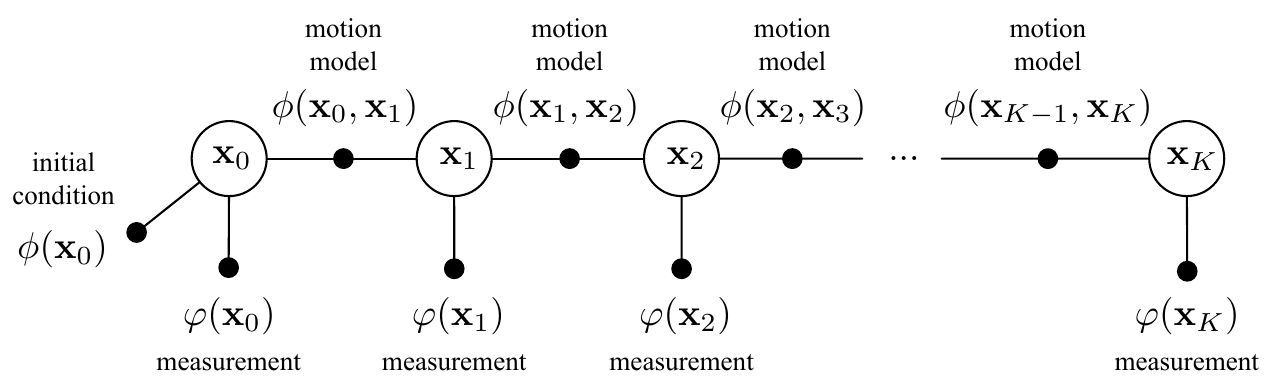}
     \caption{A simple factor graph for a classic estimation problem.  The unary factor on $\mbf{x}_0$ represents the prior knowledge about the initial state, the binary factors represent the motion model, and the remaining unary factors represent the measurements.  In practice, we can have much more complicated factor graphs with loops.}
     \label{fig:dtfg1}
\end{figure}

Typically in state estimation, we are interested in computing a posterior density representing the likelihood of a state,
\begin{equation}
p(\mbf{x} | \mbf{v}, \mbf{y}),
\end{equation}
where $\mbf{x} = (\mbf{x}_0, \ldots, \mbf{x}_K)$ is the state at times $t_0, \ldots, t_K$, $\mbf{v} = (\pri{\mbf{x}}_0, \mbf{v}_{1,0}, \ldots, \mbf{v}_{K,K-1})$ is a set of inputs/commands, and $\mbf{y} = (\mbf{y}_0, \ldots, \mbf{y}_K)$ is a set of measurements.  In most practical scenarios, we can factor this posterior density into a product of factors, each of which depends on only a subset of the variables in the problem.  For example, we might have
\begin{equation}
p(\mbf{x} | \mbf{v}, \mbf{y}) \propto \underbrace{p(\mbf{x}_0| \pri{\mbf{x}}_0)}_{\phi(\mbf{x}_0)} \prod_{k=1}^K \underbrace{p(\mbf{x}_k | \mbf{x}_{k-1}, \mbf{v}_{k,k-1})}_{\phi(\mbf{x}_{k-1},\mbf{x}_k)} \prod_{k=0}^K \underbrace{p(\mbf{y}_k | \mbf{x}_k)}_{\varphi(\mbf{x}_k)},
\end{equation}
where each $\phi(\cdot)$ and $\varphi(\cdot)$ is a {\em factor} of the posterior density corresponding to motion priors and measurements, respectively.  The factors can be {\em unary} (involving one variable), {\em binary} (involving two variables), or higher-order (involving more than two variables).

The normalization constant associated with this factorization is typically expensive to compute so we often satisfy ourselves with finding the $\mbf{x}$ that maximizes the posterior density, or equivalently the $\mbf{x}$ that minimizes the negative log-likelihood cost:
\begin{equation}\label{eq:cost}
\mathcal{C}(\mbf{x}) = -\log \phi(\mbf{x}_0) - \sum_{k=1}^K \log \phi(\mbf{x}_k, \mbf{x}_{k-1}) - \sum_{k=0}^K \log \varphi(\mbf{x}_k).
\end{equation}
This is a common approach in estimation, referred to as {\em \ac{MAP}} estimation.

We can visually understand the factors by assembling them into a {\em factor graph}, which is a bipartite graph with two types of nodes: variable nodes and factor nodes.  The variable nodes represent the variables in the problem, such as $\mbf{x}_0, \ldots, \mbf{x}_K$, and the factor nodes represent the factors, such as $\phi(\mbf{x}_0)$, $\phi(\mbf{x}_{k-1},\mbf{x}_k)$, and $\varphi(\mbf{x}_k)$.  The edges connect variable nodes to factor nodes, indicating which variables are involved in each factor.  This graphical representation allows us to see the dependencies between variables and factors in a very intuitive way.  Figure~\ref{fig:dtfg1} shows a simple factor graph for a classic discrete-time estimation problem.

\begin{figure}[t]
     \includegraphics[width=\textwidth]{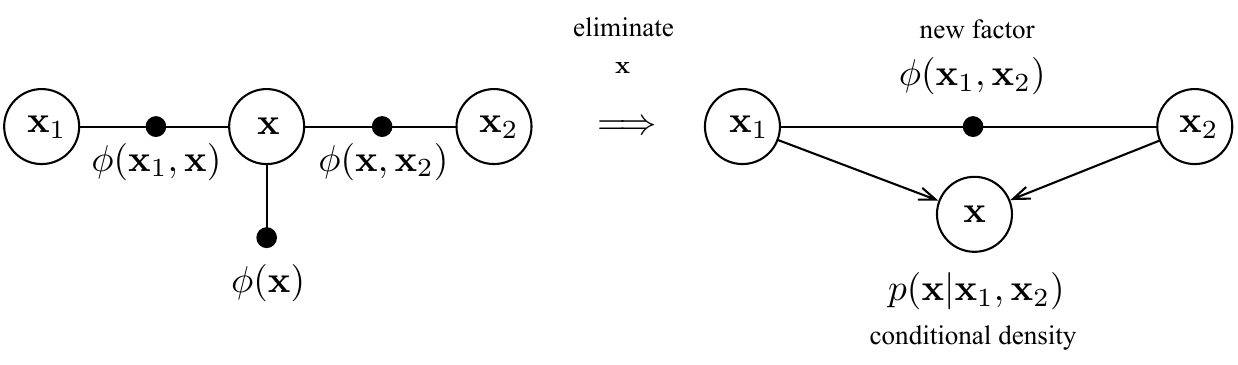}
     \centering
     \caption{Simple example of the elimination algorithm.  The left shows the factor graph before elimination and the right shows the result of eliminating $\mbf{x}$ from the factor graph, resulting in a new factor and a conditional density for $\mbf{x}$.}
     \label{fig:elim_main_body}
\end{figure}

Aside from the benefits to intuition, factor graphs also lay the groundwork for efficient computation via the \emph{elimination algorithm}. The key idea behind the iterations of this algorithm is shown in Figure~\ref{fig:elim_main_body}: we eliminate one variable at a time, and update the remaining factors in the graph accordingly. This update includes generating new factors between neighbouring, non-eliminated variables and identifying \emph{conditional dependencies} of the eliminated variables (indicated by arrows in the factor graph).
Once all variables have been eliminated, we are left with a structure -- the so-called \emph{Bayes net} --  that allows for efficient recovery of the optimal posterior density.

In the Appendix, we provide additional details about factor graphs, including the elimination algorithm (Section~\ref{sec:elim-algo}), show standard manipulations of factor graphs (Section~\ref{sec:fgelimfus}), and apply the factor graph approach to linear, discrete-time estimation (Section~\ref{sec:ldte}). We encourage readers that are not familiar with factor graphs to first read this appendix before proceeding to the next section.

\section{Factor Graphs for Linear Continuous-Time Estimation}% Formerly "Main Solve"
\begin{figure}[!h]
    \centering
    \includegraphics[width=0.95\textwidth]{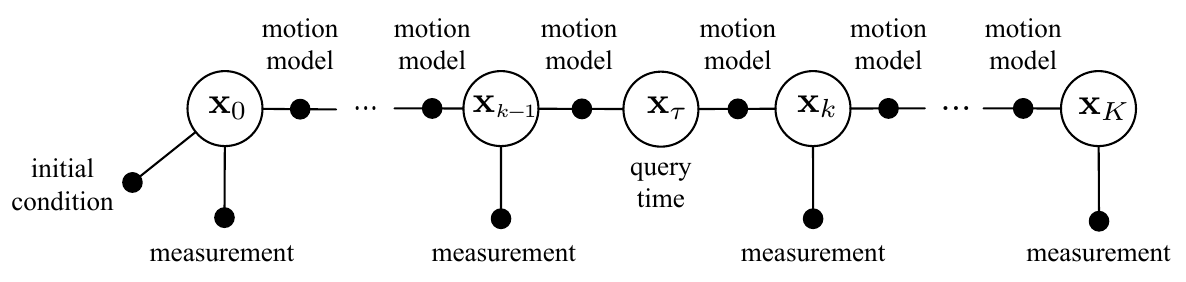}
    \caption{A continuous-time estimation problem with measurements at discrete times and a single query time.  We refer to this as `continuous-time' because the query(s) could be at any time along the trajectory.}
    \label{fig:ctfg}
\end{figure}

In the rest of this chapter, we will show that our insights from the previous chapter follows naturally when continuous-time estimation is cast into the factor graph framework. 

\begin{figure}[t]
     \centering
     \includegraphics[width=0.95\textwidth]{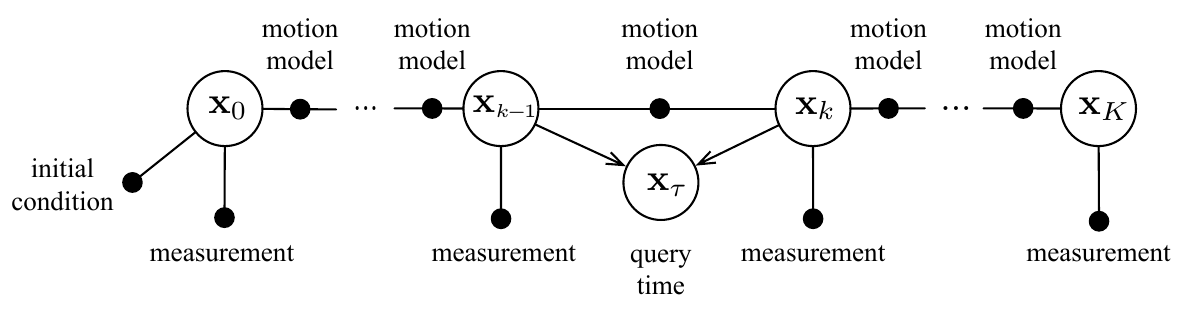}
     \caption{The same factor graph as Figure~\ref{fig:ctfg} with the state at the query time, $\mbf{x}_\tau$, eliminated.}
     \label{fig:ctfg2}
\end{figure}

We first focus on how \ac{GP}-based, continuous-time state-estimation problems can be solved exactly by applying a combination of standard discrete-time estimation tools and \ac{GP} interpolation equations~\citep{barfoot_rss14}. 
Consider the factor graph of a simple continuous-time estimation problem shown in Figure~\ref{fig:ctfg}. We have introduced a single query state, $\mbf{x}_{\tau}$, at arbitrary time, $\tau$, to the otherwise standard discrete-time factor graph. As one might expect from the previous chapter, we can break the problem down into two main steps:
\begin{enumerate}
    \item {\em main solve}: solve for the states at the measurement times
    \item {\em interpolation}: solve for the states at the query time
\end{enumerate}
The key to this decomposition is to use the elimination algorithm to remove the query state from the factor graph, resulting in the situation in Figure~\ref{fig:ctfg2}. The remaining factor graph then corresponds to a linear discrete-time factor graph, which we can readily solve with standard tools. Once we have found the solution at the measurement times, it follows that the distribution of the interpolated state, $\mbf{x}_{\tau}$, can be determined by the conditional dependence depicted by the arrows in Figure~\ref{fig:ctfg2}.  Importantly, the main solve step is completely agnostic to the query time, meaning we can select as many query times as we like after the main solve. The next two sections work through the details of these steps.

\subsection{Main Solve}
The first step in our strategy is to solve for the states at the measurement times in Figure~\ref{fig:ctfg2}.  We would like our solution to obey an important property:  whether we solve the factor-graph problem in Figure~\ref{fig:ctfg} ($\mbf{x}_\tau$ estimated as a variable in the main solve) or the one in Figure~\ref{fig:ctfg2} ($\mbf{x}_\tau$ computed afterwards), the solutions should agree.  This comes down to choosing the details of the motion model appropriately.  We will set up this motion model in this section, and show the desired property holds in the next section.

Recall our generic, continuous-time, linear system model from Section~\ref{sec:gp-regression}:
\begin{equation}\label{eq:linsysct}
\dot{\mbs{x}}(t) = \mbs{A} \mbs{x}(t) + \mbs{v}(t) + \mbs{L} \mbs{w}(t),
\end{equation}
where $\mbs{x}(t)$ is the state, $\mbs{v}(t)$ the input, and $\mbs{w}(t) \sim \mathcal{GP}(\mbf{0}, \mbs{Q} \delta(t-t^\prime))$ a white-noise process.  More concretely, choosing the coefficient matrices to be $\mbs{A} = \bbm \mbf{0} & \mbf{I} \\ \mbf{0} & \mbf{0} \ebm$ and $\mbs{L} = \bbm \mbf{0} \\ \mbf{I} \ebm$ and setting the inputs to zero, $\mbs{v}(t) = \mbf{0}$, our state represents position and velocity, $\mbs{x}(t) = \bbm \mbs{p}(t) \\ \dot{\mbs{p}}(t) \ebm$, and our motion model becomes {\em \ac{WNOA}}:
\begin{equation}
\bbm \dot{\mbs{p}}(t) \\ \ddot{\mbs{p}}(t) \ebm = \bbm \mbf{0} & \mbf{I} \\ \mbf{0} & \mbf{0} \ebm \bbm \mbs{p}(t) \\ \dot{\mbs{p}}(t) \ebm + \bbm \mbf{0} \\ \mbf{I} \ebm \mbs{w}(t) \quad \Leftrightarrow \quad \ddot{\mbs{p}}(t) = \mbs{w}(t).
\end{equation}
For now, we will work with~\eqref{eq:linsysct} due to its generality, but our examples will often use the specific case of \ac{WNOA}.

By analogy with discrete-time case,\footnote{See Appendix~\ref{app:add-fg-bg}.} we expect that our motion-prior factor between $\mbf{x}_{k-1}$ and $\mbf{x}_k$ should be of the form
\begin{multline}
     \phi(\mbf{x}_k, \mbf{x}_{k-1}) = -\log p(\mbf{x}_k | \mbf{x}_{k-1}, \mbf{v}_{k,k-1}) \\ \propto || \mbf{x}_k - \mbf{A}_{k,k-1} \mbf{x}_{k-1} - \mbf{v}_{k,k-1} ||^2_{\mbf{Q}_{k,k-1}}.
\end{multline}
To fill in the details, we need to (stochastically) integrate the motion model in~\eqref{eq:linsysct} over the time interval, $[t_{k-1}, t_{k}]$.  It can be shown that the solution to this \ac{SDE} is given by
\begin{equation}
p(\mbf{x}_k | \mbf{x}_{k-1}, \mbf{v}_{k,k-1}) = \mathcal{N}\left(\mbf{A}_{k,k-1} \mbf{x}_{k-1} + \mbf{v}_{k,k-1}, \mbf{Q}_{k,k-1} \right),
\end{equation}
where
\beqn{}
\mbf{v}_{k,k-1} & = &  \int_{t_{k-1}}^{t_{k}} \exp\left( \mbs{A} (t_{k} - s) \right) \mbs{v}(s) \, ds,   \\
\mbf{Q}_{k,k-1} & = &  \int_{t_{k-1}}^{t_{k}} \exp\left( \mbs{A} (t_{k} - s) \right) \mbs{L} \mbs{Q} \mbs{L}^T \exp\left( \mbs{A} (t_{k} - s)\right)^T \, ds,  \quad\;\;\;   \\
\mbf{A}_{k,k-1} & = &  \exp\left( \mbs{A} \, t_{k,k-1} \right),
\eeqn
with $t_{k,k-1} = t_{k} - t_{k-1}$.  We can often simplify these expressions for particular choices of $\mbs{A}$, $\mbs{L}$, and $\mbs{v}(t)$; for example, for \ac{WNOA} we have $\mbf{v}_{k,k-1} = \mbf{0}$ and
\begin{equation}
\mbf{Q}_{k,k-1} = \bbm \frac{1}{3}  t_{k,k-1}^3 \mbs{Q} & \frac{1}{2} t_{k,k-1}^2 \mbs{Q} \\  \frac{1}{2} t_{k,k-1}^2 \mbs{Q} &  t_{k,k-1} \mbs{Q} \ebm, \;\; \mbf{A}_{k,k-1} = \bbm \mbf{I} &  t_{k,k-1} \mbf{I} \\ \mbf{0} & \mbf{I} \ebm,
\end{equation}
This is everything we need to populate the motion-model factors in Figure~\ref{fig:ctfg2}.\footnote{See Figure~\ref{fig:dtfg2} for more details on the exact forms of the factors. Though this figure deals with the discrete-time case, the forms are equivalent for continuous-time.}  From here, we can use the \ac{RTS} smoother or any other solver (e.g., a factor graph solver) to solve for the states at the measurement times.  

\subsection{Querying the Solution}\label{sec:linafterquery}
% formerly After-Main-Solve Querying

\begin{figure}[t]
     \centering
     \includegraphics[width=0.95\textwidth]{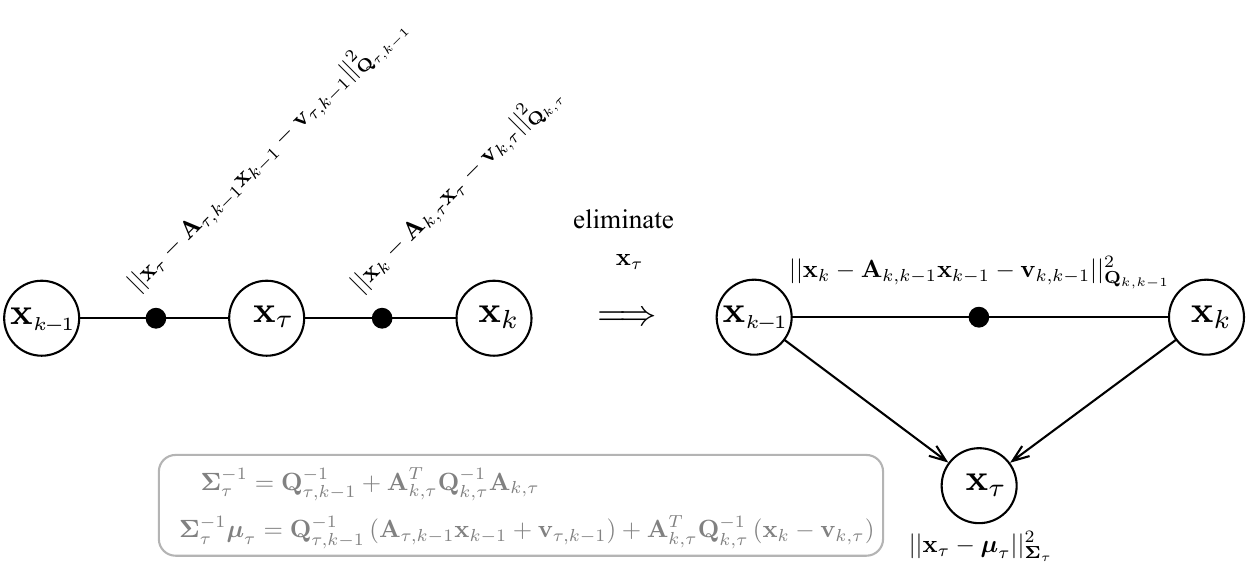}
     \caption{Elimination of the query state, $\mbf{x}_\tau$, from the factor graph in Figure~\ref{fig:ctfg2}.  The result is a new factor graph with a binary factor between the measurement states, $\mbf{x}_{k-1}$ and $\mbf{x}_{k}$, and a Gaussian conditional density for the query state.}
     \label{fig:gausselim2}
\end{figure}

In this section, we discuss querying the trajectory at some time(s) of interest, $\tau$, as in Figure~\ref{fig:ctfg2}, where the query state, $\mbf{x}_\tau$, does not have any measurements associated with it.  In this case, querying is a straightforward application of the elimination algorithm from Figure~\ref{fig:gausselim}.  In other words, we can use the Gaussian elimination formula,~\eqref{eq:gausselim}, to write
\begin{multline}
|| \mbf{x}_\tau - \mbf{A}_{\tau,k-1} \mbf{x}_{k-1} - \mbf{v}_{\tau,k-1} ||^2_{\mbf{Q}_{\tau,k-1}} + || \mbf{x}_{k} - \mbf{A}_{k,\tau} \mbf{x}_\tau - \mbf{v}_{k,\tau} ||^2_{\mbf{Q}_{k,\tau}}  \\ = || \mbf{x}_{k} - \mbf{A}_{k,k-1} \mbf{x}_{k-1} - \mbf{v}_{k,k-1} ||^2_{\mbf{Q}_{k,k-1}} + || \mbf{x}_\tau - \mbs{\mu}_\tau ||^2_{\mbs{\Sigma}_\tau},
\end{multline}
where $\mbs{\mu}_\tau$ and $\mbs{\Sigma}_\tau$ are the mean and covariance of the Gaussian conditional density for $\mbf{x}_\tau$ and are given in Figure~\ref{fig:gausselim2}, which depicts our elimination graphically.  

Critically, the new binary factor between the measurement states, $\mbf{x}_{k-1}$ and $\mbf{x}_{k}$, is identical to the form we would have used had we constructed it directly between these states to begin with; this ensures the presence/absence of the query state does not affect the solution.  It also means that we do not need to know about the query time when carrying out the main solve; we can simply solve for the states at the measurement times, and then query at any time(s) of interest later.

We want the posterior estimate for the query state, $p(\mbf{x}_\tau | \mbf{v}, \mbf{y})$, whereas we have the conditional density, $p(\mbf{x}_\tau | \mbf{x}_{k-1}, \mbf{x}_{k}) = \mathcal{N}(\mbs{\mu}_\tau, \mbs{\Sigma}_\tau)$.  We must therefore marginalize the separator states (i.e., convolve the conditional density with the posterior for the separator states) as follows:
\begin{eqnarray}
p(\mbf{x}_\tau | \mbf{v}, \mbf{y}) & = & \mathcal{N}\left( \est{\mbf{x}}_\tau, \est{\mbf{P}}_\tau \right) \nonumber \\ 
& = &  \int p(\mbf{x}_\tau | \mbf{x}_{k-1}, \mbf{x}_{k}) p(\mbf{x}_{k-1}, \mbf{x}_{k} | \mbf{v}, \mbf{y}) \, d\mbf{x}_{k-1} d\mbf{x}_{k}. \qquad
\end{eqnarray}
The marginal for the measurement states must be extracted from the main solve and is given by
\begin{equation}
p(\mbf{x}_{k-1}, \mbf{x}_{k} | \mbf{v}, \mbf{y}) = \mathcal{N}\left(\bbm \est{\mbf{x}}_{k-1} \\ \est{\mbf{x}}_{k} \ebm, \bbm \est{\mbf{P}}_{k-1} & \est{\mbf{P}}_{k,k-1}^T \\ \est{\mbf{P}}_{k,k-1} & \est{\mbf{P}}_{k} \ebm\right).
\end{equation}
Carrying out the convolution gives us the marginal posterior for the query state:
\beqn{fgquery}
\est{\mbf{x}}_\tau & = & \mbs{\eta}_\tau + \bbm \mbs{\Lambda}_\tau & \mbs{\Psi}_\tau \ebm \bbm \est{\mbf{x}}_{k-1} \\ \est{\mbf{x}}_{k} \ebm,  \\
\est{\mbf{P}}_\tau & = & \mbs{\Sigma}_\tau + \bbm \mbs{\Lambda}_\tau & \mbs{\Psi}_\tau \ebm \bbm \est{\mbf{P}}_{k-1} & \est{\mbf{P}}_{k,k-1}^T \\ \est{\mbf{P}}_{k,k-1} & \est{\mbf{P}}_{k} \ebm \bbm \mbs{\Lambda}_\tau^T \\ \mbs{\Psi}_\tau^T \ebm,
\eeqn
where
\beqn{}
\mbs{\eta}_\tau & = & \mbs{\Sigma}_\tau \left(\mbf{Q}_{\tau,k-1}^{-1} \mbf{v}_{\tau,k-1} - \mbf{A}_{k,\tau}^T \mbf{Q}_{k,\tau}^{-1} \mbf{v}_{k,\tau} \right), \\
\mbs{\Lambda}_\tau & = & \mbs{\Sigma}_\tau \mbf{Q}_{\tau,k-1}^{-1} \mbf{A}_{\tau,k-1}, \\ 
\mbs{\Psi}_\tau & = & \mbs{\Sigma}_\tau \mbf{A}_{k,\tau}^T \mbf{Q}_{k,\tau}^{-1}.
\eeqn
We can carry out this type of query for as many query times as we like.  The complexity of each query is $O(1)$, since we only need the marginal of the measurement states from the main solve. 

This result is equivalent to the traditional \ac{GP} interpolation result that we showed in the previous chapter. We only need show that
\beqn{}
\mbs{\eta}_t & = & \pri{\mbf{x}}_\tau - \bbm \mbs{\Lambda}_\tau & \mbs{\Psi}_\tau \ebm \bbm \pri{\mbf{x}}_{k-1} \\  \pri{\mbf{x}}_{k} \ebm, \\
\mbs{\Sigma}_t & = & \pri{\mbf{P}}_\tau - \bbm \mbs{\Lambda}_\tau & \mbs{\Psi}_\tau \ebm \bbm \pri{\mbf{P}}_{k-1,k-1} & \pri{\mbf{P}}_{k,k-1}^T \\ \pri{\mbf{P}}_{k,k-1} & \pri{\mbf{P}}_{k,k} \ebm \bbm \mbs{\Lambda}_\tau^T \\ \mbs{\Psi}_\tau^T \ebm,
\eeqn
which we leave as an exercise for the reader. However, we argue that the factor-graph approach is more intuitive and elegant because the sparse structure of the problem is plain to see; the separator for $\mbf{x}_\tau$ is clearly $\{ \mbf{x}_{k-1}, \mbf{x}_k\}$, so that the query will clearly be an $O(1)$ operation. 

\subsection{Example}

\begin{figure}[t]
     \centering
     \includegraphics[width=\textwidth]{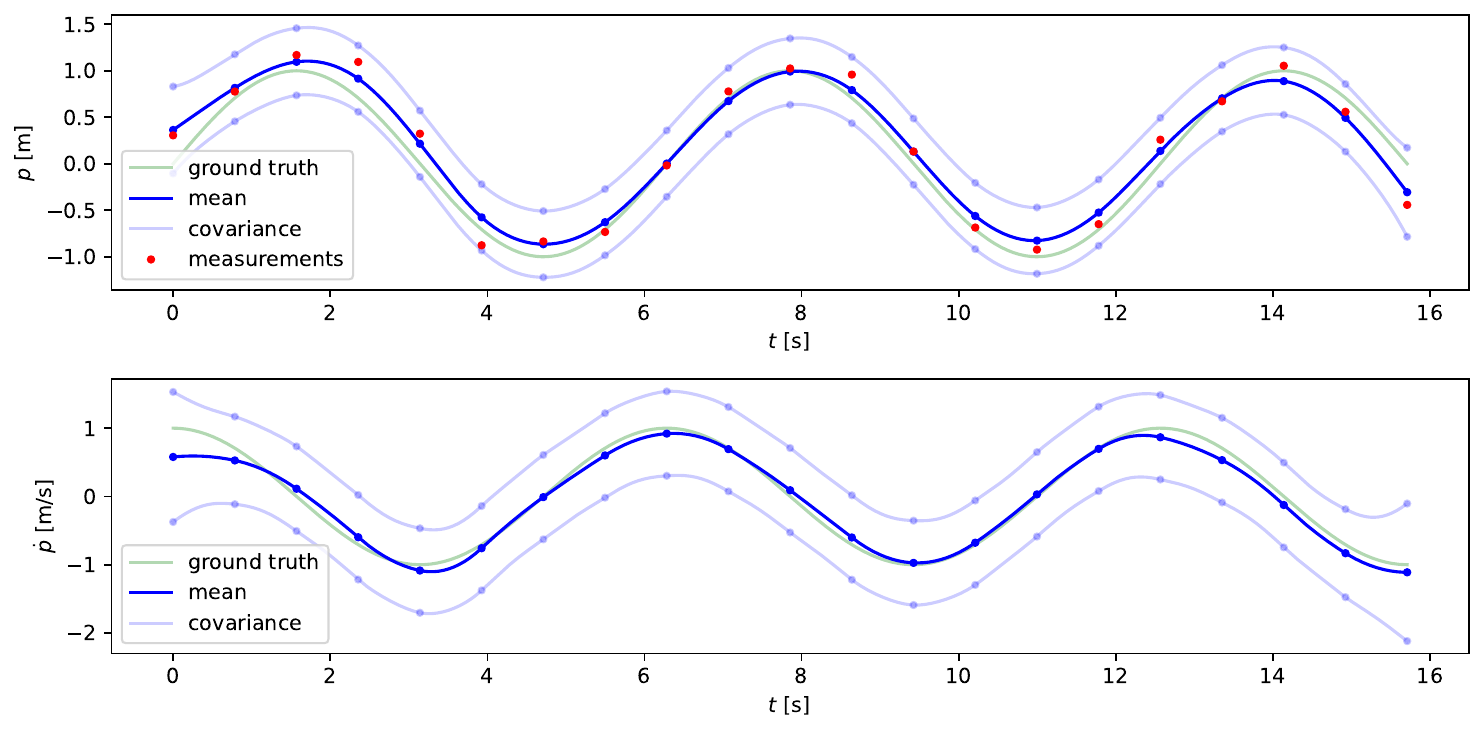}
     \caption{The ground-truth sinusoidal trajectory is shown in green.  Noisy position-only measurements are shown in red.  Our continuous-time solver first computes the solution at the measurement times (blue dots) and then interpolates (both mean and covariance) at several query times between the measurements times (blue line).  We see that the solution is smooth and continuous, as expected, following the ground truth closely.}
     \label{fig:linear_WNOA_example}
\end{figure}

As a simple example of our continuous-time estimation framework, we consider a system that moves according to a sinusoid,
\begin{equation}
p(t) = \sin(t),
\end{equation}
and only noisy measurements of this position are available at discrete times.  Our estimator does not know the true motion model, but instead models it as \ac{WNOA} with zero inputs.  The estimator therefore assumes the following system model:
\beqn{}
\bbm \dot{p}(t) \\ \ddot{p}(t) \ebm & = & \bbm 0 & 1 \\ 0 & 0 \ebm \bbm p(t) \\ \dot{p}(t) \ebm + \bbm 0 \\ 1 \ebm w(t), \\
y_k & = & \bbm 1 & 0 \ebm \bbm p(t_k) \\ \dot{p}(t_k) \ebm + n_k.
\eeqn
We follow the procedure outlined in the previous sections: (i) solve for the state at the measurement times, then (ii) interpolate at some query times.  Figure~\ref{fig:linear_WNOA_example} shows the results.

\section{Interpolating States Associated with Measurement}
\label{sec:interpolation_with_measurements}

\begin{figure}[t]
     \centering
     \includegraphics[width=0.90\textwidth]{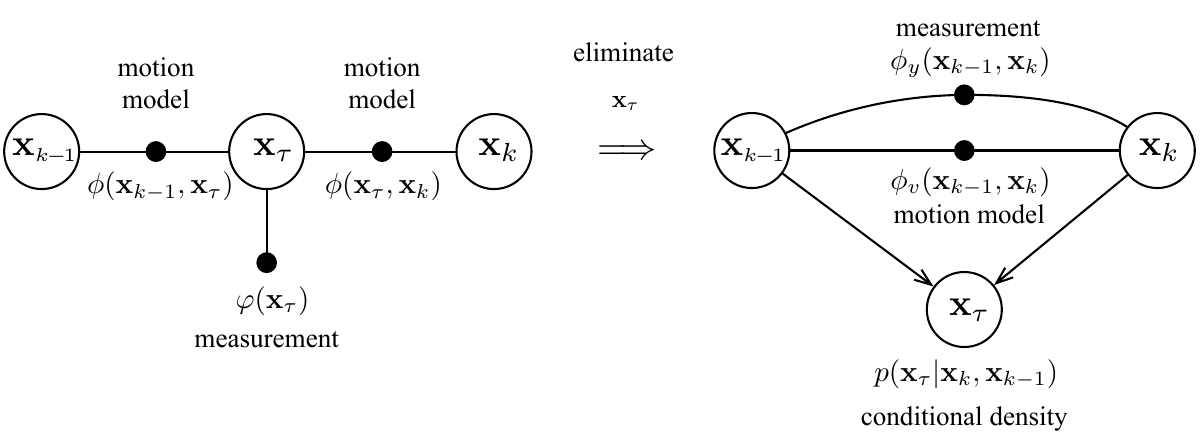}
     \caption{When eliminating an interpolated state with a measurement, we can exactly decompose the residual factor into two factors: a measurement factor $\phi_y(\mbf{x}_{k-1}, \mbf{x}_{k})$ and a motion-prior factor $\phi_v(\mbf{x}_{k-1}, \mbf{x}_{k})$.}
     \label{fig:elim-w-meas}
\end{figure}

Up to this point, we have only considered the case where our interpolated query states are not associated with any measurements. Owing to our particular choice of continuous-time motion model, we showed that elimination of these query states resulted in a new motion factor corresponding to the same motion model. 

We would also like to consider what happens when we eliminate a state that \emph{does} have an associated measurement, with the ultimate goal of incorporating measurements without needing to explicitly include an associated state in the optimization. Practically speaking, this is useful in two scenarios:
\begin{enumerate}
     \item It can reduce the size of the optimization problem when dealing with measurements from high-rate sensors, such as \ac{lidar}, \ac{radar}, or an \ac{imu}.
     \item It can allow us to regularly space our optimization states in time even when measurements are asynchronous. This regular spacing can be conducive to more accurate approximation of the posterior density \citep{zhang_tro24}
\end{enumerate}

As shown in Figure~\ref{fig:elim-w-meas}, when we eliminate a single state with a measurement, we get \emph{two} residual factors: one that recovers the motion model (as in the last section) and one that represents the \emph{effect} of the measurement. 
To see why this is true, consider that the probability density function of the factor graph on the left of Figure~\ref{fig:elim-w-meas} can be factored into a term that contains only measurement factors and a term that contains only motion-model prior factors,
\begin{equation}
     p(\mbf{x}_k, \mbf{x}_{\tau} | \mbf{x}_{k-1}, \mbf{y}, \mbf{v}) \propto p(\mbf{y}| \mbf{x}_{\tau}) p(\mbf{x}_k, \mbf{x}_{\tau}| \mbf{x}_{k-1}, \mbf{v}).
\end{equation}
Assuming a motion prior of the type discussed in this article, it turns out that the second term can be refactored as
\begin{equation}
     p(\mbf{x}_k, \mbf{x}_{\tau} | \mbf{x}_{k-1}, \mbf{v}) = p(\mbf{x}_{\tau}|\mbf{x}_k, \mbf{x}_{k-1}, \mbf{v}) p(\mbf{x}_k|\mbf{x}_{k-1}, \mbf{v}_{k,k-1}), 
\end{equation} 

The residual factor, $\phi(\mbf{x}_{k-1}, \mbf{x}_{k})$, is obtained by marginalizing\footnote{Recall that marginalization (a probabilistic operation) and elimination (a linear-algebra operation) are equivalent for Gaussian variables and linear factors.} the full joint distribution with respect to the interpolated variable:
\begin{equation}
     \phi(\mbf{x}_{k-1}, \mbf{x}_{k}) = \underbrace{\int p(\mbf{y}| \mbf{x}_{\tau}) p(\mbf{x}_{\tau}|\mbf{x}_k, \mbf{x}_{k-1}, \mbf{v})  \, d\mbf{x}_{\tau}}_{\phi_y(\mbf{x}_{k-1}, \mbf{x}_{k})} \, \underbrace{p(\mbf{x}_k|\mbf{x}_{k-1}, \mbf{v}_{k,k-1})}_{\phi_v(\mbf{x}_{k-1}, \mbf{x}_{k})},
\end{equation}
where we have moved the rightmost term out of the integral since it does not depend on $\mbf{x}_{\tau}$. The second factor in the equation above is exactly the motion prior:
\begin{equation}
     \phi_v(\mbf{x}_{k-1}, \mbf{x}_{k}) = \mathcal{N}\left(\mbf{A}_{k,k-1}\mbf{x}_{k-1} + \mbf{v}_{k,k-1}, \mbf{Q}_{k,k-1} \right),
\end{equation}

Assuming that the measurement is linear, $p(\mbf{y}| \mbf{x}_{\tau})=\mathcal{N}\left(\mbf{C} \mbf{x}_{\tau}, \mbf{R}\right)$, it is fairly straightforward to show that this measurement factor has the following distribution:
\begin{equation}\label{eq:wrapped_factor_simple}
     \phi_{y}(\mbf{x}_{k-1}, \mbf{x}_{k}) = \mathcal{N}\left(\mbf{C} \check{\mbf{x}}_{\tau}, \mbf{C} \mbf{\Sigma}_{\tau}\mbf{C}^T + \mbf{R} \right).
\end{equation}
The interpretation of this distribution is intuitive: the mean is obtained by applying the measurement model to the bordering states after mapping through the motion model to the measurement time via interpolation. The covariance of the measurement is inflated proportionate to the temporal distance from the bordering states to measurement times, accounting for the uncertainty of the model. 

The now-eliminated state $\mbf{x}_{\tau}$ can actually be dropped from the optimization,\footnote{We cannot completely forget about this state if we are dealing with a linearization of a nonlinear state-estimation problem. In this case, it is important to keep a record of $\mbf{x}_{\tau}$ to update the linearization point at each iteration.} leaving a motion prior and a new measurement factor between the non-interpolated states, $\mbf{x}_{k-1}$ and $ \mbf{x}_{k}$. In effect, we have removed $\mbf{x}_{\tau}$ from the optimization problem, but maintained the effect of its measurement.

A similar situation also holds when we eliminate multiple query states with measurements between two non-interpolated states. However, we need to make an additional approximation to transform the original measurements into \emph{separate} factors. The interested reader can find a more detailed derivation for the case of multiple factors in Appendix~\ref{app:interp-during-solve}.

%!TEX root =  FnTarticle.tex

\chapter{Continuous-Time Systems on Lie Groups}\label{chapLieGroups}

In this chapter, we will adapt the factor-graph approach to continuous-time systems on Lie groups, which is a common representation for robot motion.  To do so will require some approximation, but we will show that the approximation is very good in practice.  

\section{Background on Lie Groups}

Lie groups are widely used in several fields including robotics, computer vision, and graphics (see e.g., \citet{chirikjian01,stillwell08,absil09,chirikjian09,chirikjian16,sola18,gallierDifferentialGeometryLie2020,boumal23}).  Within robotics, they are employed within modelling \citep{deleuterio85,park95,lynch17}, state estimation \citep{wang06,wang08,bonnabel08,blanco10,wolfe11,long12,forster15,dellaertFactorGraphsRobot2017,bonnabel17,mahony21,barfoot_ser24}, and control theory \citep{murray94,bullo95,sastry99}.  They provide the mathematical tools to describe important concepts such as rotations and rigid-body motions, to name a few.  We will primarily use the notation of \citet{barfoot_ser24} with some improvements from~\citet{sola18} and refer to those sources for more details.

We consider a general Lie group quantity $\Tsmall \in G$, where $G$ is a Lie group.  The Lie group is a smooth manifold with a group structure, which means that it has a smooth multiplication operation and an inverse operation.
The tangent space at the identity element of the Lie group is called the Lie algebra, denoted $\mathfrak{g}$.  
Although the elements of the Lie algebra often have non-trivial matrix structures, the Lie algebra itself is an $m$-dimensional vector space, so we can represent each element using a coordinate vector, $\mbs{\xi}\in\Real^m$. For each Lie group, we can define a unique, isomorphic mapping (called the {\em wedge operator}), $\wdg:~\Real^m\rightarrow \mathfrak{g}$, as well as its associated inverse map, $\vee:~\mathfrak{g}\rightarrow \Real^m$, so that $\mbs{\xi}^\wdg \in \mathfrak{g}$ is in the Lie algebra . Given this isomorphism, we often refer to $\mbs{\xi}$ as an element of the Lie algebra.

The exponential map, $\mbox{exp}: \mathfrak{g} \to G$, maps elements of the Lie algebra to the Lie group, while the logarithm map, $\mbox{log}: G \to \mathfrak{g}$, maps elements of the Lie group back to the Lie algebra: 
\begin{equation}
\Tsmall = \mbox{exp}(\mbs{\xi}^\wdg)= \mbox{Exp}(\mbs{\xi}), \quad \mbs{\xi} = \mbox{log}(\Tsmall)^\vee=\mbox{Log}(\Tsmall),
\end{equation}
where $\mbox{exp}$ and $\mbox{log}$ are the {\em matrix exponential and logarithm},\footnote{For general manifolds, these maps will not necessarily correspond to the matrix exponential and logarithm, but for the matrix Lie groups used in robotics, this is most often the case.} respectively, and we define convenience operators, $\mbox{Exp}$ and $\mbox{Log}$, following~\citet{solaMicroLieTheory2021}.

The interested reader can find several examples of Lie groups, their Lie algebras and mapping operators in~\citet{solaMicroLieTheory2021} and~\citet{barfoot_ser24}.
Some common Lie groups include the special orthogonal group $SO(3)$, which represents rotations in three-dimensional space, and the special Euclidean group $SE(3)$, which represents rigid-body motions in three-dimensional space.  The Lie algebra for $SO(3)$ is $\mathfrak{so}(3)$, which can be represented as a skew-symmetric matrix, while the Lie algebra for $SE(3)$ is $\mathfrak{se}(3)$, which can be represented as a six-dimensional vector (the twist) that combines rotation and translation.

For optimization on Lie groups, what we really want is a well-defined way to perturb an element of the Lie group.
For the groups in which we are interested, such as $SE(3)$, there are constraints on the scalar elements of the Lie group elements, $\mbf{T}$.  This makes it a bit challenging to work with these elements directly, so we will often work with the Lie algebra elements, $\mbs{\xi}$, instead (since they are unconstrained). 

The exponential map, $\mbox{Exp}(\mbs{\xi})$, is a local diffeomorphism around the identity element of the Lie group, which means that we can use it to define Lie group elements around the identity. To perturb a general Lie group element, we can simply multiply by a perturbation of Identity,
\begin{equation}\label{eq:lie-perturb}
\Tsmall = \Tsmall_{\rm op}\mbox{Exp}(\mbs{\xi}) \approx \Tsmall_{\rm op} \underbrace{(\mbf{I}+\mbs{\xi}^\wdg)}_{\substack{\text{approx.} \\\text{perturb. of}\\\mbf{I}}} = \underbrace{\Tsmall_{\rm op} + \Tsmall_{\rm op} \mbs{\xi}^\wdg}_{\substack{\text{approx.} \\\text{perturb. of}\\\Tsmall_{\rm op}}},
\end{equation}
where $\Tsmall_{\rm op}$ is the operating point of the perturbation. The approximation in~\eqref{eq:lie-perturb} is only valid if $\mbs{\xi}$ is small, but is quite convenient when linearizing error terms involving elements of a Lie group.  For example, in state estimation, it is common to optimize over measurement-based error functions of the following form: 
\begin{equation}
     \mbf{e} = \mbf{y} - \mbf{f}(\mbf{T}^{-1} \mbf{p}),
\end{equation}
where $\mbf{y}$ is a measurement, $\mbf{T}$ is a Lie group element representing a pose, $\mbf{p}$ is the position of a point or feature of interest, and $\mbf{f}$ is some arbitrary measurement function. We can linearize this model using the technique above,
\begin{multline}
\mbf{e} = \mbf{y} - \mbf{f}(\mbf{T}^{-1} \mbf{p}) = \mbf{y} - \mbf{f}\left(\mbox{Exp}(-\mbs{\xi}) \mbf{T}_{\rm op}^{-1} \mbf{p}\right) \\ \approx \mbf{y} - \mbf{f}\left((\mbf{I}-\mbs{\xi}^\wdg) \mbf{T}_{\rm op}^{-1} \mbf{p}\right) = \underbrace{\mbf{y} - \mbf{f}(\mbf{T}_{\rm op}^{-1} \mbf{p})}_{\mbf{e}_{\rm op}} + \frac{\partial \mbf{f}}{\partial \mbf{T}^{-1}\mbf{p}} \bigg|_{\mbf{T}_{\rm op}^{-1} \mbf{p}} \mbs{\xi}^\wdg \mbf{T}_{\rm op}^{-1} \mbf{p} \\ =\mbf{e}_{\rm op} + \mbf{E} \, \mbs{\xi},
\end{multline}
where $\mbf{E}$ is the Jacobian of the error with respect to the perturbation $\mbs{\xi}$.  This simple approach can be used to linearize factors comprising elements of Lie groups.

Representing uncertainty on Lie groups is also facilitated by the exponential map and a Lie algebra.  We would like to have a Gaussian distribution on the Lie group.  To accomplish this we define a Gaussian distribution on the Lie algebra, and then map it to the Lie group as follows:           
\begin{equation}
\mbf{T} \sim \mathcal{N}(\est{\mbf{T}}, \est{\mbf{P}}) \quad \Leftrightarrow \quad \mbf{T} = \est{\mbf{T}} \, \mbox{Exp}(\mbs{\xi}), \quad \mbs{\xi} \sim \mathcal{N}(\mbf{0}, \est{\mbf{P}}).
\end{equation}
In other words, we define a zero-mean Gaussian distribution on the Lie algebra, and then map it to the Lie group using the exponential map and the closure property of the Lie group.  This allows us to represent uncertainty on the Lie group in a way that is consistent with the geometry of the Lie group.\footnote{For more details on representing uncertainty on Lie groups, see~\cite{barfoot_tro14}.}  This approach is only appropriate for fairly concentrated Gaussians.

\section{Motion Priors on Lie Groups}

For Lie groups, a kinematic motion model is typically of the form
\begin{equation}
\dot{\Tsmall}(t) = \Tsmall(t) \mbs{\varpi}(t)^\wdg,
\end{equation}
where $\Tsmall(t)$ is the kinematic configuration expressed in some Lie group (e.g., $SE(3)$), $\mbs{\varpi}(t)$ is the generalized velocity (e.g., $\mathbb{R}^6$ twist for $SE(3)$), and $\wdg$ is the wedge operator for the Lie group. Here we assume that $\Tsmall$ maps homogeneous points from a local frame to a global frame. The challenge with this model is that not only is it nonlinear, the state does not live in a vector space.  Alternatively, we could write this same motion model in the Lie algebra as
\begin{equation}
\dot{\mbs{\xi}}(t) = \Jbig(\mbs{\xi}(t))^{-1} \mbs{\varpi}(t),
\end{equation}
where $\mbs{\xi}(t)$ is the state in the Lie algebra (i.e., $\Tsmall(t) = \mbox{Exp}(\mbs{\xi}(t))$) and $\Jbig(\mbs{\xi}(t))$ is the (right) Jacobian of the exponential map from the Lie algebra to the (adjoint of the) Lie group.\footnote{The derivation of these terms is quite technical, but is provided in detail in~\cite{barfoot_ser24}.}  The state now lives in vector space, but the motion model is still nonlinear.

To directly leverage our earlier results, where we employed linear systems to build our motion priors, we could choose a linear motion model in the Lie algebra.  For example, a \ac{WNOA} model can be written down as
\begin{equation}
\ddot{\mbs{\xi}}(t) = \mbs{w}(t),
\end{equation}
where $\mbs{w}(t)$ is a white-noise process.  As a first-order system, this can be written as
\begin{equation}
\begin{bmatrix}
\dot{\mbs{\xi}}(t) \\
\ddot{\mbs{\xi}}(t)
\end{bmatrix} = 
\begin{bmatrix}
\mbf{0} & \mbf{I} \\
\mbf{0} & \mbf{0}
\end{bmatrix}
\begin{bmatrix}
\mbs{\xi}(t) \\
\dot{\mbs{\xi}}(t)
\end{bmatrix} +
\begin{bmatrix}
\mbf{0} \\
\mbf{I}
\end{bmatrix}
\mbs{w}(t), 
\end{equation}
which matches the form we assumed in~\eqref{eq:linsysct}.  This linear model is only useful near the identity element of the Lie group, so we will need several of these motion models initiated at key times along the trajectory.  In general, we assume that we can write a motion model as
\begin{equation}
\dot{\mbs{\gamma}}(t) = \mbs{A} \mbs{\gamma}(t) + \mbs{v}(t) + \mbs{L} \mbs{w}(t),
\end{equation}
where $\mbs{\gamma}(t)$ is a stacking of $\mbs{\xi}(t)$ and an appropriate number of its derivatives. This looks just like the models we were using for our linear systems.  We will refer to $\mbs{\gamma}(t)$ as the {\em local \ac{GP} variable}.

\begin{figure}[t]
     \centering
     \includegraphics[width=0.70\textwidth]{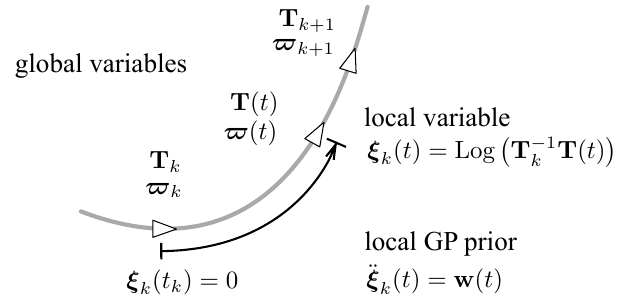}
     \caption{To make use of \acp{GP} on a Lie group, we define local variables in the Lie algebra around each measurement time and apply linear \ac{GP} models to these local variables.}
     \label{fig:localgp}
\end{figure}

\begin{figure}[t]
     \centering
     \includegraphics[width=\textwidth]{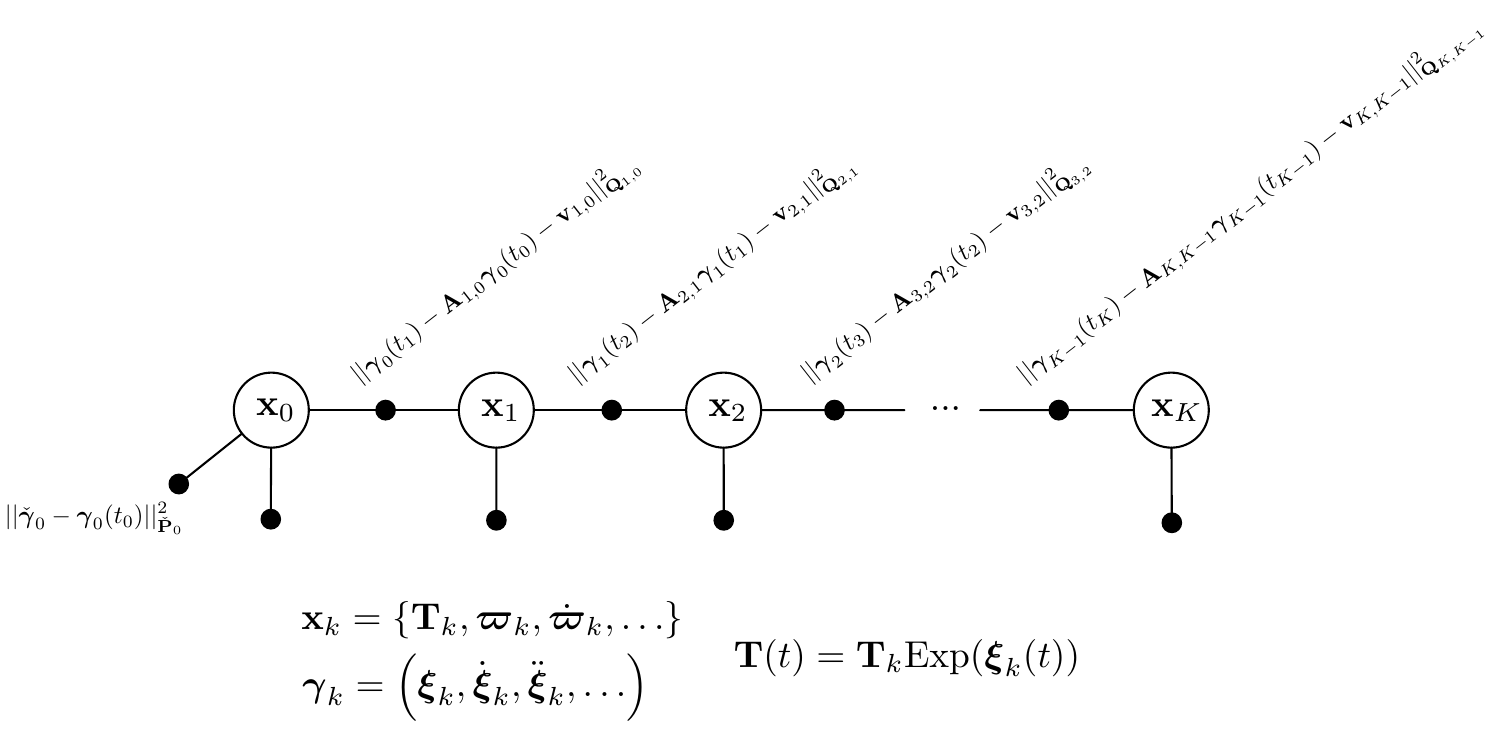}
     \caption{General form of a motion prior defined over a Lie group trajectory.  Between each pair of measurement times, we use a linear motion model in the Lie algebra.  The variables for which we are actually optimizing live in the Lie group.}
     \label{fig:lgfg}
\end{figure}

To stitch several of these models together, for example one per measurement time, we can redefine $\mbs{\xi}(t)$ to be
\begin{equation}\label{eq:local-lie-var}
\mbs{\xi}_k(t) = \mbox{Log} \left(\Tsmall_k^{-1} \Tsmall(t)\right),
\end{equation}
where $\Tsmall_k$ is the Lie group element at the $k$th measurement time.  Figure~\ref{fig:localgp} depicts this idea.  Then, from one measurement time to the next, we can write
\begin{multline}
p\left(\mbs{\gamma}_{k-1}(t_k) | \mbs{\gamma}_{k-1}(t_{k-1}), \mbf{v}_{k,k-1}\right) \\ = \mathcal{N}\left( \mbf{A}_{k,k-1} \mbs{\gamma}_{k-1}(t_{k-1}) + \mbf{v}_{k,k-1}, \mbf{Q}_{k,k-1}\right),
\end{multline}
recalling our linear results to compute $\mbf{A}_{k,k-1}$, $\mbf{v}_{k,k-1}$, and $\mbf{Q}_{k,k-1}$.  Our negative log-likelihood cost term is then
\begin{equation}\label{eq:lgmp}
|| \underbrace{\mbs{\gamma}_{k-1}(t_k) - \mbf{A}_{k,k-1} \mbs{\gamma}_{k-1}(t_{k-1}) - \mbf{v}_{k,k-1}}_{\mbf{e}_{k,k-1}} ||_{\mbf{Q}_{k,k-1}}^2.
\end{equation}
We can leave our motion model in this form and rely on the chain rule during optimization, but we can also use \eqref{eq:local-lie-var} to re-express the local $\mbs{\gamma}_k(t)$ variables in terms of Lie group elements and generalized velocities.  For example, if we want to use \ac{WNOA} as our motion model, we can substitute in the definitions of $\mbs{\xi}_k(t)$ to see that
\begin{equation}
\mbs{\gamma}_{k-1}(t) = \bbm \mbox{Log} \left(\Tsmall_{k-1}^{-1} \Tsmall(t)\right) \\ \Jbig\left(\mbox{Log} \left(\Tsmall_{k-1}^{-1} \Tsmall(t)\right)\right)^{-1} \mbs{\varpi}(t) \ebm,
\end{equation}
and at $t=\{t_k, t_{k-1}\}$,
\begin{equation}\label{eq:gamma-tk-tkm1}
\mbs{\gamma}_{k-1}(t_k) = \bbm \mbox{Log} \left(\Tsmall_{k-1}^{-1} \Tsmall_k\right) \\ \Jbig\left(\mbox{Log} \left(\Tsmall_{k-1}^{-1} \Tsmall_k\right)\right)^{-1} \mbs{\varpi}_k \ebm, \quad \mbs{\gamma}_{k-1}(t_{k-1}) = \bbm \mbf{0} \\ \mbs{\varpi}_{k-1} \ebm.
\end{equation}
Substituting \eqref{eq:gamma-tk-tkm1} into \eqref{eq:lgmp}, the negative log-likelihood cost term becomes
\begin{equation}
\left|\left| \bbm \mbox{Log} \left(\Tsmall_{k-1}^{-1} \Tsmall_k\right) - t_{k,k-1} \mbs{\varpi}_{k-1} \\ \Jbig\left(\mbox{Log} \left(\Tsmall_{k-1}^{-1} \Tsmall_k\right)\right)^{-1} \mbs{\varpi}_k - \mbs{\varpi}_{k-1} \ebm \right|\right|_{\mbf{Q}_{k,k-1}}^2.
\end{equation}
The top row is comparing the change in $\Tsmall(t)$ to the generalized velocity times the time difference, while the bottom row is essentially comparing the generalized velocity at the two times.  Figure~\ref{fig:lgfg} shows the general form of a motion prior defined over a Lie-group trajectory.

\section{Main Solve}

To make use of our earlier results involving linear systems, we will need to somehow linearize our motion model defined over a Lie group.  To do this, we assume that we can carry out perturbations of the Lie group element and the generalized velocity and its derivatives as follows:
\begin{equation}
\Tsmall_k = \Tsmall_{{\rm op},k} \mbox{Exp}(\mbs{\ep}_k), \quad \mbs{\varpi}_k = \mbs{\varpi}_{{\rm op},k} + \mbs{\eta}_k, \quad \ldots
\end{equation}
where $\mbf{x}_{{\rm op},k} = \{\Tsmall_{{\rm op},k}, \mbs{\varpi}_{{\rm op},k}, \ldots \}$ is the operating point of the linearization and $\mbs{\varepsilon}_k = (\mbs{\ep}_k, \mbs{\eta}_k, \ldots)$ is the perturbation.  

\begin{figure}[t]
     \centering
     \includegraphics[width=0.95\textwidth]{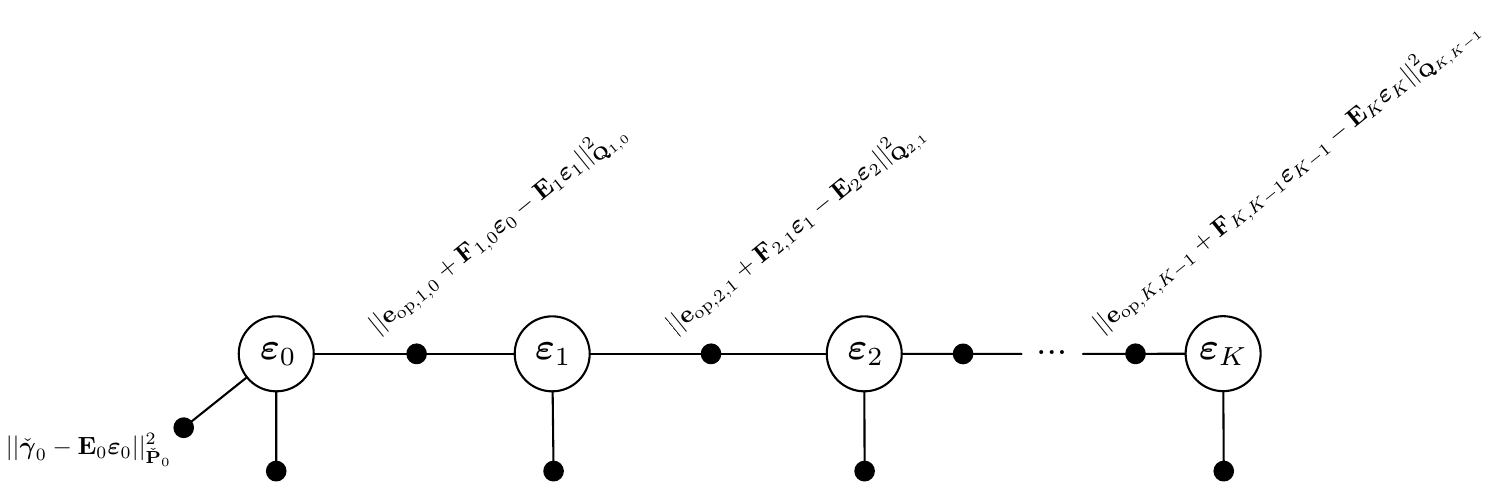}
     \caption{Linearized Lie-group motion prior.  The factor graph is now expressed over the perturbation states rather than the original Lie group variables.  We can use our linear system results to solve this factor graph, then iterate to improve the linearization.}
     \label{fig:ctfg3}
\end{figure}

We assume then that we can linearize the error term in~\eqref{eq:lgmp} as
\begin{equation}
\mbf{e}_{k,k-1} \approx \mbf{e}_{{\rm op},k,k-1} + \mbf{F}_{k,k-1} \mbs{\varepsilon}_{k-1} - \mbf{E}_k \mbs{\varepsilon}_k,
\end{equation}
where $\mbf{F}_{k,k-1}$ and $\mbf{E}_k$ are the Jacobians of the error term with respect to the perturbations.  Figure~\ref{fig:ctfg3} shows the linearized factor graph, which is now expressed over the perturbation states rather than the original Lie group variables.  We can use our linear system results to solve this factor graph, then iterate to convergence.  At each iteration, after finding the optimal perturbations, we update the operating points $\mbf{x}_{{\rm op},k}$ and recompute the Jacobians $\mbf{F}_{k,k-1}$ and $\mbf{E}_k$; in detail the update scheme is
\begin{equation}
\Tsmall_{{\rm op},k} \leftarrow \Tsmall_{{\rm op},k} \mbox{Exp}(\mbs{\ep}_k), \quad \mbs{\varpi}_{{\rm op},k} \leftarrow \mbs{\varpi}_{{\rm op},k} + \mbs{\eta}_k, \quad \ldots
\end{equation}
which ensures the Lie group element remains in the Lie group.

\section{After-Main-Solve Querying}
\label{sec:lie-after-query}
After the main solve, we can query the states at any time by interpolating between the measurement times.  We will do this in slightly different ways for the mean and the covariance.

\begin{figure}[t]
     \centering
     \includegraphics[width=\textwidth]{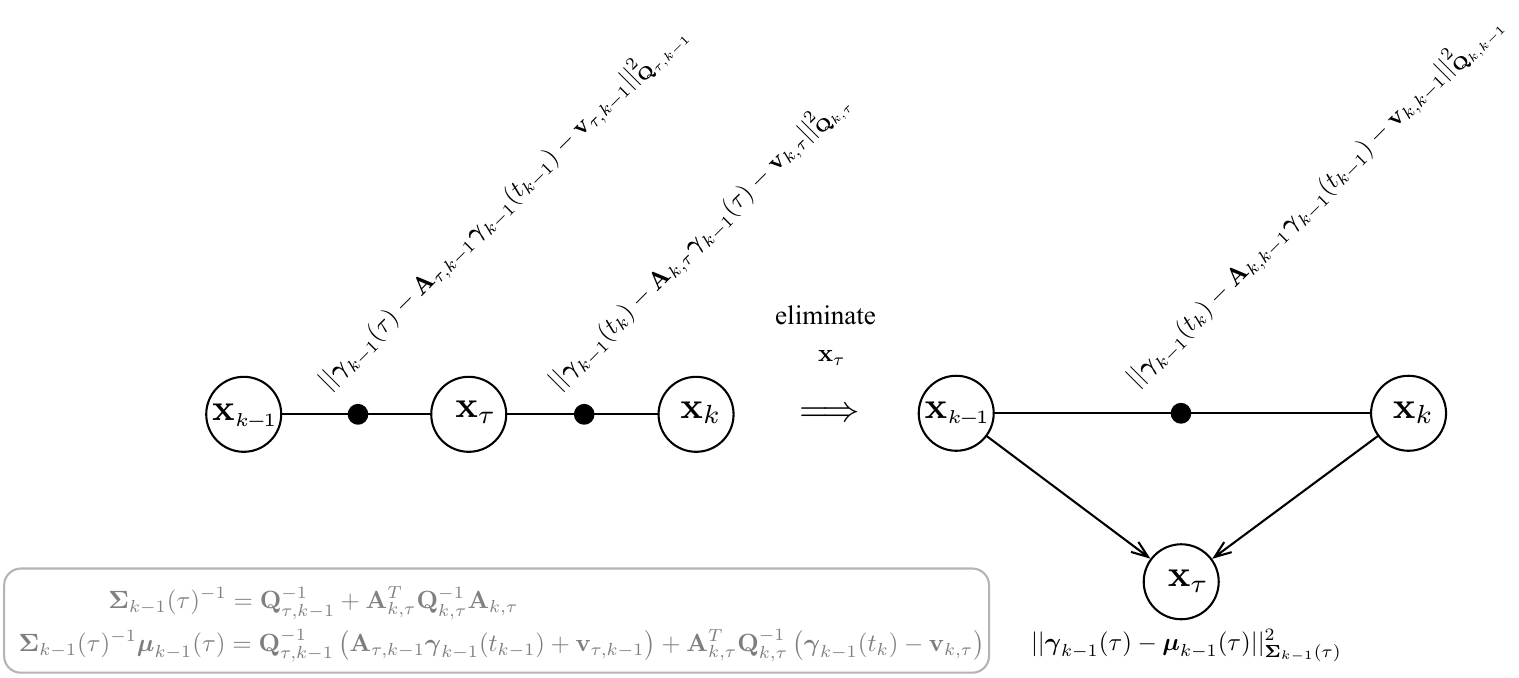}
     \caption{To query the mean of the state at time $t_{k-1} < \tau < t_k$, we start with the situation on the left.  Notice that the factors are expressed in terms of the local \ac{GP} variable, $\mbs{\gamma}_{k-1}$.  We then eliminate $\mbs{\gamma}_{k-1}(\tau)$ resulting in the situation on the right.}
     \label{fig:gausselim4}
\end{figure}

Figure~\ref{fig:gausselim4} illustrates the process for determining the mean at the query time.  We work directly with the local \ac{GP} variables, $\mbs{\gamma}_{k-1}(t_{k-1})$, $\mbs{\gamma}_{k-1}(\tau)$, and $\mbs{\gamma}_{k-1}(t_k)$.  The figure tells us how to compute $\mbs{\mu}_{k-1}(\tau) = (\mbs{\xi}_{k-1}(\tau), \dot{\mbs{\xi}}_{k-1}(\tau), \ldots)$, the mean of the conditional density for the local variable at time $\tau$.  Once we have this, we need to apply these to the global states at the start of the interval to get the estimated mean, $\est{\mbf{x}}_\tau = \left\{ \est{\Tsmall}_\tau, \est{\mbs{\varpi}}_\tau, \ldots \right\}$ at time $\tau$:
\begin{equation}
\est{\Tsmall}_\tau = \Tsmall_{k-1} \mbox{Exp}(\mbs{\xi}_{k-1}(\tau)), \quad \est{\mbs{\varpi}}_\tau = \mbs{\varpi}_{k-1} + \Jbig(\mbs{\xi}_{k-1}(\tau)) \dot{\mbs{\xi}}_{k-1}(\tau), \quad \ldots
\end{equation}
While it is possible to compute the covariance of the state at the query time using the approach outlined in Figure~\ref{fig:gausselim4} as well, this is a bit complicated because we actually want the covariance associated with {\em perturbation} of the Lie group variable not the {\em local \ac{GP} variable}, as this corresponds to the usual {\em Laplace approximation}.

\begin{figure}[t]
     \centering
     \includegraphics[width=0.95\textwidth]{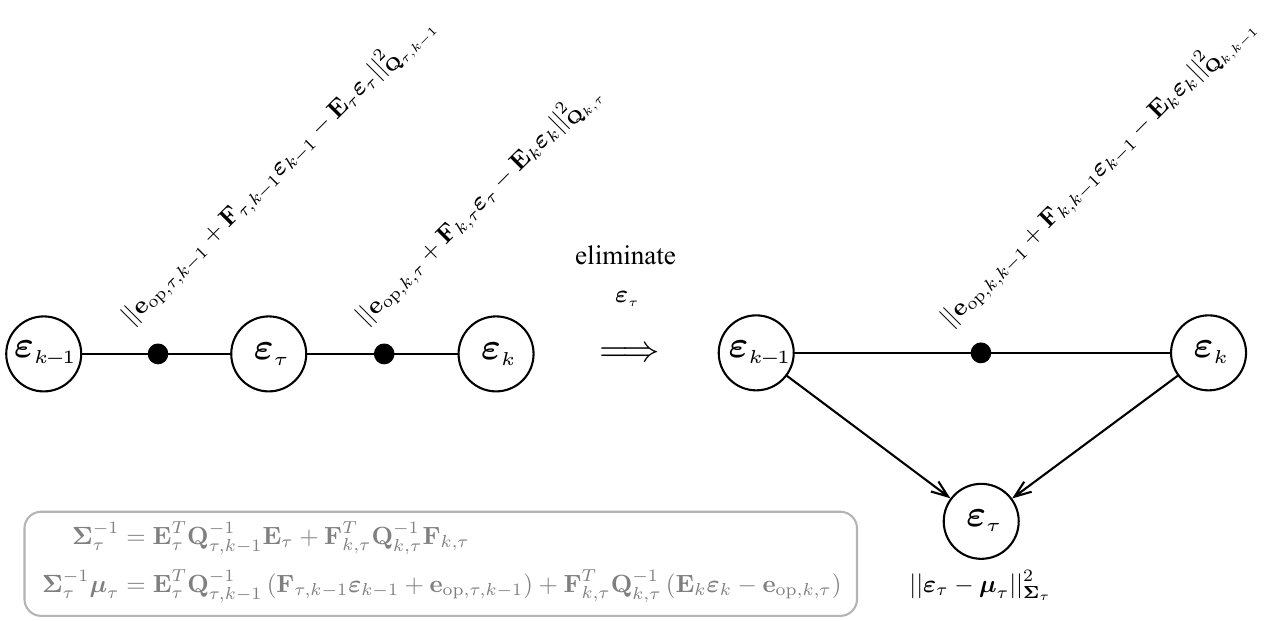}
     \caption{To query the covariance of the state at time $t_{k-1} < \tau < t_k$, we start with the situation on the left.  Notice that the factors are expressed in terms of the perturbation variables.  We then eliminate $\mbs{\varepsilon}_\tau$ resulting in the situation on the right.}
     \label{fig:gausselim5}
\end{figure}

To query the covariance, we instead turn to the perturbation version of the factor graph in Figure~\ref{fig:ctfg3} and then consider the elimination situation in Figure~\ref{fig:gausselim5}, which gives us the mean and covariance of the conditional density for the perturbation variable $\mbs{\varepsilon}_\tau$ at the query time $\tau$:
\begin{equation}
p(\mbs{\varepsilon}_\tau | \mbs{\varepsilon}_{k-1}, \mbs{\varepsilon}_k) = \mathcal{N}(\mbs{\mu}_\tau, \mbs{\Sigma}_\tau).
\end{equation}
Since this is only the conditional density for the perturbation variable, we need to convolve this with the marginal posterior density for the separator states,
\begin{equation}
p(\mbf{x}_{k-1}, \mbf{x}_{k} | \mbf{v}, \mbf{y}) = \mathcal{N}\left(\bbm \est{\mbf{x}}_{k-1} \\ \est{\mbf{x}}_{k} \ebm, \bbm \est{\mbf{P}}_{k-1} & \est{\mbf{P}}_{k,k-1}^T \\ \est{\mbf{P}}_{k,k-1} & \est{\mbf{P}}_{k} \ebm\right),
\end{equation}
as we did in Section~\ref{sec:linafterquery}.  This time we are only interested in computing the covariance as we already have the mean from the previous step:
\begin{equation}
\est{\mbf{P}}_\tau = \mbs{\Sigma}_\tau + \bbm \mbs{\Lambda}_\tau & \mbs{\Psi}_\tau \ebm \bbm \est{\mbf{P}}_{k-1} & \est{\mbf{P}}_{k,k-1}^T \\ \est{\mbf{P}}_{k,k-1} & \est{\mbf{P}}_{k} \ebm \bbm \mbs{\Lambda}_\tau^T \\ \mbs{\Psi}_\tau^T \ebm,
\end{equation}
where
\beqn{}
\mbs{\Lambda}_\tau & = & \mbs{\Sigma}_\tau \mbf{E}_\tau^T \mbf{Q}_{\tau,k-1}^{-1} \mbf{F}_{\tau,k-1}, \\ 
\mbs{\Psi}_\tau & = & \mbs{\Sigma}_\tau \mbf{F}_{k,\tau}^T \mbf{Q}_{k,\tau}^{-1} \mbf{E}_k.
\eeqn
Putting the mean and covariance together, our estimate of the state at the query time $\tau$ is $\mathcal{N}(\est{\mbf{x}}_\tau, \est{\mbf{P}}_\tau)$.

\section{Example}

\begin{figure}[p]
     \centering
     \includegraphics[width=\textwidth]{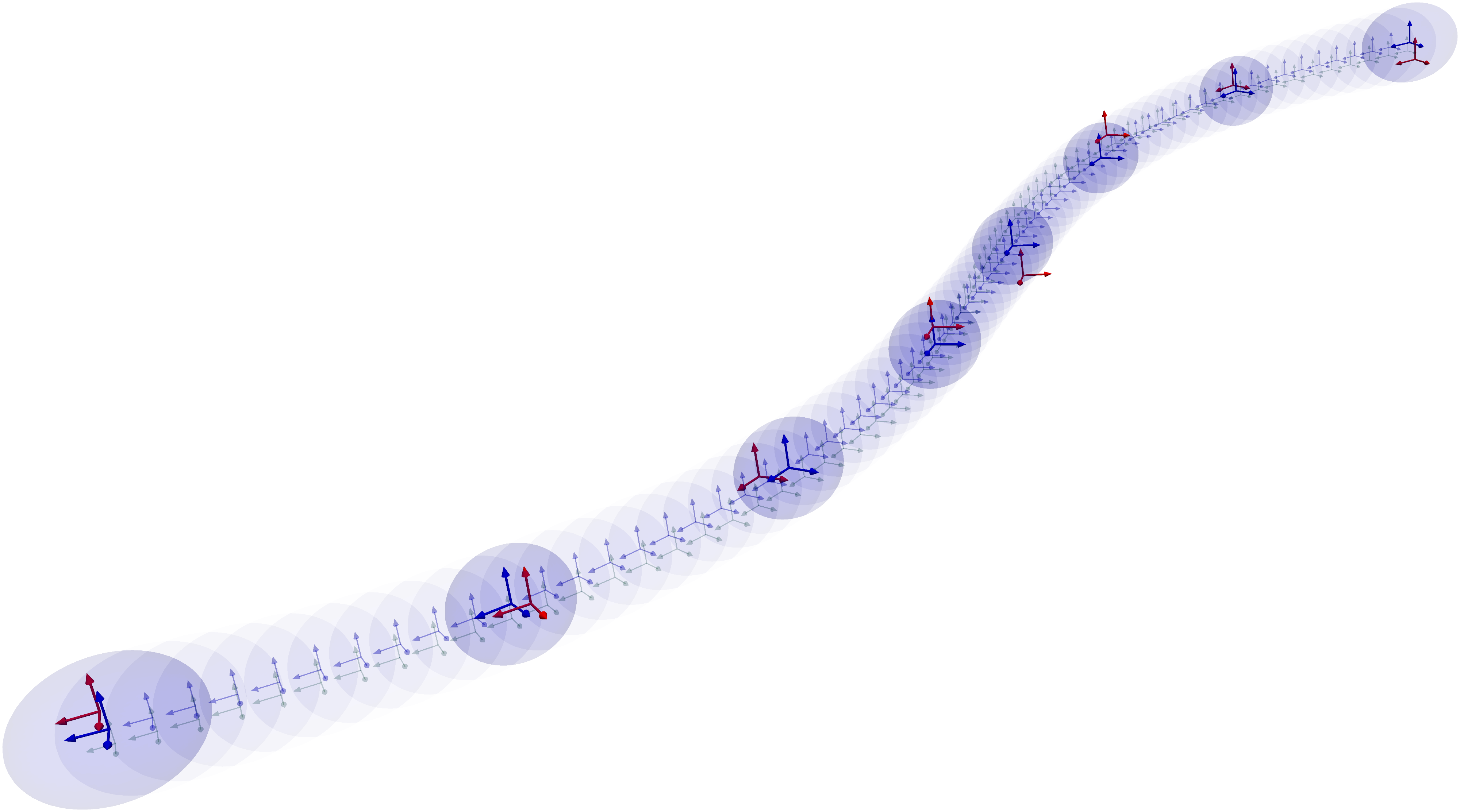}
     \caption{Example of $SE(3)$ state estimation with a \ac{WNOA} motion prior.  The red frames are noisy pose measurements.  The dark blue frames and covariance ellipsoids are computed in the main solve.  The light blue frames and covariance ellipsoids are the interpolated states.  The green frames are the ground truth.}
     \label{fig:se3_trajectory_example}
\end{figure}

\begin{figure}[p]
     \centering
     \includegraphics[width=\textwidth]{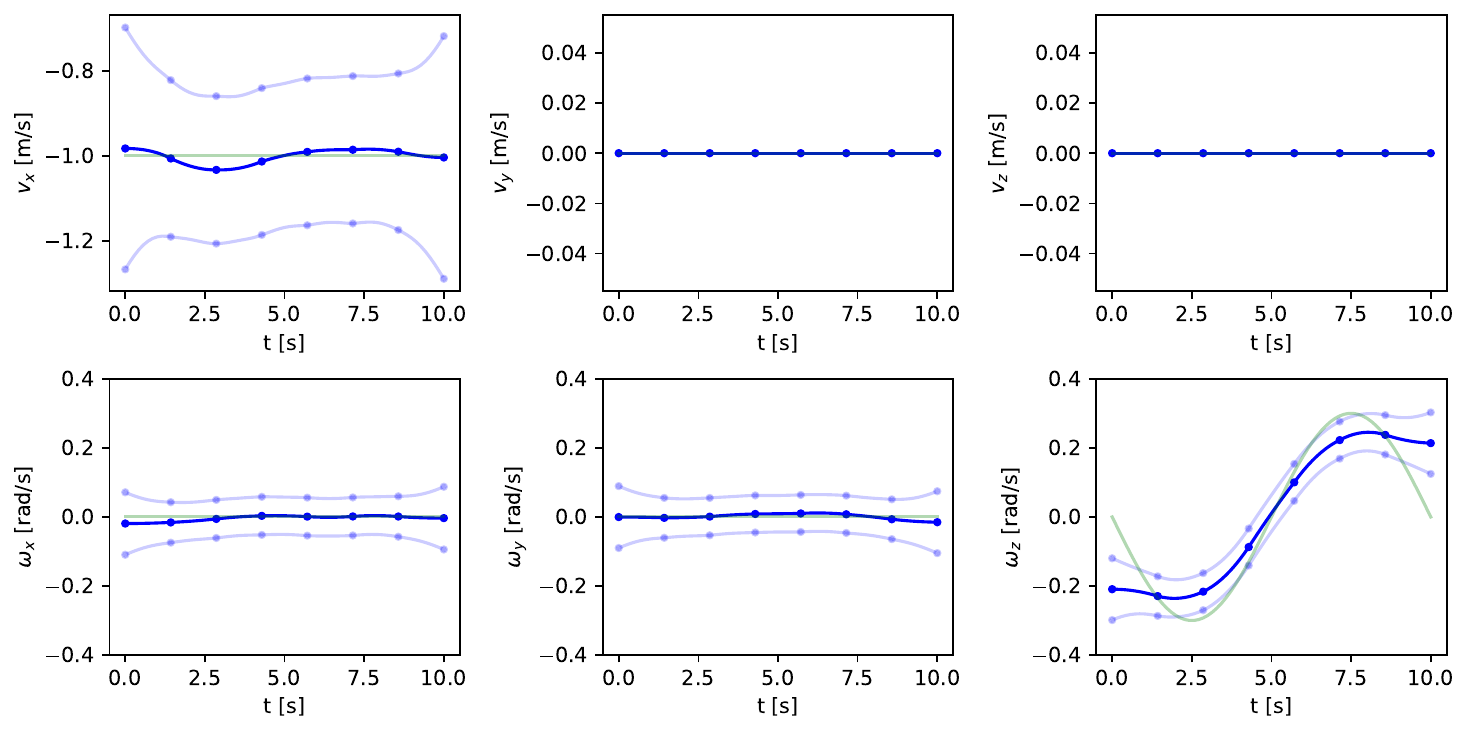}
     \caption{Generalized velocities associated with Figure~\ref{fig:se3_trajectory_example}.  The velocity is estimated at each measurement time and shown as blue dots including covariance. The interpolated velocities and covariances are shown as the blue solid line.  The green line is the ground-truth velocity.}
     \label{fig:se3_velocity_example}
\end{figure}

Figures~\ref{fig:se3_trajectory_example} shows a simple example of estimating a trajectory in $SE(3)$ using a \ac{WNOA} motion prior.  The red frames are noisy pose measurements, the dark blue frame and ellipsoids are the main solve, the light blue frames and ellipsoids are the interpolated states, and the green frames are the ground truth.  The generalized velocities associated with this trajectory are shown in Figure~\ref{fig:se3_velocity_example}, where the velocity is estimated at each measurement time and shown as blue dots including covariance.  The interpolated velocities and covariances are shown as the blue solid line, while the green line is the ground-truth velocity.

In this case, we constrained the lateral velocities to be zero during the solution to produce a very smooth motion representative of a robot moving along a path.  We see that the estimated trajectory is very close to the ground truth, and the estimated velocities are also very close to the ground truth except near the ends of the trajectory where there are fewer measurements.  

%!TEX root =  main.tex

\chapter{Robotics Examples in GTSAM}

% This chapter discusses our open-source implementation of the continuous-time estimation framework as an addition to GTSAM.
\ac{GTSAM} is a widely used C++ library that leverages factor graphs to solve large-scale optimization problems in robotics and computer vision efficiently. One of the key contributions of our work is the introduction of Gaussian-process-based continuous-time (GP-CT) estimation into the existing \ac{GTSAM} library. In this chapter, we will outline the essential functions that can be used to perform GP-CT estimation using our modified library and will provide some explanation of their implementation.

\section{GTSAM Primer}

\ac{GTSAM} provides a wide range of factor implementations that can be combined to construct arbitrary factor graphs.
In practice, users typically rely on existing factors or develop new ones tailored to their specific application.
Each factor is associated with a set of keys, which index the state variables in the factor graph.

A typical \ac{GTSAM} workflow proceeds as follows:

\begin{enumerate}
	\item Define the factors and their associated symbols, indicating to which states each factor relates.
	\item Add the factors to a factor graph object.
	\item Specify initial values for all states involved in the graph.
	\item Pass the factor graph, initial values, and solver parameters to an optimizer (e.g., Gauss-Newton or Levenberg-Marquardt).
	\item Extract the estimated means from the optimizer, and optionally compute the associated covariances.
\end{enumerate}

The implementation of our continuous-time framework within \ac{GTSAM} maintains this workflow, as it mainly introduces new factor types that integrate seamlessly with the existing architecture.
In addition, we provide functionality to compute the mean and covariance of continuous-time states at arbitrary query times, which may differ from the discrete estimation times defined in the primary optimization problem.

\section{Implementation Details}
\label{sec:imp}
In the following, we present implementation details to support the use of \ac{GTSAM} Gaussian-process–based continuous-time estimation capabilities. We illustrate these concepts through a simple example showcasing the Gaussian Process Motion Prior Factor, along with both measurement-state and after-solve interpolation functionalities.

\subsection{Gaussian Process Motion Prior Factor}

As discussed in the preceding chapters, we have shown that, by assuming that the estimated trajectory of a robot is consistent with a given \ac{SDE}, the trajectory can be represented by a \ac{GP}.
In turn, this \ac{GP} can be queried at any given time, both during and after the main optimization routine takes place.

To enforce this relationship in a factor graph setting, we must first define a \emph{motion prior factor} between two robot states that are adjacent in time.
We have implemented such a factor in \ac{GTSAM} for the \ac{WNOA} motion prior, leaving alternative motion models (e.g., \ac{WNOV}, \ac{WNOJ}) as future work. 
 
Recall that the state variable for the \ac{WNOA} prior consists of a both a pose and a velocity variable, where the velocity lives in the Lie algebra of the pose variable.
As seen in Listing~\ref{lst:wnoa}, we have defined a custom C\texttt{++} struct, \texttt{StateData}, to keep track of the variable keys (pose and velocity) and timestamp associated with a particular state.\footnote{This class has specific member functions that facilitate sorting states by time and comparing different states in a well-defined way. This allows efficient sorting for sets and maps, which are essential to some of the underlying operations in our implementation.}

Our implementation of the \ac{WNOA} requires two such state structures as well as the power spectral density matrix of the \ac{WNOA} prior. Alternatively, it can be defined directly on the \ac{GTSAM} keys as long as the time difference between the states is provided.

\begin{lstlisting}[style=C++Style, label={lst:wnoa}, caption={Example setup for WNOA factors.}]
  // Suppose we have two poses and velocities with timestamps.
  Key p0 = Symbol('x', 0);
  Key v0 = Symbol('v', 0);
  double t0 = 0.0;
  // Defined similarly for p1, v1, t1.
  
  // Define states on (p0, v0) and (p1, v1) as timestamped pose-vel pairs.
  StateData state0(p0, v0, t0);
  StateData state1(p1, v1, t1);
  
  // Suppose that p0, p1 are 2D points.
  // Define a Power Spectral Density matrix of appropriate dimensions.
  auto Q_point2 = Vector4::Ones();
  
  // Define a WNOA factor for Point2.
  WnoaMotionFactor<Point2> factor_P2(state0, state1, Q_point2);
  
  // If p0, p1 are SE(3) poses instead, we would define the factor as
  auto Q_se3 = Vector6::Ones();
  WnoaMotionFactor<Pose3> factor_SE3(state0, state1, Q_se3);
\end{lstlisting}

Note that in this example, our factor is templated on a 2D point and an $SE(3)$ pose. In general, \texttt{WnoaMotionFactor}, as well as most novel structures below, support any vector space or Lie group defined in \ac{GTSAM}. 

\subsection{Interpolating States with Measurements}\label{sec:during-query-implem}

We can treat a subset of the variables as interpolated states and use the methodology outlined in Section \ref{sec:interpolation_with_measurements} and Appendix \ref{app:interp-during-solve} to eliminate them from the optimization, while still accounting for the effect of factors that are defined on these states.
As mentioned in the preceding chapters, this allows the practitioner to reduce the number of variables in a given optimization problem, a key reason for using continuous-time estimation.

The independence approximation introduced in Appendix~\ref{app:interp-during-solve} allows us to continue to treat each factor defined on these states separately (rather than lumping them together).
Effectively, each factor defined on an interpolated state, $\mbf{x}_{\tau}$, can be mapped to a new factor on the bordering states, removing the need to actually solve for the interpolated states.

Our implementation allows a user to convert any \ac{GTSAM} factor graph into an equivalent graph where all the interpolated states have been removed from the optimization.
The core element of this implementation is new interpolation factor \texttt{WnoaInterpFactor} that can be wrapped around any \ac{GTSAM} factor defined on one or more interpolated variables.
This `wrapper' factor automatically maps the errors and Jacobians to the bordering states via the \ac{WNOA} interpolation equations and the chain rule.
The resulting factor is defined only with respect to non-interpolated variables.

Mathematically, our wrapper factor implements the conversion from a measurement factor defined on an interpolated state, $p(\mbf{y}_n|\mbf{x}_{\tau_n})$, to a measurement factor defined the bordering states (see~\eqref{eq:wrapped_meas}), but with two key generalizations. First, the development shown in Appendix~\ref{app:interp-during-solve} applies to variables defined in a vector space, but our implementation can be used with variables defined on Lie groups, using the development shown in Chapter~\ref{chapLieGroups} for continuous-time systems on Lie groups. Second, Appendix~\ref{app:interp-during-solve} is demonstrated for unary factors on the interpolated variables, but our wrapper factor implementation can be applied to any kind of \ac{GTSAM} factor. As we show below, our framework can be applied to factors defined on any number of interpolated and non-interpolated variables.

\newpage

\begin{lstlisting}[style=C++Style, label={lst:interp}, caption={Constructing an interpolated factor graph.}]
  // Define (p0, v0, t0), ..., (p4, v4, t4) as in the previous listing.
  StateData state0(p0, v0, t0); // Defined similarly for all other states.
  // Assume all the poses are in SE(3).
  
  // Suppose we have a binary factor between p1 and p3. 
  // Assume relPose_13 is the relative pose measurement between p1 and p3
  const auto binaryNoiseModel = 
    noiseModel::Diagonal::Sigmas(Vector6::Ones());
  const auto binaryFactor = std::make_shared<BetweenFactor<Pose3>>(
      p1, p3, relPose_13, binaryNoiseModel);
  
  // Suppose we also have a prior (unary factor) on p1.
  const auto p1PriorValue = Pose3();  // origin
  const auto unaryNoiseModel = 
    noiseModel::Diagonal::Sigmas(Vector6::Ones());
  const auto unaryFactor = std::make_shared<PriorFactor<Pose3>>(
    p1, p1PriorValue, unaryNoiseModel);
  
  // Suppose all poses are connected via WNOA factors.
  auto Q_se3 = Vector6::Ones();  // Power spectral density
  
  // Now, instead of defining WNOA factors and adding all the states into
  // the graph, we want state1 and state3 to be interpolated.
  
  // Factor graph visualization before interpolation:
  //       unary
  //         |
  //  0 ---- 1 ---- 2 ----- 3 ----- 4
  //  b      i      b       i       b
  //          --- between ---
  
  // Define the set of bordering states that will remain in the main solve.
  set<StateData> borderStates = {state0, state2, state4};
  
  // All other states will be interpolated.
  set<StateData> interpStates = {state1, state3};
  
  // Define the interpolation factors to wrap the two original factors.
  // This also adds the WNOA factors to the states.
  const auto binaryFactorWrapped = WnoaInterpFactor<Pose3>(
    binaryFactor, borderStates, interpStates, Q_se3);
  const auto unaryFactorWrapped = WnoaInterpFactor<Pose3>(
    unaryFactor, borderStates, interpStates, Q_se3);

  // Create graph.
  NonlinearFactorGraph graphInterp;
  graphInterp.add(binaryFactorWrapped);
  graphInterp.add(unaryFactorWrapped);
  

\end{lstlisting}

In Listing~\ref{lst:interp}, we define two factors on the interpolated states for the first and third poses. These factors are then wrapped using our \texttt{WnoaInterpFactor} factor (templated on $SE(3)$) and added to the graph.
The resulting graph is defined only on the bordering states for the zeroth, second and fourth states (poses and velocities).
These factors allow the user to exactly specify which states and factors they wish to convert.
However, we also provide a function that automatically converts an entire factor graph into an equivalent, interpolated graph based on a list of interpolated states. In Listing~\ref{lst:convert}, the function \texttt{interpolateFactorGraph} effectively carries out the operations shown in Figure~\ref{fig:gausselim3b}: it replaces any factors defined on interpolated states with wrapped, interpolated factors and adds \ac{WNOA} motion models factors between adjacent border states. It will also remove any \ac{WNOA} motion model factors defined on interpolated states. The resulting output graph will not include any of the interpolated state variables and can be passed to any standard \ac{GTSAM} optimizer for solving.

\begin{lstlisting}[style=C++Style, label={lst:convert}, caption={Converting to an interpolated factor graph.}, firstnumber=38]
  // Replaces lines 38-48 in Listing 5.2.
  // Suppose we had first constructed the graph with the original factors.
  NonlinearFactorGraph graphOriginal;
  graphOriginal.add(binaryFactor);
  graphOriginal.add(unaryFactor);

  // Convert to an equivalent graph defined with interpolated factors.
  // We use interpolateFactorGraph to automatically construct the
  // necessary WnoaInterpFactors.
  NonlinearFactorGraph graphInterp = interpolateFactorGraph<Pose3>(
    graphOriginal, borderStates, interpStates, Q_se3);
\end{lstlisting}

\subsection{After-Main-Solve Querying}\label{sec:post-query-implem}
Given the interpolated factor graph, we now show optimizing and using after-main-solve querying to extract solutions.
A factor graph built on the factors we have defined herein can be optimized in the same way as any other \ac{GTSAM} factor graph.
However, the result includes only values for the border states.
To recover the means and covariances for the interpolated states, we use after-main-solve interpolation.
As seen in Listing~\ref{lst:optimize}, this can be easily done using \texttt{updateInterpValues}.
This function follows the procedure outlined in Section~\ref{sec:lie-after-query} to compute the interpolated means and covariances.

We must caution the reader regarding the covariances generated by the interpolated graph in two regards. First, as noted in Appendix~\ref{app:interp-during-solve}, the independence assumption treats correlated states as if they are not correlated, leading to overconfident estimates. The degree of this overconfidence depends on how correlated the interpolated states are as well as the number of interpolated factors. Second, covariances that are interpolated using the after-main-solve can sometimes exhibit a variance `bubbling' effect,\footnote{By `bubbling', we refer to swelling of the state covariance (under-confidence) in the interpolated states that is due to the approximations introduced in Appendix~\ref{app:interp-during-solve}. This effect does not appear when this approximation is not made.} especially when estimated states are particularly far apart in time or the power spectral density matrix has large eigenvalues.

\begin{lstlisting}[style=C++Style, label={lst:optimize}, caption={Optimization and after-main-solve interpolation.}, firstnumber=last]
// Continues from Listing 5.3.
// Define some initial values for the optimizer.
Values initialValues = ... 

// Set up and run the optimizer.
GaussNewtonOptimizer opt(graphInterp, initialValues);
Values resultBorder = opt.optimize(); // only contains the border states.

// Fill in the interpolated states using after-solve interpolation.
// Construct an empty covariance map to hold the output
auto covarianceMap = std::make_shared<CovarianceMap>();

// Interpolate
Values resultAll = updateInterpValues<Pose3>(graphInterp, resultBorder,
borderStates, interpStates, Q_se3, covarianceMap);

// The means and covariances for all states can be extracted from
// resultsAll and covarianceMap, respectively.
\end{lstlisting}

\begin{figure}[ht!]
	\centering
	\includegraphics[width=\textwidth]{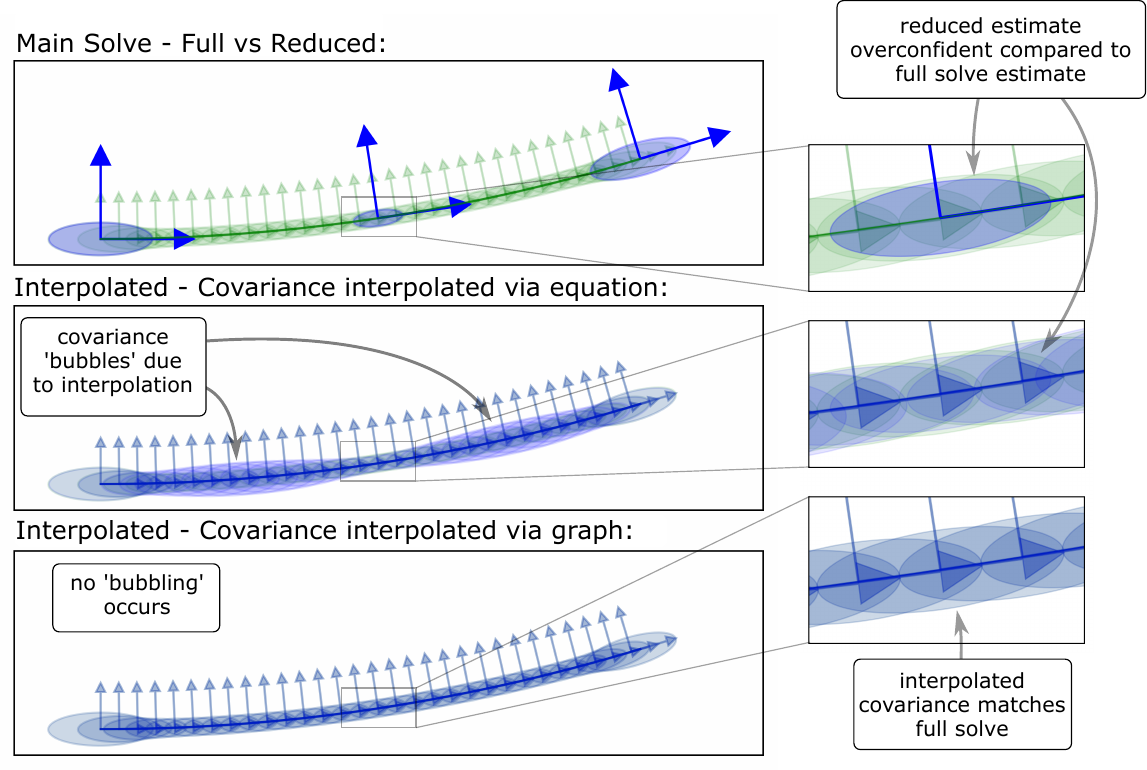}
	\caption{A simple $SE(2)$ interpolation example demonstrating a full factor graph solve including all 60 states (green) compared to a reduced solve where only three of the 60 states are actually solved (blue). A unary factor is placed on each of the 60 states. The top row shows the estimated states in each case (with covariances). The bottom two rows show two different methods of interpolating the covariance of the interpolated states compared with the full solve. The middle row shows covariances interpolated via the after-main-solve equations, while the bottom row shows the covariances interpolated using the original graph with the interpolated means. The latter method provides covariances that match exactly with the full solve, while the former is subject to both overconfidence and `bubbling'.}
	\label{fig:simple-se2-interp}
\end{figure}

Both of these phenomena are demonstrated in top two rows of Figure~\ref{fig:simple-se2-interp}. The figure provides a study of interpolation on a simple $SE(2)$ example. The ground-truth poses follow a constant velocity arc trajectory and no-noise unary measurements were generated at 60 equidistant poses along the trajectory. The figure shows a comparison of the solutions obtained by solving the full factor graph (green) versus interpolating all but three states at the beginning, middle and end. 

The final row of the figure demonstrates an alternative method for finding the covariances at the interpolated states. In this method, we still find the means via the approximations in Appendix~\ref{app:interp-during-solve}, but compute their associated covariances from Laplace's approximation of the original graph about the interpolated means. 
As seen in Listing~\ref{lst:cov-alt}, this can be accomplished quite easily using existing \ac{GTSAM} functions. 

Finally, we emphasize that these confidence issues are not due to the after-main-solve querying itself, but are a consequence of approximations in the main solve.

\newpage
\begin{lstlisting}[style=C++Style, label={lst:cov-alt}, caption={Alternative method of extracting covariances.}, firstnumber=last]
 // Continues from Listing 5.4.
 // Instead of using the covariances in covarianceMap, we can get more
 // consistent covariances using Laplace's approximation.

 // Build marginals from the original graph and the interpolated means.
 Marginals marginals(graphOriginal, resultAll);

 // Extrapolate interpolated covariance for, e.g., Pose 1.
 Matrix cov_p1 = marginals.marginalCovariance(p1);
\end{lstlisting}

\subsection{Speeding up Computation}

The above implementation can introduce computational overhead because each wrapper factor processes interpolation independently.
As a result, factors relying on the same interpolated state repeatedly recompute its mean and the associated Jacobians from the boundary states.
To avoid this redundancy, we introduce the \texttt{WnoaFactorGraph}, a derived class of the \ac{GTSAM} \texttt{NonlinearFactorGraph} that has been modified to tracks all interpolated and estimated states.
During construction, the graph receives a mapping from each interpolated state to its boundary states.
The convenience function, \texttt{interpolateFactorGraph}, has a second template argument that determines the type of factor graph that it returns. By default, the template parameter is set to \texttt{NonlinearFactorGraph}, but it can also be set to \texttt{WnoaFactorGraph} to obtain a factor graph of this form.

The \texttt{WnoaFactorGraph} overrides the \texttt{error} and \texttt{linearize} methods to compute and cache interpolated state means and Jacobians once per iteration, then pass them by reference to all wrapper factors.
Wrapper factors simply retrieve these cached values during their own evaluations.
This eliminates redundant computation and yields substantial runtime improvements, particularly when multiple wrapper factors share the same interpolated states. Moreover, the existing \ac{GTSAM} optimizers do not need to be modified to leverage these overridden methods.
Listing~\ref{lst:interp_wnoa} shows an example usage of this functionality.

We can achieve additional computational speed-up by omitting the computation of the $\mbf{C} \mbf{\Sigma}_{\tau} \mbf{C}^T$ term in the measurement covariance computation of our wrapper factors in  \eqref{eq:wrapped_meas}, since the computation of $\mbf{\Sigma}_{\tau}$ ends up being costly.
We discuss the implications of omitting this terms with respect to accuracy and computational speed later in this section.

Lastly, we note that it may be feasible to approximate the Lie group Jacobians as identity when the relative pose change between consecutive boundary states is small.
This approximation can further reduce computational cost during derivative evaluation, while typically sacrificing only minimal accuracy and preserving overall convergence behavior.

\newpage
\begin{lstlisting}[style=C++Style, label={lst:interp_wnoa}, caption={Constructing an interpolated WNOA factor graph.}]
 // Define (p0, v0, t0), ..., (p4, v4, t4) as in the previous listing.
 StateData state0(p0, v0, t0); // Defined similarly for all other states.
 // Assume all the poses are in SE(3).
	
 // Suppose we have several unary factors (e.g., priors or noisy measuremnts) on p0, p3.
 const Pose3 p0PriorValue( Rot3(), Point3(0.0, 0.0, 0.0) );  
 const Pose3 p3PriorValue1(Rot3(), Point3(1.0, 0.2, 0.0));   
 const Pose3 p3PriorValue2(Rot3(), Point3(1.0, -0.1, 0.0));   
	
 const auto unaryNoiseModel = 
 noiseModel::Diagonal::Sigmas(Vector6::Ones());
 const auto unaryFactor1 = std::make_shared<PriorFactor<Pose3>>(
 p0, p0PriorValue, unaryNoiseModel);
 const auto unaryFactor2 = std::make_shared<PriorFactor<Pose3>>(
 p3, p3PriorValue1, unaryNoiseModel);
 const auto unaryFactor3 = std::make_shared<PriorFactor<Pose3>>(
 p3, p3PriorValue2, unaryNoiseModel);
	
 // In addition, suppose all poses are connected via WNOA factors.
	
 // Factor graph visualization before interpolation:
 // unary                 unary
 //   |                     |
 //   0 ---- 1 ---- 2 ----- 3 ----- 4
 //   b      i      b       i       b
 //                         |         
 //                       unary
	
 // Define bordering and interpolated states
 set<StateData> borderStates = {state0, state2, state4};
 set<StateData> interpStates = {state1, state3};
	
 // Define the interpolation factors to wrap the two original factors.
 auto Q_se3 = Vector6::Ones();  // Power spectral density
 const auto unaryFactorWrapped1 = WnoaInterpFactor<Pose3>(
 unaryFactor2, borderStates, interpStates, Q_se3);
 const auto unaryFactorWrapped2 = WnoaInterpFactor<Pose3>(
 unaryFactor3, borderStates, interpStates, Q_se3);
	
 // Define interpolated to border states mapping
 unordered_map<StateData, pair<StateData, StateData>> interp_to_borders;
 interp_to_borders[state1] = pair(state0, state2);
 interp_to_borders[state3] = pair(state2, state4);

 // Create graph.
 WnoaFactorGraph<Pose3> graphInterp(interp_to_borders, Q_se3);

 graphInterp.add(unaryFactor1);
 graphInterp.add(unaryFactorWrapped1);
 graphInterp.add(unaryFactorWrapped2);

 // Pass to optimizer.
 Values initial = ...
 LevenbergMarquardtParams opt_params;

 Values result = LevenbergMarquardtOptimizer<Pose3>(graphInterp, initial, opt_params).optimize();

\end{lstlisting}

\section{Implementation on Datasets}

We now turn to robotics estimation problems using real-world datasets to demonstrate the capabilities of the continuous-time estimation framework across one-dimensional, two-dimensional, and three-dimensional scenarios.
We present qualitative results for each dataset to highlight its core capabilities, and provide additional quantitative evaluations on the two-dimensional dataset comparing our interpolation scheme. In particular, interpolation is used to reduce the number of estimated states in the factor graph relative to a `full' graph, which also includes the interpolated states.
While high-level descriptions are provided in the following subsections, readers are encouraged to explore the dataset examples themselves, which are available in the open-source repository associated with this paper,\footnote{Our paper and dataset repository can be found at: \url{https://github.com/utiasASRL/2025-fnt-ctfg}.} along with the plotting scripts used to generate the graphs in this section.

\subsection{1D: Giant Glass of Milk Dataset}
\label{sec:exp-1d}
\begin{figure}[ht!]
	\centering
	\includegraphics[width=0.8\textwidth]{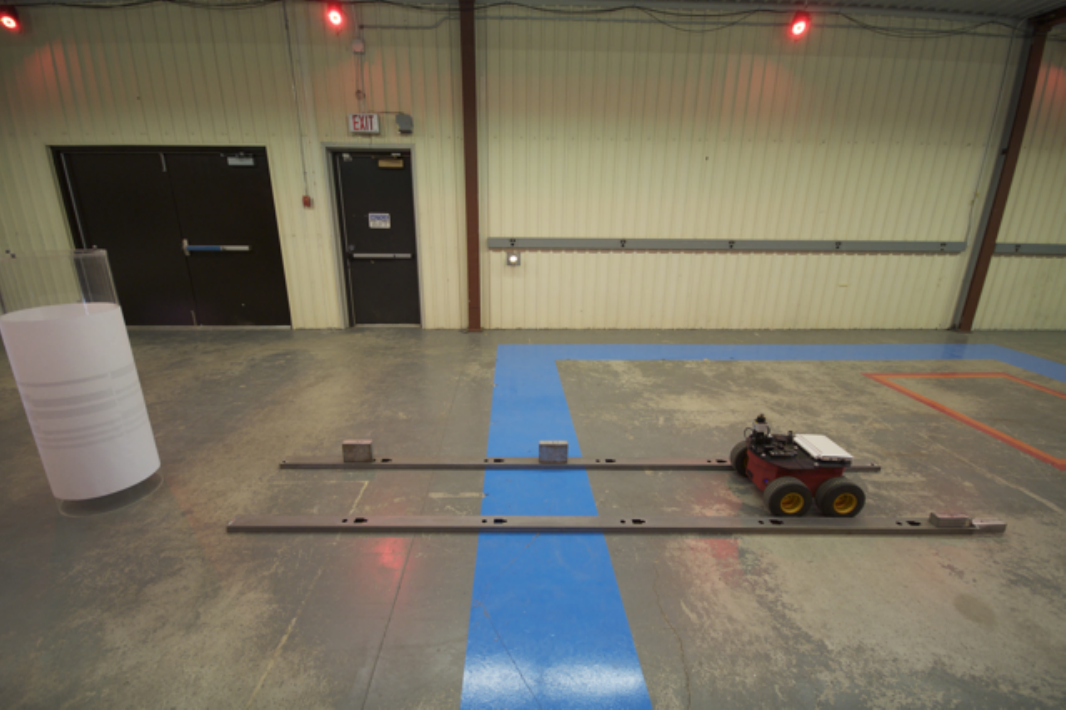}
	\caption{A mobile robot driving between rails. It is equipped with an odometer to measure translational speed and a laser rangefinder to measure the range to the large white cylinder.}
	\label{fig:giant_glass_of_milk}
\end{figure}

In our first example dataset, we investigate a linear one-dimensional problem, in which a mobile robot drives back and forth in a straight line.
The experimental setup is shown in Figure~\ref{fig:giant_glass_of_milk}, consisting of the robot, two fixed guiding rails as well as a large cylinder resembling a \emph{giant glass of milk}.
Both the robot and the cylinder are equipped with reflective markers, which are tracked by a ten-camera motion-capture system, providing ground-truth positions with millimeter-level accuracy.
Additionally, the robot is equipped with a laser rangefinder (Hokuyo URG-04LX) that provides 681 range measurements over a 240° horizontal field of view, centered straight ahead, as well as wheel odometers for measuring the robot's speed.

During data collection, the robot was driven back and forth along the two guiding rails for 20 minutes, logging laser rangefinder and odometer data at 10 Hz, while ground-truth positions were recorded at 70 Hz.
% All measurement data were temporally aligned in a post-processing step using linear interpolation.
Moreover, the distance between the robot and the cylinder was extracted from each rangefinder scan by exploiting the cylinder's circular shape.
This eliminates the need to process the raw laser scans directly, allowing the use of a simplified range measurement instead.
Given the known location of the cylinder, the range measurement can be treated as a noisy observation of the robot’s position.
The noise of each sensor was characterized using the ground-truth data provided by the motion-capture system.

\subsubsection{Qualitative Estimation Results}

\begin{figure}[ht!]
	\centering
	\includegraphics[width=1\textwidth]{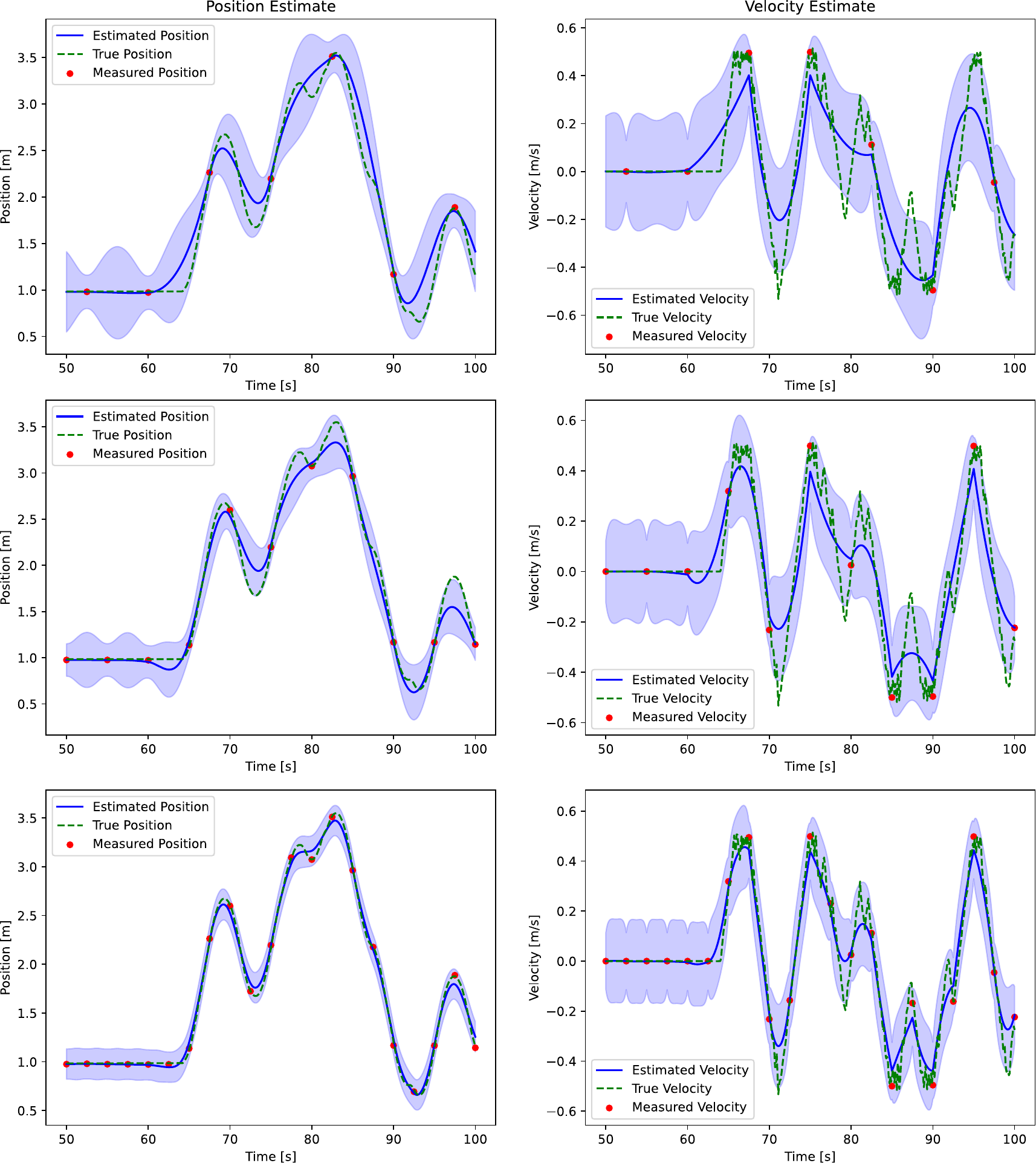}
	\caption{Robot state estimate over a subset of its trajectory for different measurement frequencies. Position and velocity measurements are provided only every 2.5 seconds, 5 seconds or 7.5 seconds, respectively. Between these times, the estimate relies solely on the Gaussian process prior. The plots show the mean estimates with 3-$\sigma$ uncertainty envelopes. Ground-truth velocity is obtained by numerically differentiating the measured position ground-truth.}
	\label{fig:giant_glass_of_milk_results}
\end{figure}

Using our continuous-time state estimation framework, we estimate the robot's position and velocity through its approximately 280~m long trajectory based on noisy position and velocity measurements.
For this, we set our problem up using \ac{GTSAM}, defining estimation states (including position and velocity) every 0.1 seconds. States are sequentially connected by relative \ac{WNOA} motion prior factors and position and velocity measurements are included via unary factors on the states.

We first show that our physics-based, \ac{WNOA} prior factor is effective at smoothing the state estimates between measurement times. 
Fig.~\ref{fig:giant_glass_of_milk_results} shows the robot estimate on a subset of the trajectory with different intervals between available measurements (2.5, 5 and 7.5 seconds).
It can be seen that our approach achieves good estimation accuracy even with relatively sparse data, leveraging the motion prior to produce a continuous estimate over time.
As expected, accuracy improves with a higher number of measurements.
Nonetheless, even with sparse data, the physics-based prior enables close approximations of the ground truth.

\begin{figure}[t]
	\centering
	\includegraphics[width=1\textwidth]{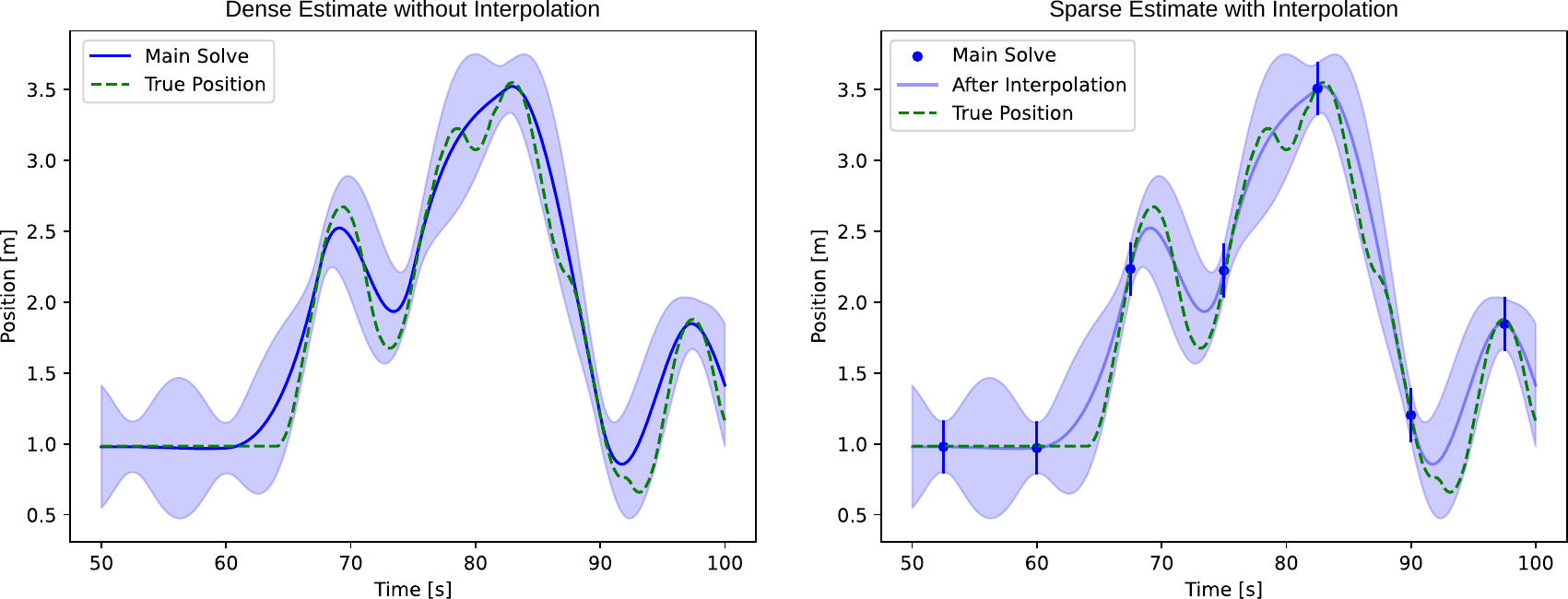}
	\caption{Comparison estimating discrete states in the factor graph at every 0.1 s (left) versus only every 5 s when measurements are available (right). In the latter case, intermediate states are recovered using our post-query interpolation scheme. For this one-dimensional linear example, both approaches yield identical solutions.}
	\label{fig:giant_glass_of_milk_interpolation}
\end{figure}

Second, we evaluate and validate the post-solve interpolation capabilities of our continuous-time state estimation framework.
To do so, we compare the solution quality when discrete states between measurements are interpolated versus explicitly included in the main solve.
As before, the remaining reduced set of states feature both unary position and velocity measurement factors as well as binary \ac{WNOA} motion prior factors.
After solving the resulting state estimation problem, we use the post-solve interpolation implementation detailed in Section \ref{sec:post-query-implem} to recover the estimated mean and covariance for states every 0.1 seconds.
The results of this approach are presented in Fig.~\ref{fig:giant_glass_of_milk_interpolation}, where they are directly compared with the alternative of explicitly solving for the dense states during the main solve.
It can be seen that the results in both cases are identical, which is expected for a linear state estimation problem.

\subsection{2D: Lost in the Woods Dataset}

\begin{figure}[ht!]
	\centering
	\includegraphics[width=\textwidth]{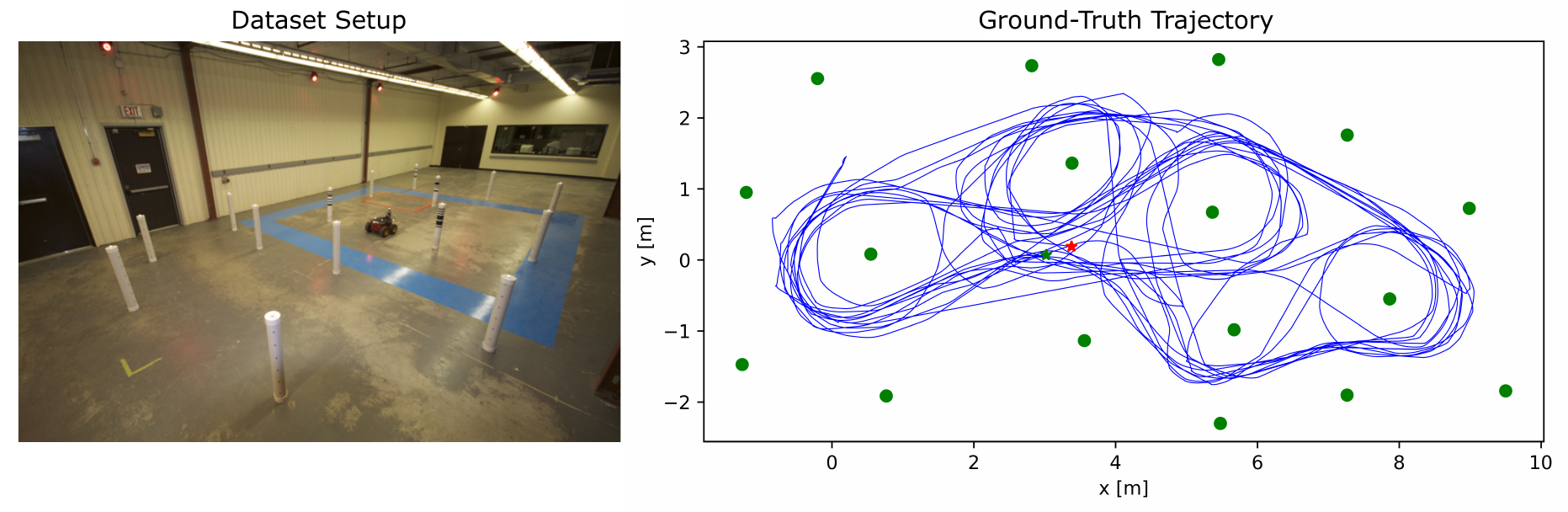}
	\caption{Dataset setup and ground-truth trajectory. The left image shows the setup of the robot and the plastic cylinders that serve as the landmark in a room with 10 camera Vicon system used to obtain the ground-truth positions and orientations of the robot throughout the dataset. The plot on the right shows the full trajectory of the robot throughout the dataset. The start and end positions are marked with a green and red star, respectively.}
	\label{fig:litw-setup}
\end{figure}

Our second dataset consists of a wheeled robot exploring a flat space amongst a forest of fixed plastic tubes that can be used as landmarks (hence the name, ``Lost in the Woods''). 
The robot is equipped with the same laser rangefinder as in the first dataset, but is also outfitted with wheel encoders that measure linear and angular speeds. 
In this case, we are interested estimating the robot's pose in $SE(2)$ and velocity (i.e., localization), but will also estimate the locations of the landmarks (i.e., \ac{SLAM}). 

Sensor data has been sanitized to facilitate the study of back-end estimation routines. In particular, biases have been removed and rangefinder measurements have been converted to bearing-range measurements to landmarks, with known data association. The ground-truth position and orientation of the rangefinder origin on the robot and landmark locations are tracked via a Vicon motion-capture system. All data in the dataset has been synchronized at 10 Hz (approximately). Images of the setup and ground-truth trajectory for the dataset are shown in Figure~\ref{fig:litw-setup}.

\subsubsection{Qualitative Estimation Results}

% First, we show that this estimation problem can be solved using standard discrete-time estimation tools: figure~\ref{fig:litw-gtsam-solve} shows a segment of the trajectory that has been optimized via \ac{GTSAM} with all available bearing-range and wheel odometry measurements. 

% \begin{figure}[ht!]
% 	\centering
% 	\includegraphics[width=0.75\textwidth]{a2_figs/litw_fig2_solve.png}
% 	\caption{\ac{GTSAM} localization solution of a segment (timestep 1000 to timestep 6000) of the Lost in the Woods dataset using both bearing-range and wheel odometry measurements. The ground-truth trajectory is shown in green, while the optimal trajectory is shown in blue. Covariance ellipsoids based on three standard deviations are plotted in blue for the estimated values.}
% 	\label{fig:litw-gtsam-solve}
% \end{figure}
% We now turn our attention to the application of the continuous-time estimation on this dataset. 

\begin{figure}[ht!]
	\centering
	\includegraphics[width=0.75\textwidth]{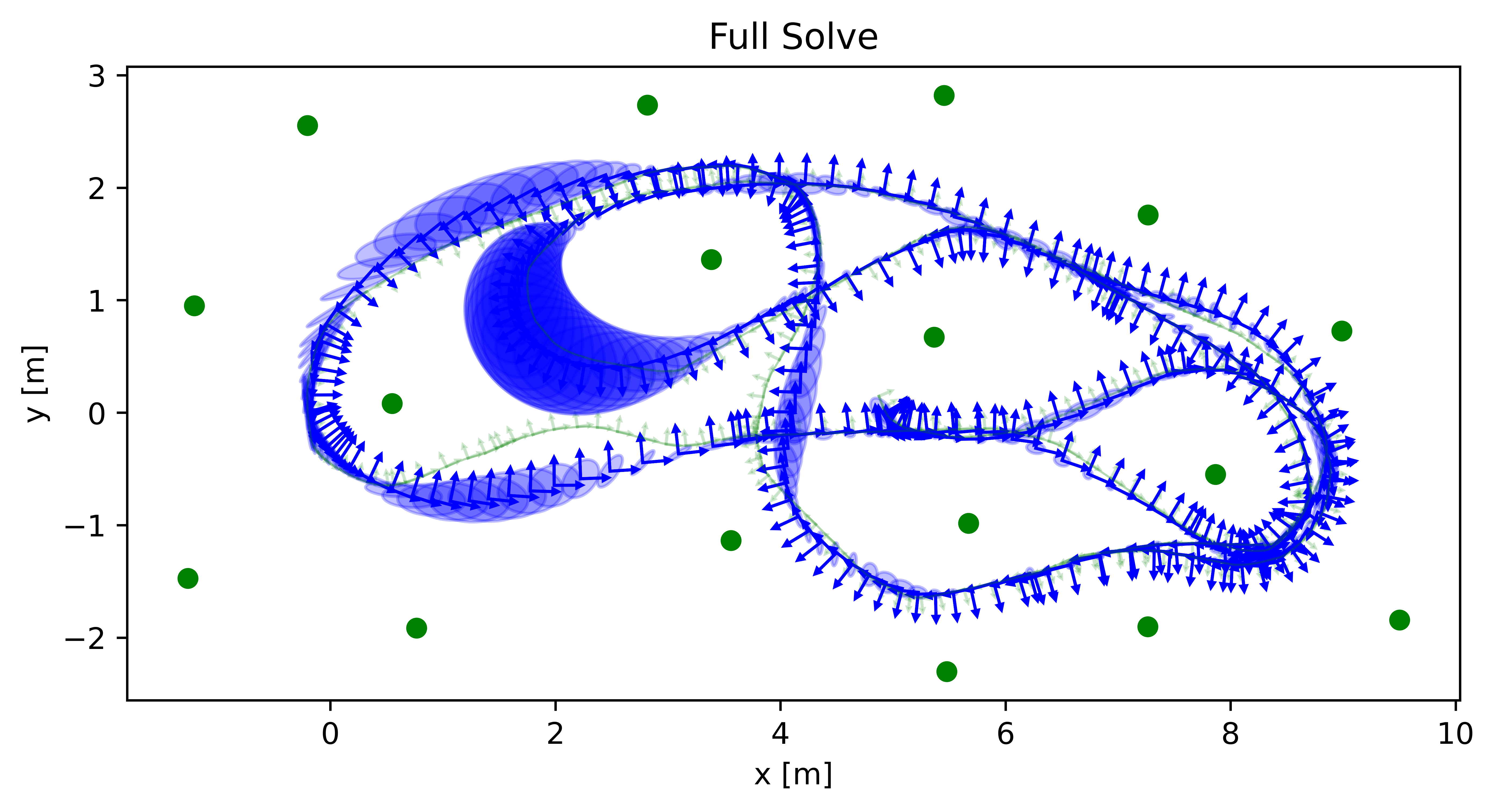}
	\includegraphics[width=0.75\textwidth]{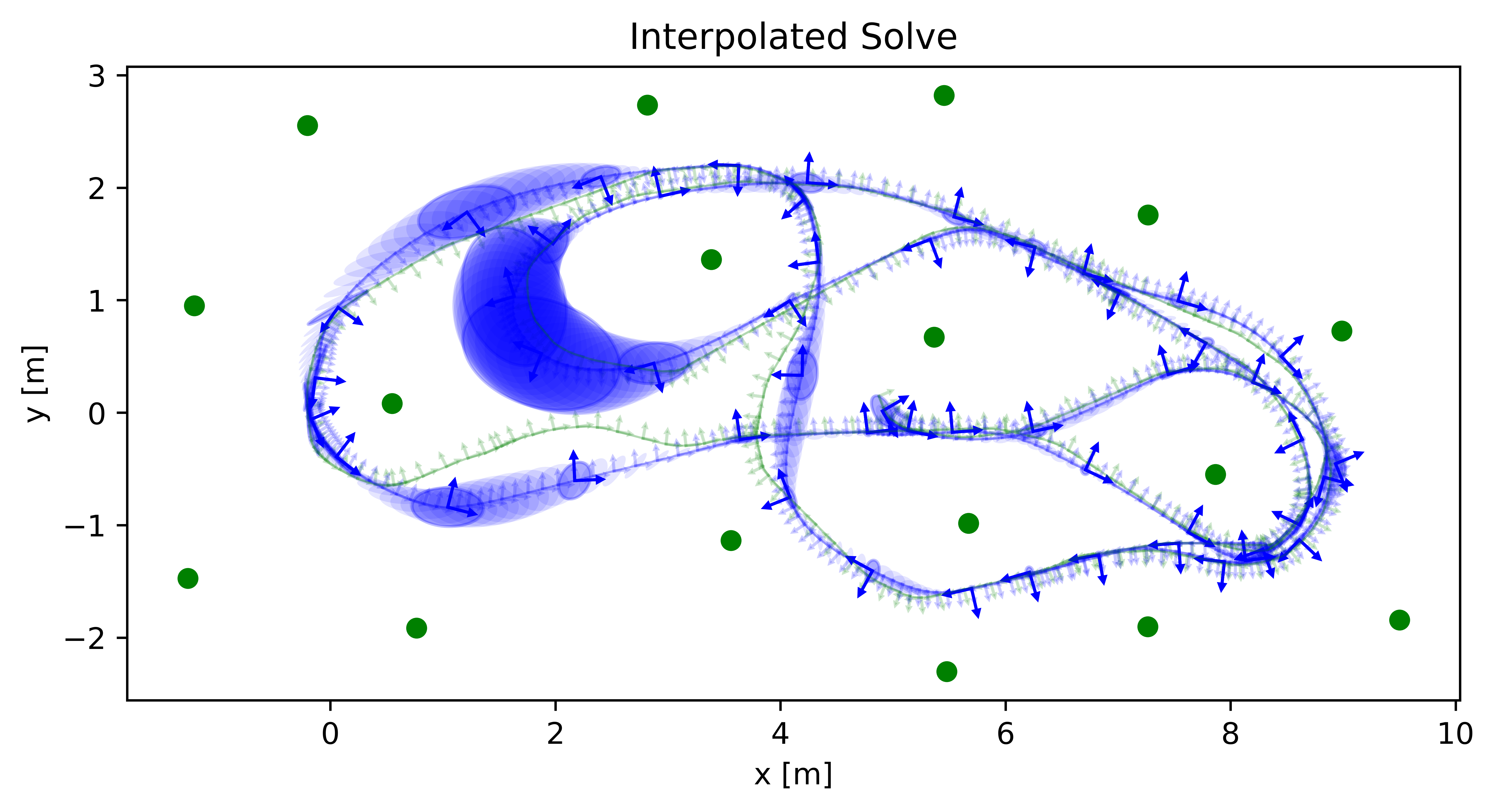}
	\caption{\ac{GTSAM} localization solution of a segment (100 seconds to 250 seconds) of the Lost in the Woods dataset with bearing-range measurements limited to 1 m and \ac{WNOA} motion model factors between adjacent states (no odometry). The ground-truth trajectory is shown in green, while the estimated trajectory is shown in blue. Covariance ellipsoids based on three standard deviations are plotted in blue. In regions with few measurements, estimate uncertainty increases since it mainly relies on the \ac{WNOA} motion prior. In the top plot, states at every measurement time are directly included in the main solve at 10 Hz. In the bottom plot, the main solve only optimizes every $30$th state in the trajectory (i.e., every 3 seconds), with the remainder of the states being interpolated. Non-interpolated states are represented with larger, darker frames and darker ellipsoids with borders. The interpolated solution matches the non-interpolated solution well.}
	\label{fig:litw-loc-wnoa}
\end{figure}

Similar to our first dataset, we first show that applying our \ac{WNOA} motion model is effective even when measurements are sparse. The top of Figure~\ref{fig:litw-loc-wnoa} shows the localized trajectory when wheel odometry and bearing-range measurements with range greater than 1 m are artificially removed. We rely on \ac{WNOA} motion model priors between adjacent states to `fill in the gaps'. The estimated trajectories deviate from the ground truth in regions where there are fewer available bearing-range measurements. In these regions, the estimate relies entirely upon the \ac{WNOA} prior, leading to a corresponding increase in uncertainty. 

Second, we demonstrate the power of the continuous-time scheme to reduce problem size:
Figure~\ref{fig:litw-loc-wnoa} (bottom) shows the same problem, but only one in every 30 trajectory states are actually computed in the main solve, with the remainder being interpolated.\footnote{Note that bearing-range factors on interpolated states were converted to new factors on the remaining states by applying our \texttt{WnoaInterpFactor} (described in Section~\ref{sec:during-query-implem})} We would like to draw the reader's attention to the fact that, despite this aggressive interpolation scheme, we are still able to recover a trajectory that is \emph{almost identical} to the full solve trajectory. The similarity of the covariances also suggests that the confidence issues due to approximations may not always be apparent in practical examples.

\begin{figure}[hb!]
	\centering
	\includegraphics[width=0.75\textwidth]{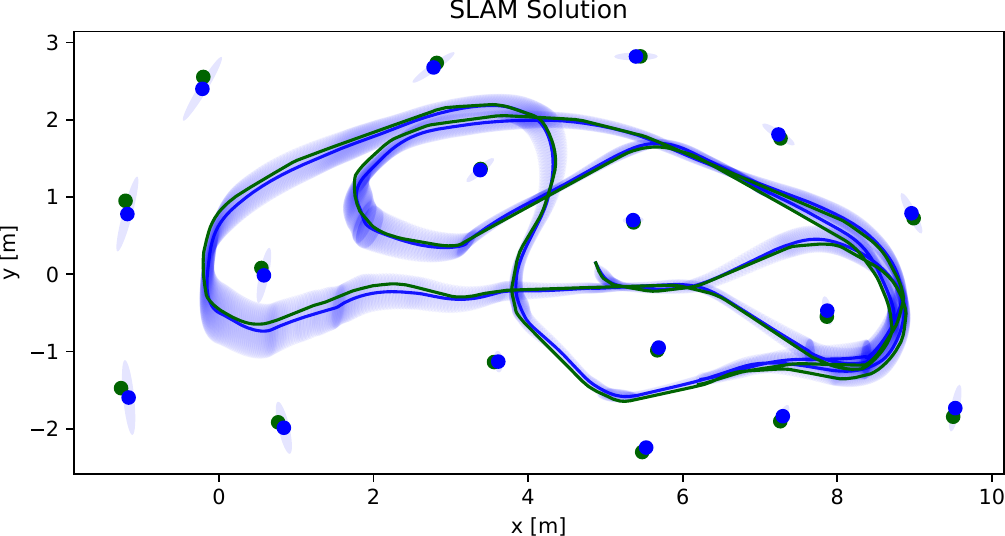}
	\caption{\ac{GTSAM} SLAM solution with interpolation for the robot trajectory between 100 seconds and 250 seconds. The main solve only optimizes every $30$th state in the trajectory (i.e., every 3 seconds), with the remainder of the states being interpolated. Ground truth is shown in green, while the estimates are shown in blue. Covariance ellipsoids are plotted based on three standard deviations. Bearing-range measurements are only included when the range is below 3 m.}
	\label{fig:litw-SLAM-wnoa-interp}
\end{figure}

We also demonstrate the ability of our custom interpolation to handle factors that are defined on a combination of interpolated and non-interpolated states.
In Figure~\ref{fig:litw-SLAM-wnoa-interp}, we solve \ac{SLAM} on the same subset of the trajectory, including the landmarks as to be estimated variables.
As in the previous example, we only keep one of every thirty states in the main solve, interpolating the remaining states and converting factors on interpolated states to \texttt{WnoaInterpFactor} factors.

\subsubsection{Quantitative Results: Runtime-Accuracy Trade-off}\label{sec:runtime-accuracy}

In this section, we use the `Lost In the Woods' localization problem to investigate the trade-offs in compute time and accuracy introduced by continuous-time interpolation. In particular, we show that, though solving the full graph (i.e., explicitly including a state at every measurement time) can be more accurate in some cases, using interpolation can result in considerable gains in compute time while maintaining accuracy. 

Overall, we found that the \texttt{WnoaInterpFactor} slightly increases linearization cost, due to the need to chain derivatives from interpolated states to their bordering states.
However, solving the resulting linearized system can be faster, as it contains significantly fewer variables than the original problem. In practice, users are advised to carefully weigh this trade-off.

\begin{figure}[t!]
	\centering
	\includegraphics[width=0.95\textwidth]{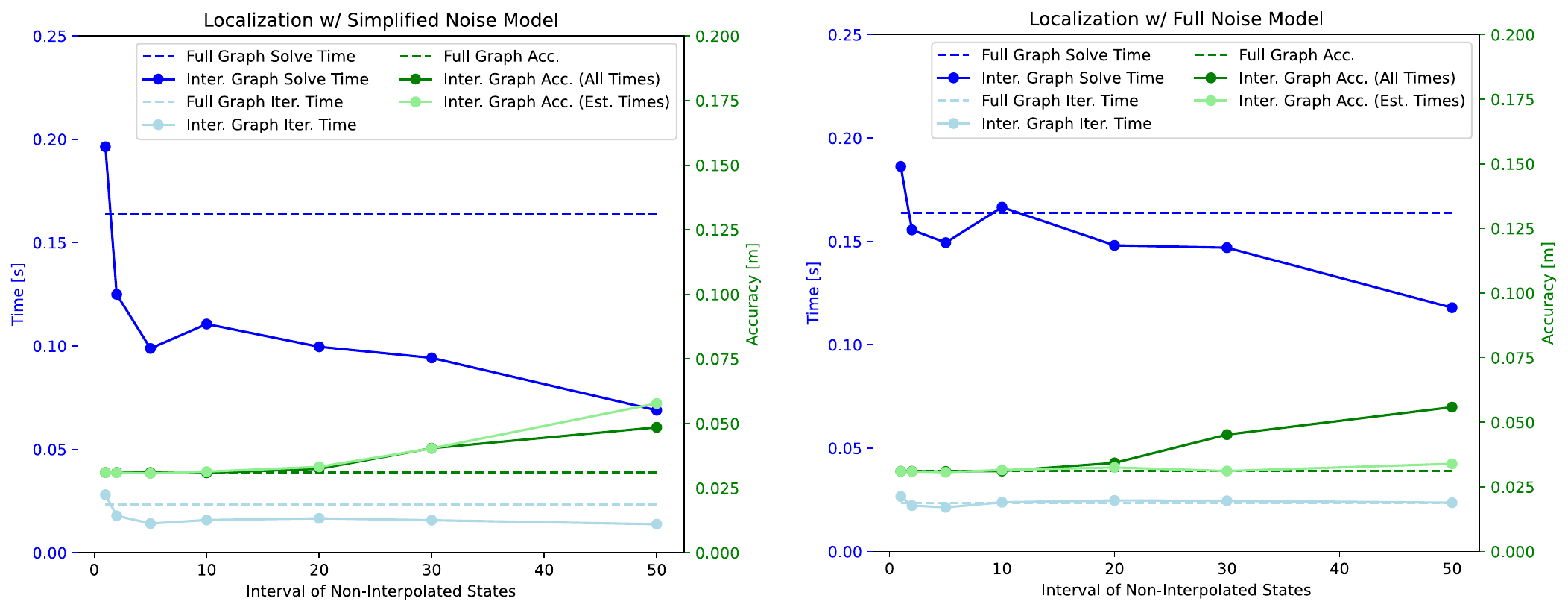}
	\caption{Quantitative comparison of solving the full factor graph by including a state at every measurement time versus relying on interpolation with varying state inclusion frequencies (one state corresponds to 0.1 seconds). We report on accuracy and computation time. We also compare interpolation with and without the approximations applied to the derived noise model for measurements at interpolated times, illustrating the associated trade-offs.}
	\label{fig:litw-quant-results}
\end{figure}

\begin{figure}[t!]
	\centering
	\includegraphics[width=0.55\textwidth]{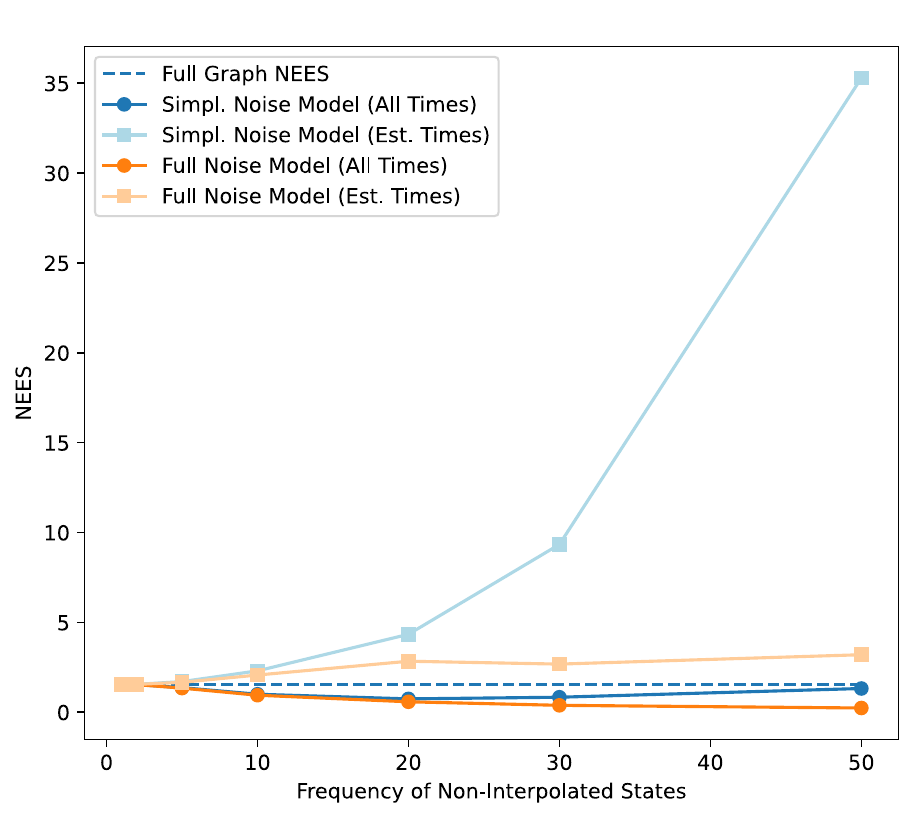}
	\caption{Comparison of the estimate's consistency when solving the problem using interpolation with and without the approximations applied to the derived noise model for measurements at interpolated times.}
	\label{fig:litw-quant-results-nees}
\end{figure}

For this experiment, we again consider a subset of the full trajectory (between 600 and 700 seconds)
and set up the problem to solely rely on bearing-landmark measurements and our \ac{WNOA} motion prior. However, in this section we consider landmarks to be observed when they are within 10 m of a given robot pose. Accuracy and computation time results are shown in Figure~\ref{fig:litw-quant-results}, while Figure~\ref{fig:litw-quant-results-nees} shows the consistency results.

Throughout our experiments, we also observed that an additional approximation of the noise model used for interpolated factors lead to considerable gains in speed. In particular, this simplification omits the $\mathbf{C}\boldsymbol{\Sigma}_\tau\mathbf{C}^\top$ in the covariance of the interpolated factor equation, \eqref{eq:wrapped_factor_simple}.\footnote{Equivalently, we drop the same term when multiple states are interpolated. See Equation~\eqref{eq:wrapped_meas}.} In the following, we refer to the noise model without this term as the ``simplified noise model'', in contrast to the ``full noise model'' when it is included.

Figure~\ref{fig:litw-quant-results} shows that increasing the number of interpolated states generally improves runtime, but (eventually) degrades the accuracy of the overall solve. Here, the $x$-axis indicates the number of interpolated states between each estimated state, while the $y$-axis reports compute time in seconds (full solve and per-iteration) and accuracy in terms of RMSE between the estimated and ground-truth trajectories, reported in meters. When interpolation is used, we also distinguish between the accuracy of states included in the main optimization (i.e., estimated states) and the accuracy of all the states (including those recovered via interpolation), denoted ``Est. Times'' and ``All Times'', respectively.

Comparing the simplified noise model (left) to the full noise model (right) in Figure~\ref{fig:litw-quant-results}, we see that the simplified noise model is much faster, suggesting that the additional covariance term is computationally demanding. Even at interpolation intervals of five states, the runtime is reducing about one third, representing substantial gains.

In terms of overall accuracy, Figure~\ref{fig:litw-quant-results} shows that both the simplified and full noise models maintain performance close to the full-graph solution for interpolation intervals up to roughly 20 states.
We attribute accuracy degradation beyond this point to violations of the constant-velocity assumption underlying the motion model and to the interpolation scheme becoming inadequate to represent the true trajectory.

However, the accuracy of the estimated states degrades more when the simplified model is used. Intuitively, we attribute this effect to the fact that the measurements interpolated states are not properly weighted in the main solve. In contrast, the accuracy of the estimated states with the full noise model remains high even for large interpolation intervals (e.g., 50 states, or 5 s). Interestingly, the accuracy degradation when all states are considered (including interpolated) is roughly equal for both models.\footnote{This is likely because the interpolated states are much less accurate than estimated states in both cases, so they dominate the metric.}

Finally, we observe that the estimates for estimated states become increasingly overconfident when using the simplified noise model. This can be seen in Figure~\ref{fig:litw-quant-results-nees}, where we evaluate the consistency of the obtained estimates using the Normalized Estimation Error Squared (NEES) metric.
This behavior is expected since our simplified model artificially removes uncertainty from measurements.
In contrast, the full noise model maintains consistency levels for the estimated states that remain much closer to those of the full-graph solution.

\subsection{3D: Starry Night Dataset}
\begin{figure}[ht!]
	\centering
	\includegraphics[width=0.75\textwidth]{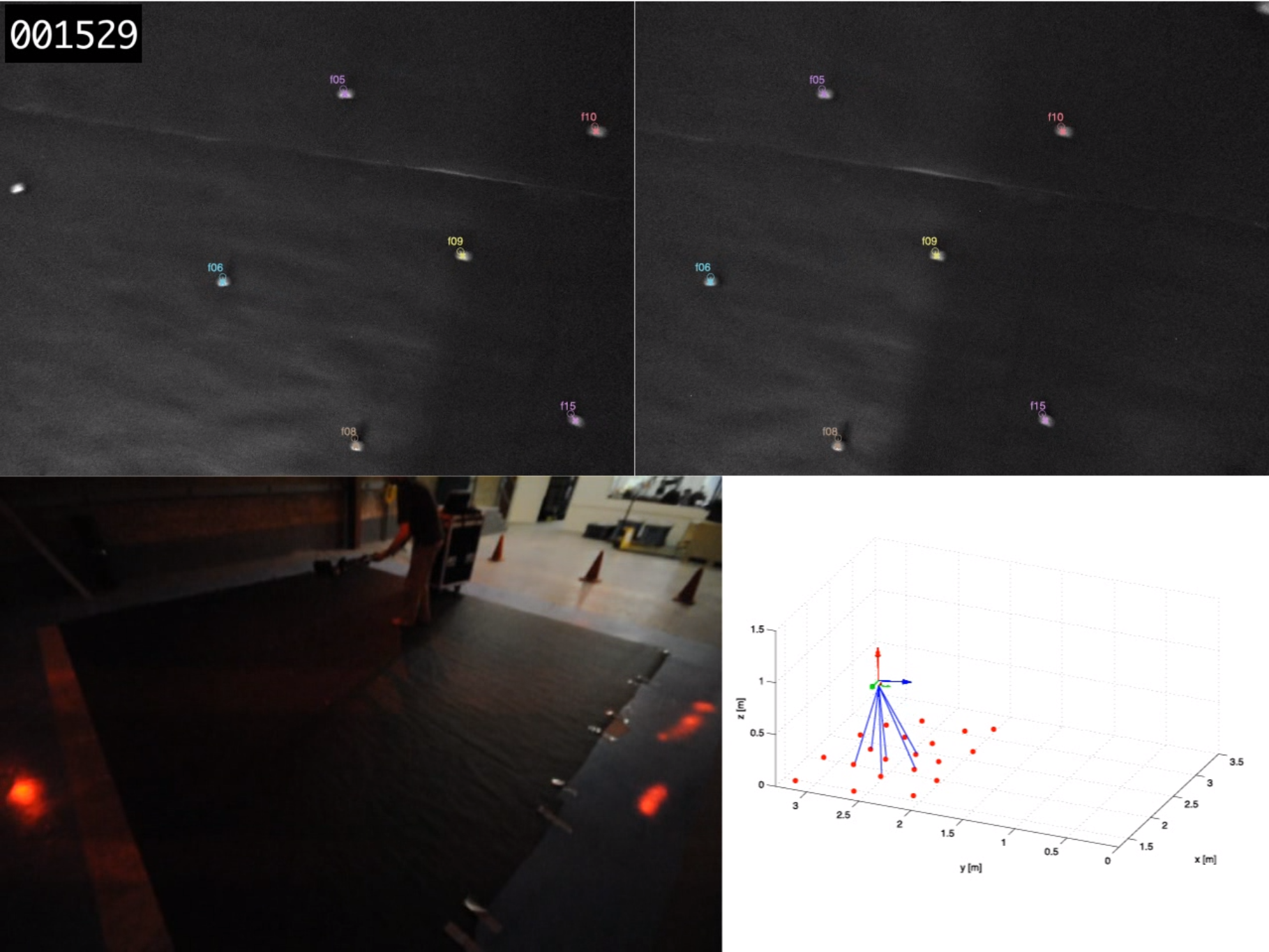}
	\caption{Starry Night dataset collection setup. A rigid sensor head is moved around in dim environment while collecting IMU data and stereo measurements to Vicon markers on the ground (top row). The ground-truth poses of the sensor head are collected using a Vicon motion capture system.}
	\label{fig:a3-setup}
\end{figure}
We now turn to a 3D dataset to demonstrate the above concepts in $SE(3)$. The setup consists of a sensor head that contains an inertial measurement unit (IMU) rigidly attached to a stereo camera. The sensor head is being moved around above a field of reflective markers acting as landmarks that resemble a \emph{``starry night''}. The ground-truth poses of the sensor head are collected using a motion capture system. The localization objective is to estimate the trajectory of the sensor head. See Fig.~\ref{fig:a3-setup} for a visualization of the data collection process.

This dataset has been sanitized to facilitate its use.
For odometry, we have derived the sensor head's (unbiased) linear and angular velocities based on the raw IMU data.
For measurements, we have extracted stereo features from the images, and have associated these features with the landmarks based on ground-truth data.
The odometry and the ground-truth poses have been linearly interpolated to coincide the measurement times.

\subsubsection{Qualitative Estimation Results}

\begin{figure}[ht!]
	\centering
	\includegraphics[width=0.75\textwidth]{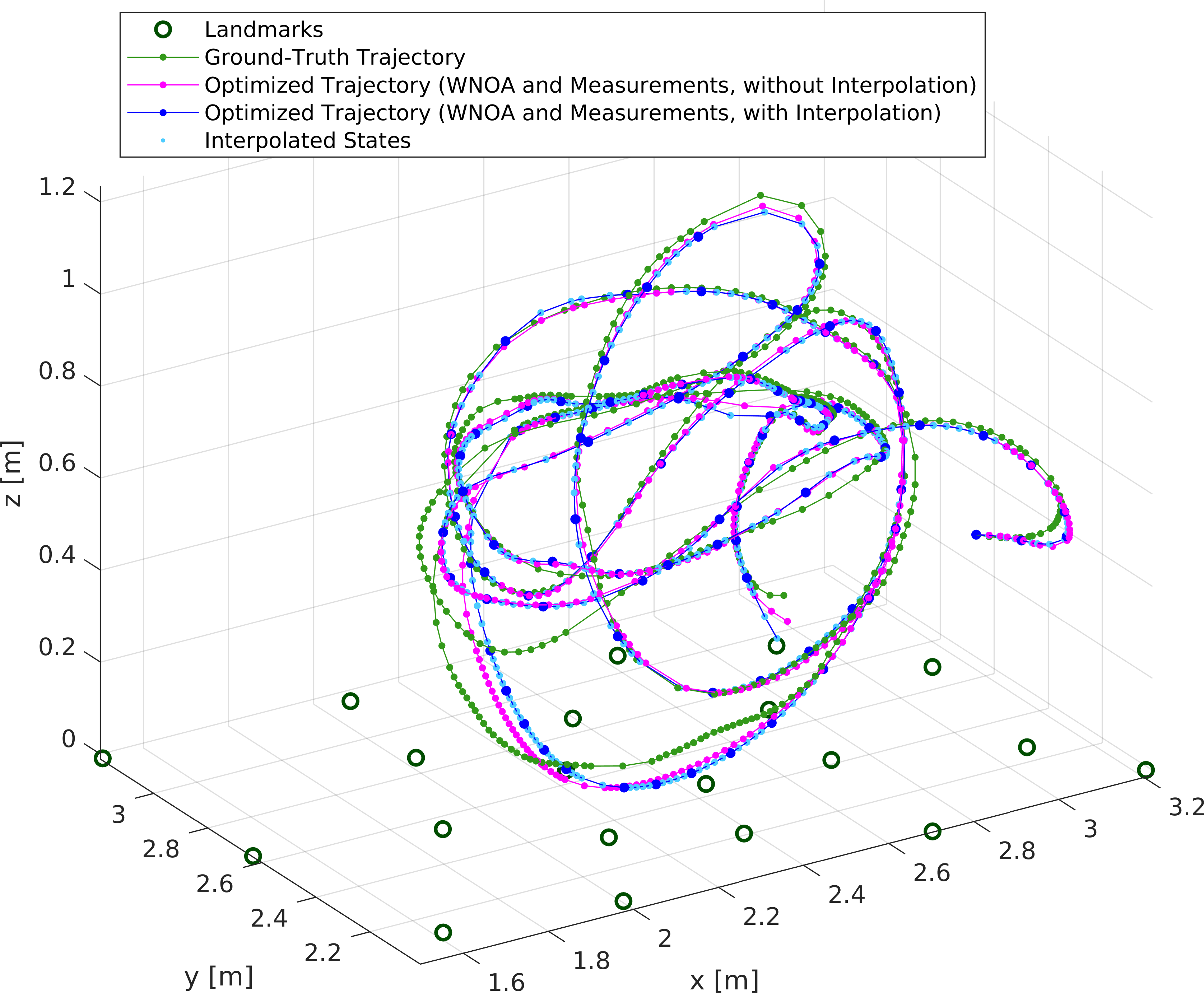}
	\caption{Comparing \ac{GTSAM} localization solutions with and without interpolation between 112 seconds and 152 seconds of the Starry Night dataset. In the magenta trajectory, all states were included in the main solve. In the blue trajectory, every $5^{th}$ state was included in the main solve, with the remainder of the states being interpolated (cyan). We see that the two solutions are fairly close, despite that using interpolation involves far fewer states within the main solve.}
	\label{fig:a3-3d-vis}
\end{figure}

\begin{figure}[ht!]
\centering
\makebox[\textwidth][c]{
 \subcaptionbox*{}{
 \includegraphics[width=0.49\textwidth]{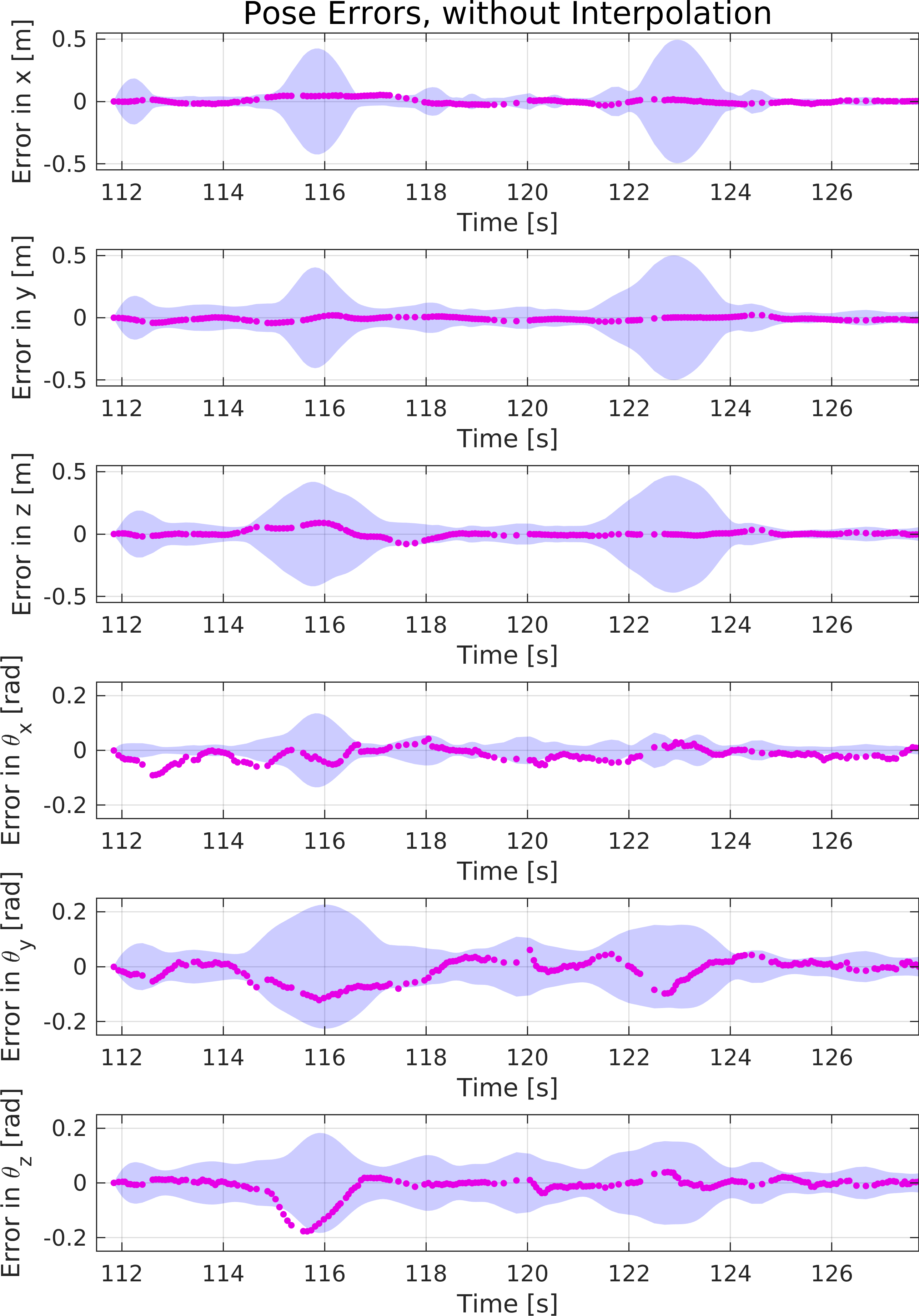}
 }
  \subcaptionbox*{}{
  \includegraphics[width=0.49\textwidth]{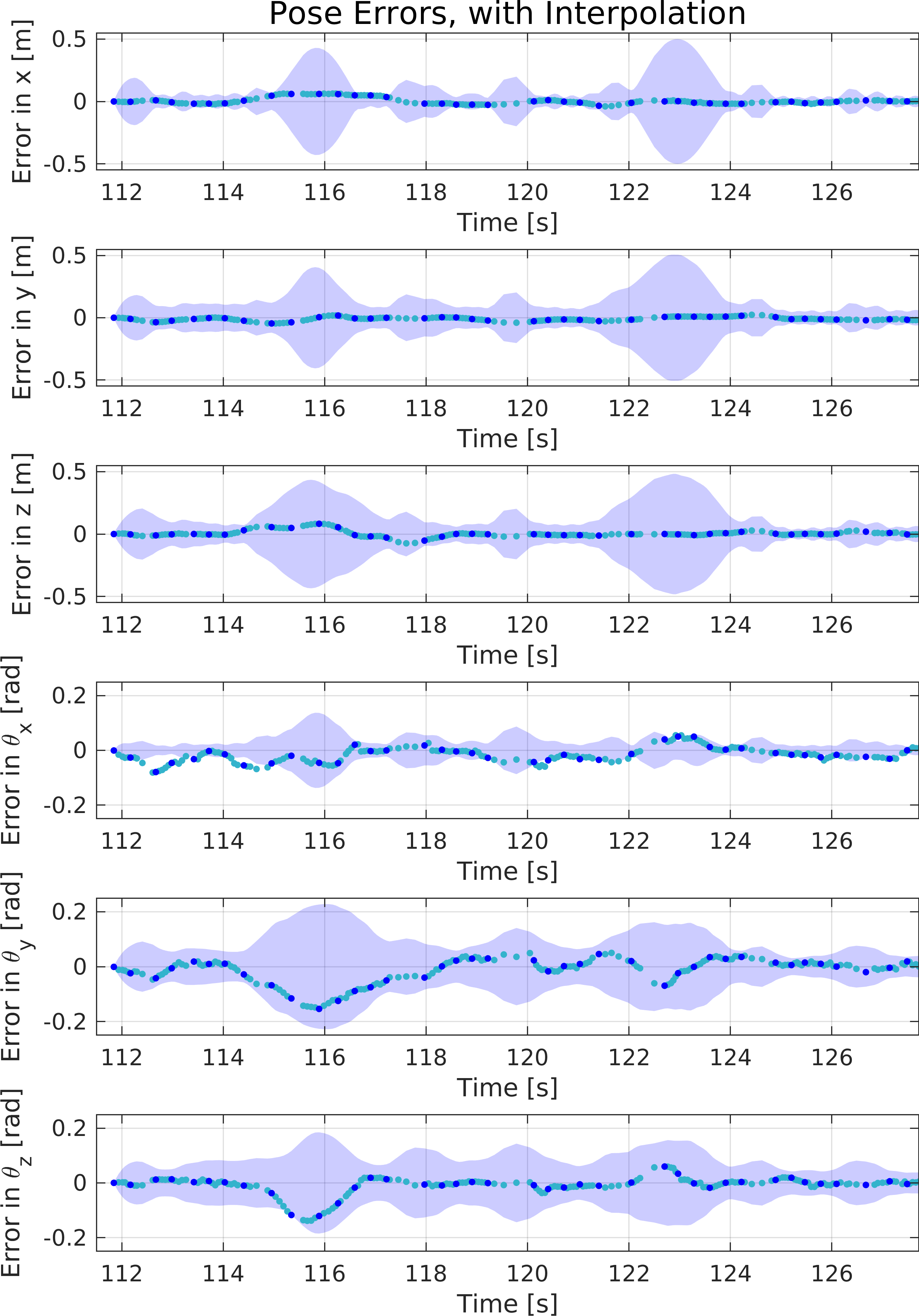}
 }
 }
	\caption{Pose error plots for the \ac{GTSAM} localization solutions in Fig.~\ref{fig:a3-3d-vis} with matching color schemes. The errors and covariances from interpolation (right) are reasonably similar to without interpolation (left), despite having only 20\% of all states included in the main solve.}
	\label{fig:a3-errors}
\end{figure}
We compare the performance of our state estimation approach both with and without interpolation on this 3D dataset in a localization scenario.
We first perform standard localization using stereo measurements and a \ac{WNOA} motion model, including a state at each measurement time during the main solve.
We then perform localization with interpolation, adding only every $5^{th}$ state into the main solve and interpolating the rest following the procedure in Section~\ref{sec:imp}.
We specifically make use of the \texttt{WnoaInterpFactor} described in Section~\ref{sec:during-query-implem}.
For simplicity, covariance interpolation is performed following Section~\ref{sec:lie-after-query}, i.e., not using the alternative method.
The two solutions and their corresponding error plots are shown in Fig.~\ref{fig:a3-3d-vis} and Fig.~\ref{fig:a3-errors}, respectively.

The interpolated solution is fairly close, but not identical, to the full-solve solution.
This is expected since we have made three major approximations: (i) the independence approximation in Section~\ref{sec:ind-approx}; (ii) approximating Lie-group perturbations in Lie algebra (see~\eqref{eq:lie-perturb}); and (iii) approximations from linearizing the model.
Expanding on the last point, since the \ac{WNOA} motion model and the stereo measurement model is nonlinear, the expressiveness of the solution is limited by the number of linearization points. Nevertheless, we see that even with $80\%$ less linearization points, the interpolated solution has a similar accuracy and consistency as the full-solve solution.
We encourage interested readers to try using other interpolation configurations, adding in the \ac{imu} odometry, or try out SLAM on this dataset using the provided code.

\subsection{Key Takeaways and Guidelines}

We have demonstrated practical examples of the proposed continuous-time estimation framework, implemented in \ac{GTSAM}, across several datasets.
In the following, we outline practical guidelines to help readers set up their own estimation problems:

\begin{itemize}
  \item Interpolation of states without measurements requires \ac{WNOA} motion model factors between states, but can be computed in a post-processing step that does not affect the runtime or accuracy of the main solve. 
	\item Interpolation of states with measurements can reduce the number of variables in the linear system solve at each iteration, but increases the time required for linearization of the resulting interpolated factors. Where possible in our implementation, we have used shared memory and parallelization to reduce this linearization cost.
	\item Interpolation can substantially reduce computation while preserving accuracy, provided the interpolation intervals are not too large and the motion generally adheres to the chosen prior.
	\item When sufficient computation is available, solving the full factor graph, i.e., including a state at every measurement time during optimization, remains the best option.
	\item Using the ``simplified noise model'' (see Section~\ref{sec:runtime-accuracy}) for factors on interpolated states reduces computation, but comes at the cost of degraded accuracy of the mean estimate and overconfidence at estimation times compared to the full-graph solution. On the other hand, the ``full noise model'' increases computation time, but maintains higher accuracy at the estimation times and may be beneficial in scenarios where only the accuracy at a specific time (e.g., a single timestamp within a LiDAR scan) matters.
\end{itemize}

%!TEX root =  FnTarticle.tex

\chapter{Conclusion and Outlook}

We have revisited the \ac{GP} approach to continuous-time estimation, through the lens of factor graphs.  We hope the new insights this provides will lead to wider adoption of the approach, as well as new applications.  To this end, we also provided implementations of the approach in \ac{GTSAM} that we hope will be useful to the community.

Looking forward, we see several interesting directions for future work:
\begin{itemize}
\item For computational reasons, we employed a sequence of local \acp{GP} between our estimation times when working on Lie groups.  While this works well in practice, it is not as elegant as the original linear-Gaussian theory, sometimes leading to artifacts in the estimated trajectory.  Perhaps there are other ways to formulate the problem that would allow us to work with a single \ac{GP} over the entire Lie group.  For example, we are currently investigating the use of differential flatness to formulate the problem in a way that would allow us to work with a single \ac{GP} over the entire Lie group along the lines of \citet{johnsonContinuousTimeTrajectoryEstimation2023} who used a similar idea with parametric continuous-time estimation.
\item It could be of interest to explore the use of multi-dimensional \acp{GP} to model more complex systems such as dynamic continuum robots \citep{teetaert_icra24}.  This would require careful consideration of the kernel functions and the structure of the factor graph.
\item Our approach still requires the user to select the estimation times, which is currently done in a heuristic way.  It would be interesting to explore how to automatically select the estimation times  based on the data, perhaps using a data-driven approach or a more principled method based on information theory.  \citet{anderson_icra14}, for example, used a wavelet approach in the parametric setting.
\end{itemize}
Happy estimating!

%BACKMATTER SEE DOCUMENTATION
% references, restarts sample
\appendix
%!TEX root =  main.tex
\chapter{Additional Factor Graph Background}\label{app:add-fg-bg}

In Section~\ref{sec:brief-fg}, we introduced the basic concept of a factor graph. In this appendix, provide additional details about how a factor graph can be used to efficiently solve state estimation problems. Following this, we will work through the specific example of classic discrete-time filtering/smoothing. Although this section can be safely skipped by readers that are already familiar with factor graphs, it sets the stage for Chapter~\ref{chap:fg-ctls}, which shows that continuous-time estimation is completely compatible with this common discrete-time setup, under specific conditions.  For a more thorough introduction to factor graphs, we refer the reader to~\citet{dellaertFactorGraphsRobot2017}.

\section{The Elimination Algorithm}\label{sec:elim-algo}

While it is possible to minimize the cost function in~\eqref{eq:cost} directly using matrix algebra, we here lean into the factor-graph representation and present the general {\em elimination algorithm}, an equivalent way to solve the minimization problem.  The key idea is to eliminate one variable at a time, updating the factors accordingly.  Once we have eliminated all variables, our factor graph has been transformed into another graph called a {\em Bayes net}; there is then a simple procedure to compute the desired posterior density from the Bayes net.

Mathematically, the elimination algorithm starts from a factor graph and ends with a new factorization of the posterior in the form
\begin{equation}\label{eq:elimfactors}
p(\mbf{x}) = p(\mbf{x}_0 | \mbf{s}_0) p(\mbf{x}_1 | \mbf{s}_1) \cdots p(\mbf{x}_K) = \prod_{k=0}^K p(\mbf{x}_k | \mbf{s}_k),
\end{equation}
where $\mbf{s}_k$ is the {\em separator} for $\mbf{x}_k$, defined as the set of variables on which $\mbf{x}_k$ is conditioned after it is eliminated.  
Without loss of generality, here $k=0\ldots K$ is the order in which the variables have been eliminated. 

Once all the conditional densities are computed, we can work backwards from $K$ to $0$ to compute the posterior densities for each variable.  This operation is well-defined because there are can be no (directed) cycles in the resulting graph; a consequence of the fact that each variable's separator cannot include any variables that have already been eliminated.

Each step of the elimination algorithm consists of three operations:  (i) select a variable to eliminate, (ii) multiply all factors involving that variable together to form the {\em product factor}, and (iii) factor the product factor into a {\em conditional density} on the eliminated variable and a {\em residual factor} on the separator variables.  The new factor is then added to the factor graph, and the eliminated variable is removed from the graph.  This process is repeated until all variables have been eliminated.  

\begin{figure}[t]
     \includegraphics[width=\textwidth]{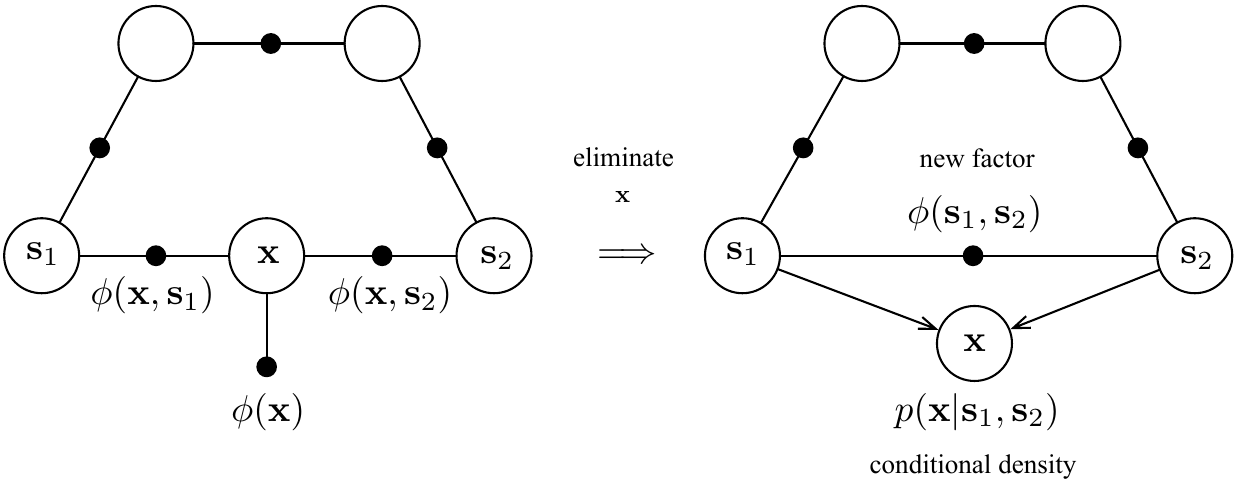}
     \centering
     \caption{Example of the elimination algorithm.  The left shows the factor graph before elimination and the right shows the result of eliminating $\mbf{x}$ from the factor graph, resulting in a new factor on the separator variables and a conditional density for $\mbf{x}$.}
     \label{fig:elim}
\end{figure}

In a bit more detail, consider the example factor graph on the left of Figure~\ref{fig:elim}.  Suppose we select $\mbf{x}$ as the variable to eliminate, then the product factor is given by
\begin{equation}
\psi(\mbf{x}, \mbf{s}) = \phi(\mbf{x}) \phi(\mbf{x}, \mbf{s}_1) \phi(\mbf{x}, \mbf{s}_2),
\end{equation}
where $\mbf{s}_1$ and $\mbf{s}_2$ are two other variables that share factors with $\mbf{x}$; the separator for $\mbf{x}$ is then $\mbf{s} = \{ \mbf{s}_1, \mbf{s}_2 \}$.  The product factor is then factored into a conditional density on $\mbf{x}$ and a new factor on the separator variables:
\begin{equation}
\psi(\mbf{x}, \mbf{s}) = p(\mbf{x} | \mbf{s}) \, \phi(\mbf{s}).
\end{equation}
The key to this last factorization is that $\phi(\mbf{s})$ has no dependence on $\mbf{x}$; it is the factor that would result if $\mbf{x}$ were {\em marginalized} from the factor graph. After all variables have been eliminated, we are left with a graph with only directed arrows called the Bayes net.

The elimination algorithm is quite powerful in that it can be applied to any factor graph.  However, it is particularly useful for problems with a lot of structure, such as those arising in robotics, as we can exploit the sparsity of the factor graph in the various operations.  Here, we will only need to work with factors and densities that are Gaussian, and we use the next section to show that the elimination algorithm can be carried out in closed form when the factors are Gaussian.  This will allow us to carry out estimation efficiently and effectively.

\section{Gaussian Factors, Elimination, Fusion}
\label{sec:fgelimfus}

Now that we understand factor graphs and the general strategy for carrying out estimation, we show that when our factors are Gaussian we can carry out elimination in closed form.  We know that a Gaussian density has the form
\begin{equation}
p(\mbf{x}) \propto \exp\left( -\frac{1}{2} (\mbf{x} - \mbs{\mu})^T \mbs{\Sigma}^{-1} (\mbf{x} - \mbs{\mu}) \right) = \exp\left( -\frac{1}{2} || \mbf{x} - \mbs{\mu} ||^2_\mbs{\Sigma} \right),
\end{equation}
where $\mbs{\mu}$ is the mean value and $\mbs{\Sigma}$ is the covariance.
After taking the negative logarithm, we will only need the $|| \mbf{x} - \mbs{\mu} ||^2_\mbs{\Sigma}$ part, which forms a term in the negative-log likelihood cost that we are seeking to minimize; we will often then label our factors on our factor graphs using this cost term for simplicity.  We may also sometimes write $p(\mbf{x}) \sim \mathcal{N}(\mbs{\mu}, \mbs{\Sigma})$ as a shorthand for Gaussian densities.

In general, we can use the following relationship -- based on the Schur complement -- to carry out elimination when all the factors are Gaussian:
\begin{multline}\label{eq:gausselim}
\underbrace{\left\| \bbm \mbf{x} \\ \mbf{s} \ebm - \bbm \mbs{\mu}_x \\ \mbs{\mu}_s \ebm \right\|^2_{\bbm \mbs{\Sigma}_{xx} & \mbs{\Sigma}_{xs} \\ \mbs{\Sigma}_{sx} & \mbs{\Sigma}_{ss} \ebm} }_{-\log p(\mbf{x}, \mbf{s})} \\ = \underbrace{|| \mbf{x} - \mbs{\mu}_x - \mbs{\Sigma}_{xs} \mbs{\Sigma}_{ss}^{-1} (\mbf{s} - \mbs{\mu}_s) ||^2_{\mbs{\Sigma}_{xx} - \mbs{\Sigma}_{xs} \mbs{\Sigma}_{ss}^{-1} \mbs{\Sigma}_{sx}}}_{-\log p(\mbf{x} | \mbf{s})} + \underbrace{|| \mbf{s} - \mbs{\mu}_s ||^2_{\mbs{\Sigma}_{ss}}}_{-\log p(\mbf{s})}. \\
\end{multline}
Here $\mbf{x}$ is the state to be eliminated and $\mbf{s}$ is the separator state.  The term on the left is the (negative log of the) {\em product factor} in a standardized form.  The second term is the (negative log of the) {\em conditional density} and the third term is the (negative log of the) {\em residual factor} on the separator.  This expression can be used to carry out any Gaussian elimination, including grouped-state elimination. 

\begin{figure}[t]
     \includegraphics[width=\textwidth]{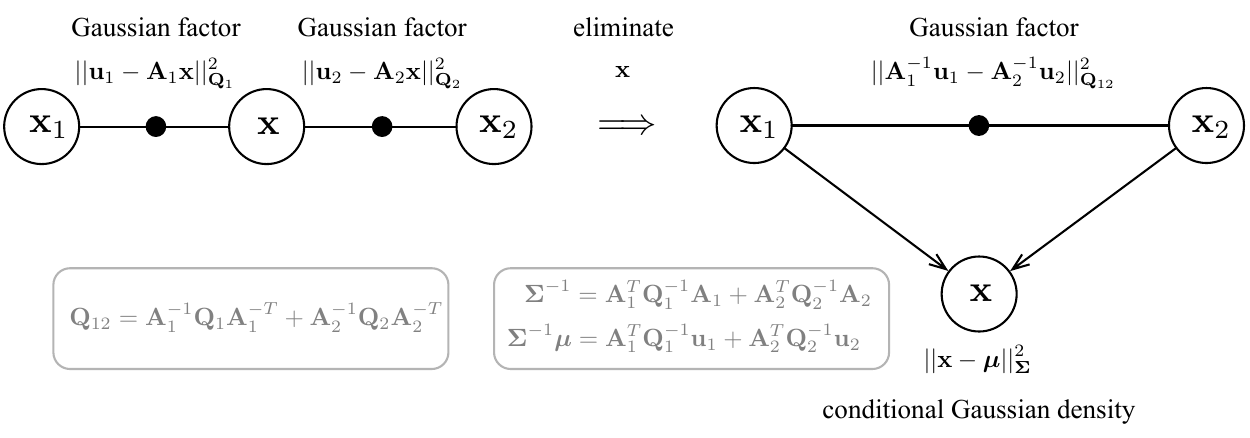}
     \caption{With Gaussian factors, eliminating $\mbf{x}$ results in a new Gaussian factor and a Gaussian conditional density for $\mbf{x}$.  This is a specific example of the general Gaussian elimination formula in~\eqref{eq:gausselim}.}
     \label{fig:gausselim}
\end{figure}

For example, consider the situation on the left of Figure~\ref{fig:gausselim}; we have $\mbf{x}$ linked to two other variables via Gaussian factors (i.e., $\mbf{x}$'s separator).  When we eliminate $\mbf{x}$, this results in the situation on the right, with a new binary factor between the separator variables as well as a Gaussian conditional density for $\mbf{x}$.  Note, we assume that $\mbf{u}_1$ depends on $\mbf{x}_1$ and $\mbf{u}_2$ depends on $\mbf{x}_2$, typically linearly, but the elimination procedure still works if $\mbf{x}$ has a binary factor on one side and a unary factor on the other. 

% The Gaussian elimination in Figure~\ref{fig:gausselim} follows from the very handy sum-of-squares identity,
% \begin{equation}\label{eq:sumofsquares}
% || \mbf{u}_1 - \mbf{A}_1 \mbf{x} ||^2_{\mbf{Q}_1} + || \mbf{u}_2 - \mbf{A}_2 \mbf{x} ||^2_{\mbf{Q}_2}  = || \mbf{A}_1^{-1} \mbf{u}_1 - \mbf{A}_2^{-1} \mbf{u}_2 ||^2_{\mbf{Q}_{12}} + || \mbf{x} - \mbs{\mu}||^2_\mbs{\Sigma},
% \end{equation}
% where $\mbf{Q}_{12}$, $\mbs{\Sigma}$, $\mbs{\mu}$,  are defined in Figure~\ref{fig:gausselim} and $\mbf{A}_1$ and $\mbf{A}_2$ are for now assumed invertible.  We will use this closed-form Gaussian elimination time and again in what follows.  

\begin{figure}[b]
     \centering
     \includegraphics[width=0.8\textwidth]{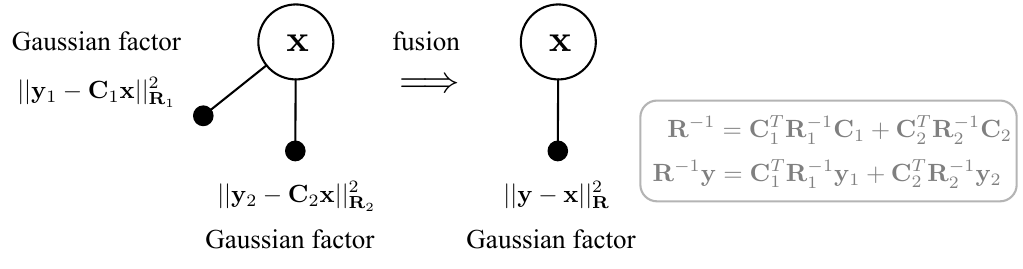}
     \caption{We can fuse multiple Gaussian factors into a single one in order to declutter our factor graph.}
     \label{fig:gaussfusion}
\end{figure}

Another operation that we will perform frequently with Gaussian factors is fusion.  By this we mean lumping together multiple factors into a single one as depicted in Figure~\ref{fig:gaussfusion} for unary factors.  In fact, it is a well-known result that we can fuse an arbitrary number of Gaussians in this way, allowing us to declutter our factor graph.

We are now prepared to discuss classic discrete-time estimation from a factor-graph perspective, which serves as a larger example of the elimination and fusion operations we have just introduced.

\section{Linear Discrete-Time Estimation}
\label{sec:ldte}

Classic discrete-time state estimation has a chain-like factor graph as depicted in Figure~\ref{fig:dtfg1}.  There is a unary initial-condition factor capturing what we know about the state at the initial time, binary motion-model factors predicting how the system moves from one discrete time to the next, and unary measurement factors that tell us something about the state at each discrete time. 

% \begin{figure}[b]
%      \centering
%      \includegraphics[width=0.95\textwidth]{dtfg.pdf}
%      \caption{Factor graph for classic discrete-time filtering/smoothing estimators.}
%      \label{fig:dtfg}
% \end{figure}

\begin{figure}[t]
     \centering
     \includegraphics[width=\textwidth]{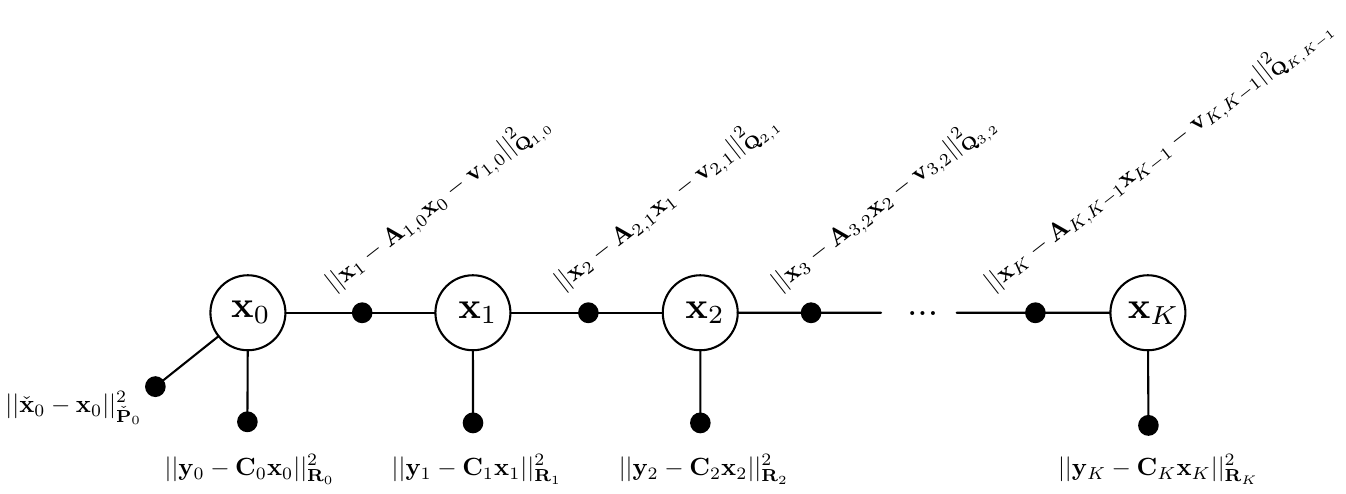}
     \caption{Populated factor graph for the discrete-time estimation of the linear system in~\eqref{eq:linsys}.}
     \label{fig:dtfg2}
\end{figure}

For example, we might have a stochastic linear system of the form
\beqn{linsys}
\mbf{x}_k & = & \mbf{A}_{k,k-1} \mbf{x}_{k-1} + \mbf{v}_{k,k-1} + \mbf{w}_{k,k-1},  \\
\mbf{y}_k & = & \mbf{C}_k \mbf{x}_k + \mbf{n}_k,
\eeqn
where $k \in 0 \ldots K$ is the discrete-time index, $\mbf{x}_k$ is the state, $\mbf{v}_{k,k-1}$ is an input/command, $\mbf{y}_k$ is an output/measurement, $\mbf{w}_{k,k-1} \sim \mathcal{N}(\mbf{0}, \mbf{Q}_{k,k-1})$ is Gaussian process noise, and $\mbf{n}_k \sim \mathcal{N}(\mbf{0}, \mbf{R}_k)$ is Gaussian measurement noise.  We additionally assume that at $k=0$ the initial condition is drawn from a Gaussian, $\mbf{x}_0 \sim \mathcal{N}\left(\pri{\mbf{x}}_0, \pri{\mbf{P}}_0\right)$.  In this case, we can populate our factor graph as in Figure~\ref{fig:dtfg2}, where all of the factors are now Gaussian.  

An important point to make when looking at Figure~\ref{fig:dtfg2} is that the factor graph has a lot of structure in the form of sparsity.  Not all nodes are connected to all other nodes; rather, each node is connected only to its temporal neighbours.  This sparsity arises because $\mbf{x}$ is a {\em Markovian} state for~\eqref{eq:linsys}.  In other words, it captures everything we need to know to predict the next state of the system.  

Our state-estimation problem is now to use the initial-state knowledge, $\pri{\mbf{x}}_0$, the inputs, $\mbf{v}_{k,k-1}$, and the measurements, $\mbf{y}_k$, to work out the likelihood of $\mbf{x}$ taking on all possible values.   This can be done exactly for this linear-Gaussian system; for nonlinear systems we will require some approximations.  We will show that by using a combination of the Gaussian elimination and fusion operations from Section~\ref{sec:fgelimfus}, we can recover the classic \ac{RTS} smoother, the forward pass of which is the \ac{KF}.

Looking to Figure~\ref{fig:dtfg2}, our estimation game plan will be to first fuse the two unary factors on $\mbf{x}_0$, then eliminate $\mbf{x}_0$, then repeat this process with $\mbf{x}_1$ and so on until we arrive at $\mbf{x}_K$.  The result of this forward pass will be a Bayes net.  After this we will carry out a backward pass to work out the posterior densities for $\mbf{x}_K$ all the way back to $\mbf{x}_0$.

\begin{figure}[p]
     \centering
     \includegraphics[width=\textwidth]{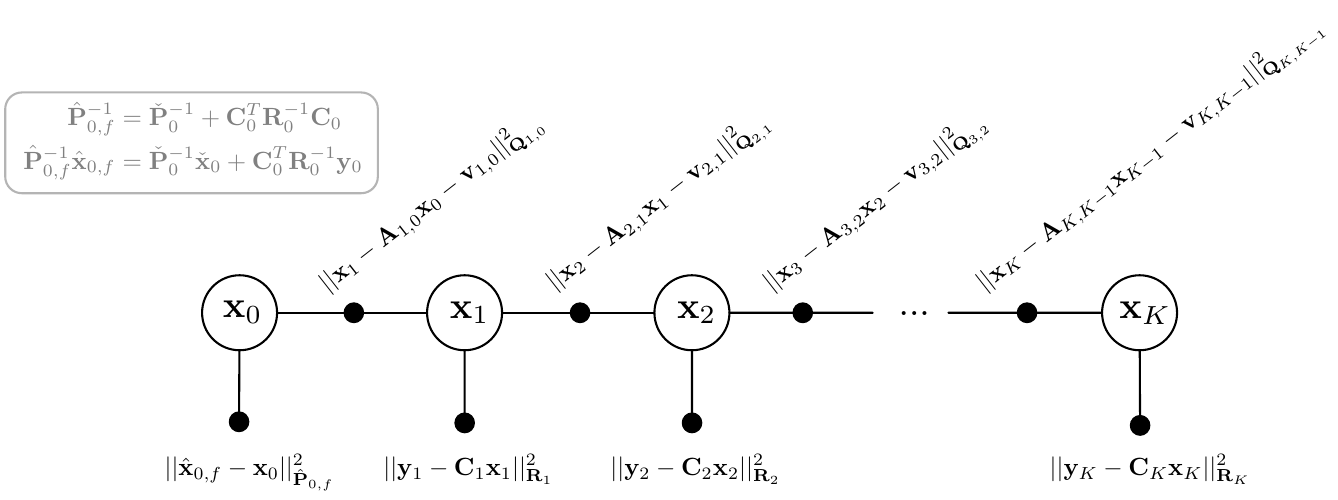}
     \caption{Result after fusing the two unary factors on $\mbf{x}_0$ in Figure~\ref{fig:dtfg2}.}
     \label{fig:dtfg3}
\end{figure}
\begin{figure}[p]
     \centering
     \includegraphics[width=\textwidth]{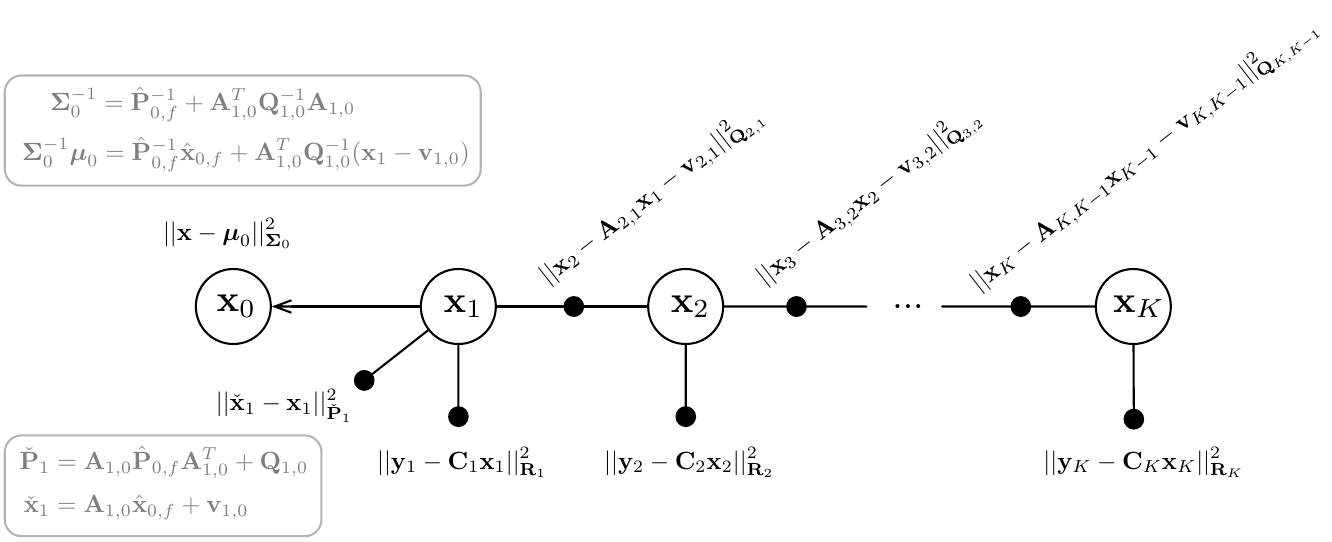}
     \caption{Result after eliminating $\mbf{x}_0$ from the factor graph in Figure~\ref{fig:dtfg3}.}
     \label{fig:dtfg4}
\end{figure}
\begin{figure}[p]
     \centering
     \hspace*{0.5in}\includegraphics[width=0.87\textwidth]{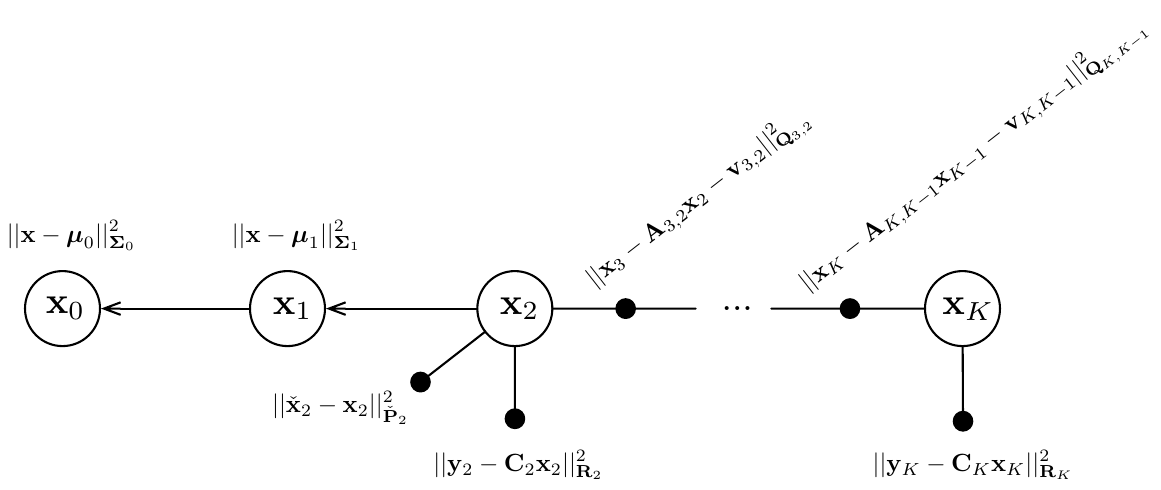}
     \caption{Result after fusing the two unary factors on $\mbf{x}_1$ in Figure~\ref{fig:dtfg4} and then also eliminating $\mbf{x}_1$.}
     \label{fig:dtfg5}
\end{figure}

\begin{figure}[t]
     \centering
     \includegraphics[width=0.85\textwidth]{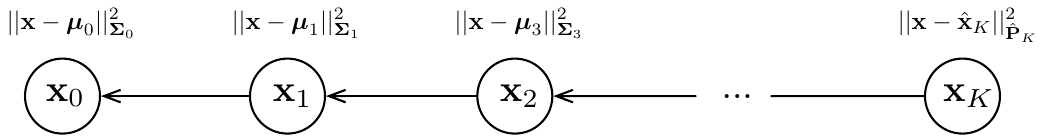}
     \caption{After eliminating all the variables in a forward pass, we are left with a Bayes net.  We can then carry out a backward pass to calculate the posterior densities on all the variables.}
     \label{fig:dtfg6}
\end{figure}

Figure~\ref{fig:dtfg3} shows the result of fusing the two unary factors on $\mbf{x}_0$ into a single Gaussian factor.  Figure~\ref{fig:dtfg4} then shows the result of eliminating $\mbf{x}_0$ from the factor graph, resulting in a new unary factor on $\mbf{x}_1$ as well as the conditional density for $\mbf{x}_0$.  We now notice that the remaining factor graph looks identical to the one we started with, only shifted forward by one time step.  Figure~\ref{fig:dtfg5} shows what happens after another round of fusing the unary factors on $\mbf{x}_1$ then eliminating $\mbf{x}_1$.  If we keep applying this same procedure all the way to time step $K$, our factor graph is completely eliminated and we are left with the Bayes net in Figure~\ref{fig:dtfg6}.  We can then work backward from $\mbf{x}_K$ back to $\mbf{x}_0$ using the conditional densities to compute the posterior densities for each state.

After some algebraic manipulations, it can be shown that this procedure is identical to the classic \ac{RTS} smoother, consisting of the same forward and backward passes:
\begin{subequations}
\label{eq:rts}
\begin{eqnarray}
\mbox{forward:} \quad\;\;\; & & \nonumber \\
\mbox{($k=1\ldots K$)} \; & & \nonumber \\
\pri{\mbf{P}}_{k,f} & = & \mbf{A}_{k,k-1} \est{\mbf{P}}_{k-1,f} \mbf{A}_{k,k-1}^T + \mbf{Q}_{k,k-1}, \\
\pri{\mbf{x}}_{k,f} & = & \mbf{A}_{k,k-1} \est{\mbf{x}}_{k-1,f} + \mbf{v}_{k,k-1}, \\
\mbf{K}_k & = & \pri{\mbf{P}}_{k,f} \mbf{C}_k^T \left( \mbf{C}_k \pri{\mbf{P}}_{k,f} \mbf{C}_k^T + \mbf{R}_k \right)^{-1}, \\
\est{\mbf{P}}_{k,f} & = & \left( \mbf{1} - \mbf{K}_k \mbf{C}_k \right) \pri{\mbf{P}}_{k,f}, \\
\est{\mbf{x}}_{k,f} & = & \pri{\mbf{x}}_{k,f} + \mbf{K}_k \left( \mbf{y}_k - \mbf{C}_k \pri{\mbf{x}}_{k,f} \right), \\
\mbox{backward:}\quad & & \nonumber \\
\mbox{($k=K\ldots 1$)} \; & & \nonumber \\
\est{\mbf{x}}_{k-1} & = & \est{\mbf{x}}_{k-1,f} + \left(\est{\mbf{P}}_{k-1,f} \mbf{A}_{k,k-1}^T \pri{\mbf{P}}_{k,f}^{-1} \right)\left( \est{\mbf{x}}_k - \pri{\mbf{x}}_{k,f} \right), \qquad\quad \\
\est{\mbf{P}}_{k-1} & = & \est{\mbf{P}}_{k-1,f} + \left( \est{\mbf{P}}_{k-1,f} \mbf{A}_{k,k-1}^T \pri{\mbf{P}}_{k,f}^{-1}\right) \left( \est{\mbf{P}}_k - \pri{\mbf{P}}_{k,f} \right) \nonumber \\  &  & \qquad\qquad\qquad\quad \times \; \left( \est{\mbf{P}}_{k-1,f} \mbf{A}_{k,k-1}^T \pri{\mbf{P}}_{k,f}^{-1}\right)^T \label{eq:rtsbackcov3}, 
\end{eqnarray}
\end{subequations}
which are initialized with 
\begin{subequations}
\begin{eqnarray}
\est{\mbf{P}}_{0,f} & = & ( \mbf{1} - \mbf{K}_0 \mbf{C}_0 ) \pri{\mbf{P}}_0, \\
\est{\mbf{x}}_{0,f} & = & \pri{\mbf{x}}_0 + \mbf{K}_0 ( \mbf{y}_0 - \mbf{C}_0 \pri{\mbf{x}}_0), \\
\est{\mbf{x}}_K & = & \est{\mbf{x}}_{K,f}, \\
\est{\mbf{P}}_K & = & \est{\mbf{P}}_{K,f},
\end{eqnarray}
\end{subequations}
and $\mbf{K}_0 = \pri{\mbf{P}}_{0} \mbf{C}_0^T \left( \mbf{C}_0 \pri{\mbf{P}}_{0} \mbf{C}_0^T + \mbf{R}_0 \right)^{-1}$.  We will also be interested in computing the cross-covariance between two consecutive states, which is given by
\begin{equation}
\est{\mbf{P}}_{k,k-1} = \est{\mbf{P}}_{k} \left( \est{\mbf{P}}_{k-1,f} \mbf{A}_{k,k-1}^T \pri{\mbf{P}}_{k,f}^{-1}\right)^T
\end{equation}
and can be calculated during the backward pass.  The computational complexity of the \ac{RTS} smoother is $O(K)$, in other words, linear in the length of the trajectory.

\begin{figure}[ht]
     \centering
     \includegraphics[width=0.95\textwidth]{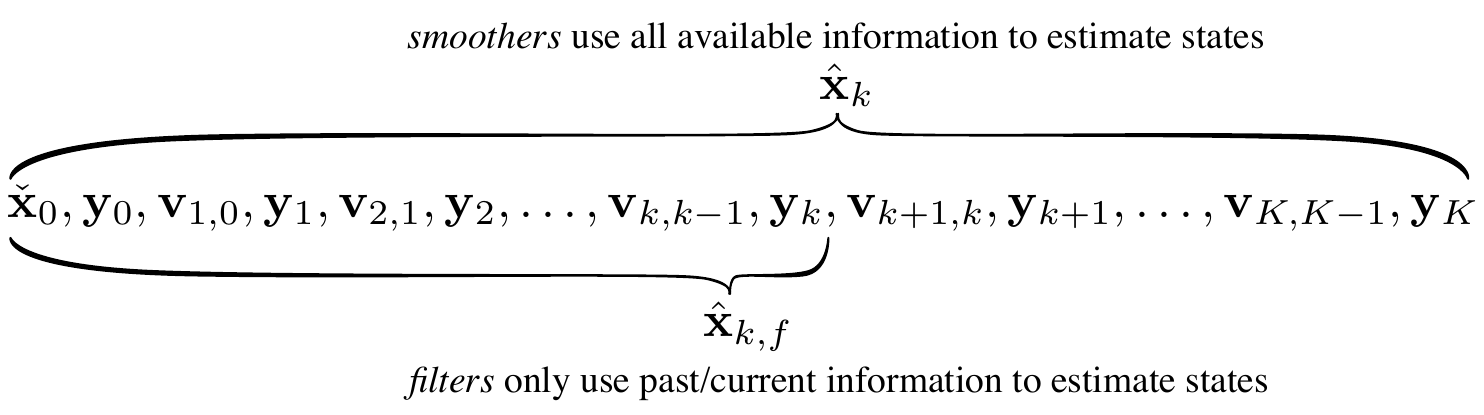}
     \caption{Contrasting filtering vs. smoothing.  We are primarily interested in smoothing in this article.}
     \label{fig:filtersmoother}
\end{figure}

The forward pass is the \acf{KF}, which we now understand from a factor-graph perspective.  The $p(\mbf{x}_k | \mbf{v}_{0:k}, \mbf{y}_{0:k}) = \mathcal{N}\left(\est{\mbf{x}}_{k,f}, \est{\mbf{P}}_{k,f} \right)$ quantities can be referred to as {\em filtered} (accounting only for up-to-the-present measurements) while the $p(\mbf{x}_k | \mbf{v}_{0:K}, \mbf{y}_{0:K}) = \mathcal{N}\left( \est{\mbf{x}}_k, \est{\mbf{P}}_k \right)$ quantities are {\em smoothed} (accounting for all measurements: past, present, and future); see Figure~\ref{fig:filtersmoother}.

It turns out that this discrete-time estimation setup is completely compatible with extending our results to continuous time, we just need to fill in the details of~\eqref{eq:linsys} in the right way.  This is the subject of Chapter~\ref{chap:fg-ctls}.

%!TEX root =  main.tex

\chapter{Interpolating Multiple States with Measurements}\label{app:interp-during-solve}
%Formerly During-the-solve Querying
Discrete-time state estimation often runs into trouble when dealing with measurement data from many unsynchronized, high-rate sensors, \eg, \ac{lidar}, \ac{radar}, \ac{imu} and rolling shutter cameras. The naive approach of defining a state for every measurement may be untenable in practice due to the size of the resulting optimization problem. Methods to aggregate measurement data across time can lead to tractable problems, but also introduce motion distortion errors~\citep{talbot_tro25}. 
On the other hand, continuous-time estimation offers a principled alternative to the approximations that are typically required to aggregate high-rate sensor data. 

In this chapter, we show that continuous-time estimation can be used to include the \emph{effect} of measurements without needing to explicitly define a state variable at their associated measurement times. Instead, we can use \ac{GP} interpolation to reformulate the associated measurement cost terms as a function of a \emph{reduced set of states} that are included in the optimization problem. The size of this reduced set then leads to a clear trade-off between the accuracy and compute time. 

We will see that the factor graph lens continues to bring clarity to the underlying mechanisms of this approach, as well as shedding new light on approximations that are commonly used in practice. However, in contrast to the previous chapter, we now consider the case where interpolated states \emph{do} have measurements associated with them.

In the remainder of this chapter, we show that the elimination operations and some suitable approximations that will lead to a practical approach to dealing with asynchronous, high-frequency sensor data.

\section{Grouped Interpolation}

\begin{figure}[t]
     \centering
     \includegraphics[width=0.90\textwidth]{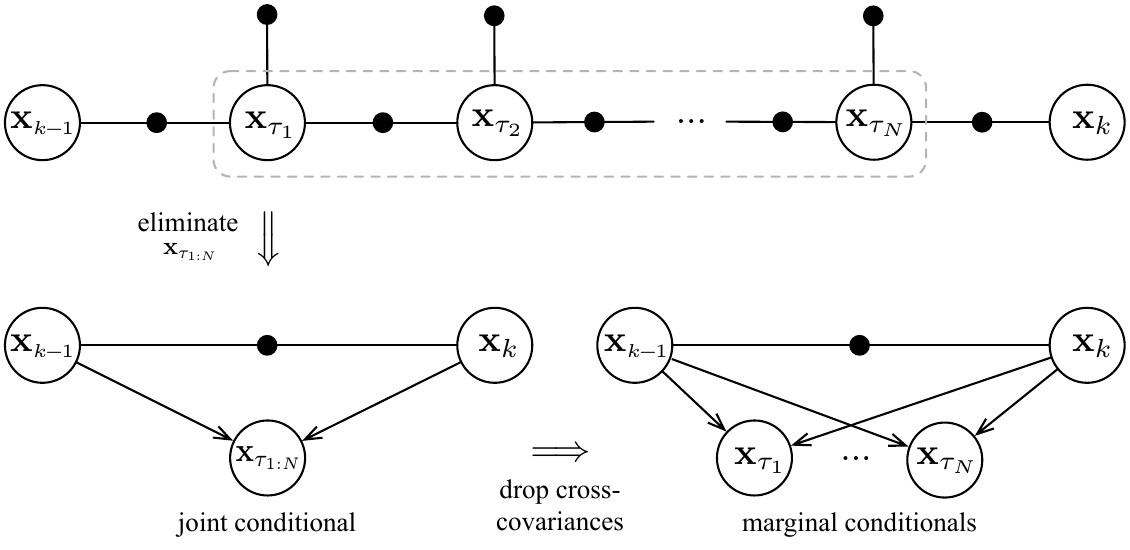}
     \caption{To reduce the number of states in the factor graph during the main solve, we can eliminate several states at once, even if they have measurements associated with them.  An example of this is classic IMU preintegration.}
     \label{fig:gausselim3}
\end{figure}

The core idea of this chapter is shown in Figure~\ref{fig:gausselim3}: we wish to interpolate a group of states, $\mbf{x}_{\tau_1}, \ldots, \mbf{x}_{\tau_N}$, to reduce the size of the estimation problem to be solved. In the context of high-rate or asynchronous sensor applications, these states are introduced to register the exact measurement time of the data.

In the previous chapter, we saw that interpolation of a continuous-time variables is equivalent to including the variable in the factor graph and subsequently eliminating it. By extension, eliminating a group of states en masse leads to the factor graph in the bottom left of the figure. The measurements associated with the interpolated variables are collected into a single factor connecting the bordering states, $\mbf{x}_{k-1}$ and $\mbf{x}_{k}$. The resulting conditional density is expressed over the joint variable $\mbf{x}_{\tau_{1:N}}$, meaning it includes the full covariance between all the eliminated states. If the cross-covariances are not required, we can drop these and only compute individual marginal conditional densities as in the bottom right of the figure. 

Grouped-state Gaussian elimination is handled by the general formula in~\eqref{eq:gausselim}.  If we do not want the full covariance between the eliminated states, we can compute only the desired diagonal blocks of $\mbs{\Sigma}_{xx} - \mbs{\Sigma}_{xs} \mbs{\Sigma}_{ss}^{-1} \mbs{\Sigma}_{sx}$, resulting in the marginal conditional densities for each of the individual eliminated states.

In our context, we may not even care about solving the problem at the interpolated states since they were introduced only to account for certain measurements. In this case, we can just drop the conditionals from the factor graph entirely.\footnote{We may not want to do this in nonlinear factor graphs, where the interpolated states can still serve as linearization points for the remaining factors.} We are left with a single, reduced factor between the states of interest, $\mbf{x}_{k-1}$ and $\mbf{x}_{k}$. The remaining sections will show how this factor can be further decomposed.

% There are a few use cases we can imagine for this type of grouped-state elimination:
% \begin{itemize}
% \item For a linear system, if we can find several places in the trajectory to carry out group-state eliminate, these calculations could potentially be done in parallel and then the main solve would also be much faster as the remaining factor graph would be much smaller.
% \item If we actually have a nonlinear system that has been linearized, we may still want to group-state eliminate several states at once (e.g., classic IMU preintegration), and then we will need the marginal conditionals to update the operating points of the linearizations for the next iteration.  The main solve might carry out several iterations with the grouped elimination only updating its operating point at a lower frequency.
% \end{itemize}

\section{Decomposing the Residual Factor}

An important remaining question is whether there is additional structure to be exploited in the residual factor between the bordering states, $\mbf{x}_{k-1}$ and $\mbf{x}_{k}$, after the grouped-state elimination. It turns out that this factor can actually be split into two factors; one corresponding to our prior model and the other corresponding to the measurements on the eliminated states. 
To see why this is true, consider that the probability density function of the original factor graph in Figure~\ref{fig:gausselim3} can be factored into a term that contains only measurement factors and a term that contains only motion-model prior factors,
\begin{equation}
     p(\mbf{x}_k, \mbf{x}_{\tau_{1:N}} | \mbf{x}_{k-1}, \mbf{y}, \mbf{v}) \propto p(\mbf{y}| \mbf{x}_{\tau_{1:N}}) p(\mbf{x}_k, \mbf{x}_{\tau_{1:N}}| \mbf{x}_{k-1}, \mbf{v}).
\end{equation}
In the original factor graph, the measurement and motion model terms are further factored as follows:
\beqn{}
     p(\mbf{y}| \mbf{x}_{\tau_{1:N}}) & = & \prod\limits_{n=1}^{N} p(\mbf{y}_n| \mbf{x}_{\tau_{n}}), \\
     p(\mbf{x}_k, \mbf{x}_{\tau_{1:N}} | \mbf{x}_{k-1}, \mbf{v}) & = & p(\mbf{x}_k|\mbf{x}_{\tau_N}, \mbf{v}_{k,\tau_N})\prod\limits_{n=2}^{N} p(\mbf{x}_{\tau_{n}}| \mbf{x}_{\tau_{n-1}}, \mbf{v}_{\tau_n, \tau_{n-1}}) \nonumber \\ &   & \qquad \qquad \times \; p(\mbf{x}_{\tau_1}|\mbf{x}_{k-1}, \mbf{v}_{\tau_1, k-1}).
\eeqn
However, given a motion prior of the type discussed in this article, it turns out that the second term can be alternatively factored as
\begin{equation}
     p(\mbf{x}_k, \mbf{x}_{\tau_{1:N}} | \mbf{x}_{k-1}, \mbf{v}) = p(\mbf{x}_{\tau_{1:N}}|\mbf{x}_k, \mbf{x}_{k-1}, \mbf{v}) p(\mbf{x}_k|\mbf{x}_{k-1}, \mbf{v}_{k,k-1}), 
\end{equation} 
where
\begin{equation}
     p(\mbf{x}_k|\mbf{x}_{k-1}, \mbf{v}_{k,k-1}) = \mathcal{N}\left(\mbf{A}_{k,k-1}\mbf{x}_{k-1} + \mbf{v}_{k,k-1}, \mbf{Q}_{k,k-1} \right),
\end{equation}
and, as above, the conditional density, $p(\mbf{x}_{\tau_{1:N}}|\mbf{x}_k, \mbf{x}_{k-1}, \mbf{v})$, can be found by applying the general formula in~\eqref{eq:gausselim}. 
% Note that we have also introduced a unary prior term, $p(\mbf{x}_1)$.

\begin{figure}[t]
     \centering
     \includegraphics[width=0.90\textwidth]{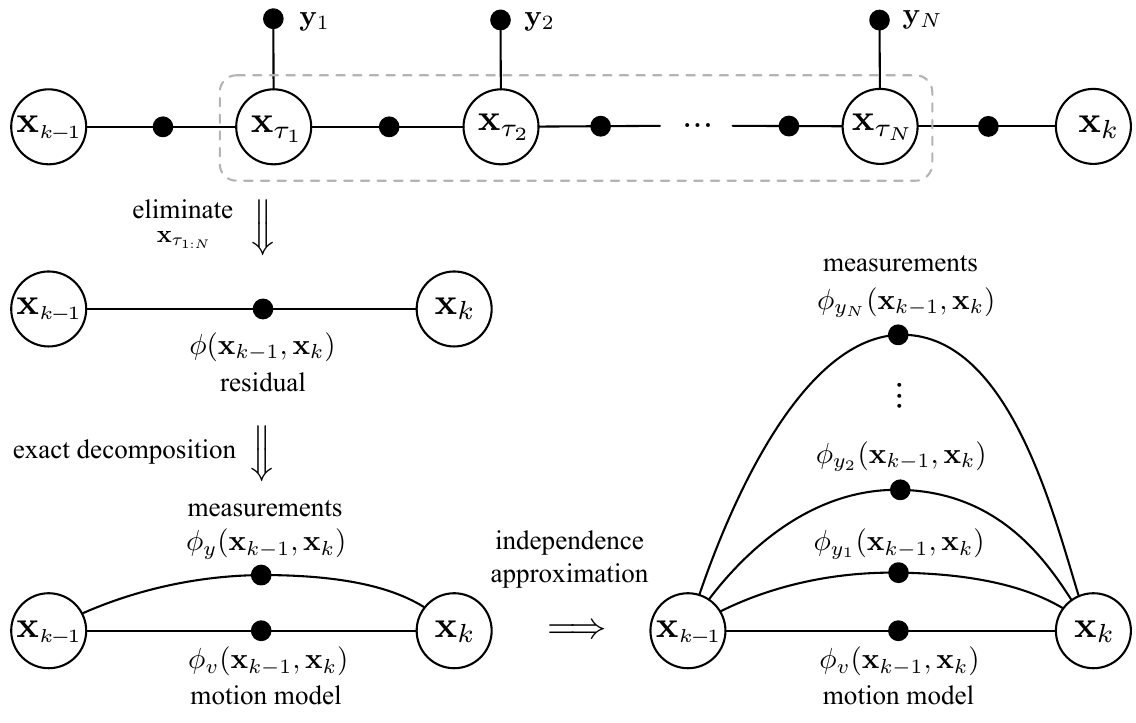}
     \caption{Following on to Figure~\ref{fig:gausselim3}, we can exactly decompose the residual factor into two factors: a measurement factor $\phi_y(\mbf{x}_{k-1}, \mbf{x}_{k})$ and a motion-prior factor $\phi_v(\mbf{x}_{k-1}, \mbf{x}_{k})$ (bottom left).  By making a further approximation, we can drop the cross-covariances between the interpolated states, resulting in a factor graph that can be solved as if the measurements were independent (bottom right).}
     \label{fig:gausselim3b}
\end{figure}

The residual factor, $\phi(\mbf{x}_{k-1}, \mbf{x}_{k})$, is obtained by marginalizing the full joint distribution with respect to the interpolated variables:
\begin{equation}
     \phi(\mbf{x}_{k-1}, \mbf{x}_{k}) = \int p(\mbf{y}| \mbf{x}_{\tau_{1:N}}) p(\mbf{x}_{\tau_{1:N}}|\mbf{x}_k, \mbf{x}_{k-1}, \mbf{v}) p(\mbf{x}_k|\mbf{x}_{k-1}, \mbf{v}_{k,k-1}) \, d\mbf{x}_{\tau_{1:N}} .
\end{equation}
The rightmost term, $p(\mbf{x}_k|\mbf{x}_{k-1}, \mbf{v}_{k,k-1})$, can be pulled out of the integral and corresponds to a motion-prior factor between bordering states, $\mbf{x}_k$ and $\mbf{x}_{k-1}$. The remaining terms in the integral correspond to the measurement distribution conditioned on the border states.  Thus, our original factor in Figure~\ref{fig:gausselim3} can be broken into two parts:
% \begin{equation}
%      p(\mbf{y}| \mbf{x}_k,\mbf{x}_{k-1}, \mbf{v}) = \int p(\mbf{y}| \mbf{x}_{\tau_{1:N}}) p(\mbf{x}_{\tau_{1:N}}|\mbf{x}_k, \mbf{x}_{k-1}) d\mbf{x}_{\tau_{1:N}}.
% \end{equation} 
\begin{equation}
     \phi(\mbf{x}_{k-1}, \mbf{x}_{k}) = \underbrace{\int p(\mbf{y}| \mbf{x}_{\tau_{1:N}}) p(\mbf{x}_{\tau_{1:N}}|\mbf{x}_k, \mbf{x}_{k-1}, \mbf{v})  \, d\mbf{x}_{\tau_{1:N}}}_{\phi_y(\mbf{x}_{k-1}, \mbf{x}_{k})} \, \underbrace{p(\mbf{x}_k|\mbf{x}_{k-1}, \mbf{v}_{k,k-1})}_{\phi_v(\mbf{x}_{k-1}, \mbf{x}_{k})}.
\end{equation}
The resulting factor graph from this factorization is shown in Figure~\ref{fig:gausselim3b} in the bottom left.

At this point, it is worth noting that, even though the original measurement distribution, $p(\mbf{y}| \mbf{x}_{\tau_{1:N}}) $, can be factorized into individual measurements, our new measurement distribution cannot be similarly factorized. This is because the interpolated states, $\mbf{x}_{\tau_{1:N}}$, are fully correlated variables.\footnote{This correlation arises due to the fact that these variables are assumed to come from the same trajectory in time. Their distributions are linked by their dependence on the bordering states.} When we eliminate these states in the factor graph, we necessarily obtain a new (large) lumped measurement factor, $\phi_y(\mbf{x}_{k-1}, \mbf{x}_{k})$, that cannot be further separated without making additional approximations.

\section{Independence Approximation}
\label{sec:ind-approx}

To keep the batch optimization lightweight, we may want to treat each measurement as a separate factor, even after marginalization of the interpolated states. This can be accomplished by dropping the cross-covariances between the interpolated variables. In other words, we can make the approximating assumption that the distribution of the $\mbf{x}_{\tau_{i}}$ variables are independent when conditioned on the bordering states:
\begin{equation}
     p(\mbf{x}_{\tau_{1:N}}|\mbf{x}_k, \mbf{x}_{k-1}, \mbf{v}) = \prod\limits_{n=1}^{N} p(\mbf{x}_{\tau_{n}}|\mbf{x}_k, \mbf{x}_{k-1}, \mbf{v}).
\end{equation}
The exact form of the individual distributions was derived above in Section~\ref{sec:linafterquery} and is given by
\begin{equation}
     p(\mbf{x}_{\tau_{n}}|\mbf{x}_k, \mbf{x}_{k-1}, \mbf{v}) = \mathcal{N}\biggl(\underbrace{\mbs{\eta}_{\tau_n}+ \bbm \mbs{\Lambda}_{\tau_n} & \mbs{\Psi}_{\tau_n} \ebm \bbm \mbf{x}_{k-1} \\ \mbf{x}_{k} \ebm}_{\check{\mbf{x}}_{\tau_n}}, \mbf{\Sigma}_{\tau_n}\biggr),
\end{equation} 
In turn, our lumped measurement factor can now be broken into factors corresponding to individual measurements:
\beqn{}
     \phi_y(\mbf{x}_{k-1}, \mbf{x}_{k}) & = & \prod_{n=1}^N \phi_{y_n}(\mbf{x}_{k-1}, \mbf{x}_{k}),\\
     \phi_{y_n}(\mbf{x}_{k-1}, \mbf{x}_{k}) & = & \int p(\mbf{y}_n| \mbf{x}_{\tau_{n}}) p(\mbf{x}_{\tau_{n}}|\mbf{x}_k, \mbf{x}_{k-1}, \mbf{v}) d\mbf{x}_{\tau_n}.\label{eqn:interp-factor}
\eeqn
Assuming that the measurements are linear, $p(\mbf{y}_n| \mbf{x}_{\tau_n})=\mathcal{N}\left(\mbf{C}_n \mbf{x}_{\tau_n}, \mbf{R}_n\right)$, it is fairly straightforward to show that these measurement factors have the following distribution:
\begin{equation}
	 \label{eq:wrapped_meas}
     \phi_{y_n}(\mbf{x}_{k-1}, \mbf{x}_{k}) = \mathcal{N}\left(\mbf{C}_n \check{\mbf{x}}_{\tau_n}, \mbf{C}_n \mbf{\Sigma}_{\tau_n}\mbf{C}_n^T + \mbf{R}_n \right).
\end{equation}
The interpretation of this distribution is straightforward: the mean is obtained by applying the measurement model to the bordering states after mapping through the motion model to the measurement time. The covariance of the measurement is inflated proportionate to the temporal distance from the bordering states to measurement times, accounting for the uncertainty of the model. 

Dropping the cross-covariance allows us to re-separate the measurements factors even after marginalization. This approximation also treats a set of correlated measurements as if they were independent. In practice, this typically does not affect the mean estimate, but can lead to overconfident estimates (i.e., artificially reduces the estimated posterior covariance).

Many prior works in continuous-time estimation -- including the \ac{STEAM} framework -- further approximate the measurement distribution by removing the $\mbf{C}_i \mbf{\Sigma}_{\tau_i}\mbf{C}_i^T$ term from the covariance~\citep{burnett_tro25}. This has the effect of removing any dependence of measurement certainty on the time that measurement was observed.

Our full procedure in this chapter is summarized in the factor graphs of Figure~\ref{fig:gausselim3b}: the final factor graph no longer requires separate states at each measurement time. Moreover, the new factors correctly account for the measurement times of the data, providing a principled way to deal with high-rate or asynchronous sensor data without needing to define a state for each measurement time. 

Although we considered unary prior factors on the interpolated states above, the derivation shown here can be easily extended to factors that involve other variables. For instance, a binary factor between a landmark variable and the interpolated variable would lead to a ternary factor between the bordering variables and the landmark variable.

%!TEX root =  main.tex

\chapter{Code API}

The \ac{API} for the code implemented as part of this project is available in the repository associated with this document: \url{https://github.com/utiasASRL/2025-fnt-ctfg/tree/main/latex/api}

\printbibliography

@book{lynch17,
	author = {Lynch, Kevin M and Park, Frank C},
	publisher = {Cambridge University Press},
	title = {Modern robotics},
	year = {2017}}

@book{boumal23,
	author = {Boumal, Nicolas},
	publisher = {Cambridge University Press},
	title = {An introduction to optimization on smooth manifolds},
	year = {2023}}

@article{dellaert17,
	author = {Dellaert, Frank and Kaess, Michael},
	journal = {Foundations and Trends{\textregistered} in Robotics},
	number = {1-2},
	pages = {1--139},
	publisher = {Now Publishers, Inc.},
	title = {Factor graphs for robot perception},
	volume = {6},
	year = {2017}}

@article{blanco10,
	author = {Blanco, Jose-Luis},
	journal = {University of Malaga, Tech. Rep},
	number = {6},
	pages = {1},
	publisher = {Citeseer},
	title = {A tutorial on se (3) transformation parameterizations and on-manifold optimization},
	volume = {3},
	year = {2010}}

@article{lilge_ijrr22,
	author = {S Lilge and T D Barfoot and J Burgner-Kahrs},
	journal = {International Journal of Robotics Research (IJRR)},
	title = {Continuum Robot State Estimation Using Gaussian Process Regression on SE(3)},
	year = {2022}}

@unpublished{sola18,
	author = {Sol{\`a} Ortega, Joan and Deray, J{\'e}r{\'e}mie and Atchuthan, Dinseh},
	title = {A micro Lie theory for state estimation in robotics},
	year = {2018}}

@article{bonnabel17,
	author = {Barrau, Axel and Bonnabel, Silv{\`e}re},
	journal = {IEEE Transactions on Automatic Control},
	number = {4},
	pages = {1797-1812},
	title = {The Invariant Extended Kalman Filter as a Stable Observer},
	volume = {62},
	year = {2017}}

@inproceedings{forster15,
	author = {Forster, Christian and Carlone, Luca and Dellaert, Frank and Scaramuzza, Davide},
	booktitle = {Proceedings of {Robotics: Science and Systems}},
	organization = {Georgia Institute of Technology},
	pages = {6-15},
	title = {IMU preintegration on manifold for efficient visual-inertial maximum-a-posteriori estimation},
	year = {2015}}

@article{wong_ral20,
	author = {J N Wong and D J Yoon and A P Schoellig and T D Barfoot},
	doi = {10.1109/LRA.2020.2969153},
	journal = {IEEE Robotics and Automation Letters (RAL)},
	month = {April},
	note = {presented at ICRA 2020.},
	number = {2},
	pages = {1429-1436},
	title = {A Data-Driven Motion Prior for Continuous-Time Trajectory Estimation on SE(3)},
	url = {sbib/wong_ral20.pdf},
	volume = {5},
	year = {2020}}

@article{tang_ral19,
	author = {T Y Tang and D J Yoon and T D Barfoot},
	doi = {10.1109/LRA.2019.2891492},
	journal = {IEEE Robotics and Automation Letters},
	note = {(\href{http://arxiv.org/abs/1809.06518}{arXiv:1809.06518 [cs.RO]}), presented at ICRA 2019.},
	number = {2},
	pages = {594-601},
	title = {A White-Noise-On-Jerk Motion Prior for Continuous-Time Trajectory Estimation on SE(3)},
	url = {sbib/tang_ral19.pdf},
	volume = {4},
	year = {2019}}

@misc{anderson_phd16,
	author = {S W Anderson},
	howpublished = {Ph.D. Thesis, University of Toronto},
	title = {Batch Continuous-Time Trajectory Estimation},
	url = {bib/anderson_phd16.pdf},
	year = {2016}}

@article{bonnabel08,
	author = {Bonnabel, Silvere and Martin, Philippe and Rouchon, Pierre},
	journal = {IEEE Transactions on Automatic Control},
	number = {11},
	pages = {2514--2526},
	publisher = {IEEE},
	title = {Symmetry-preserving observers},
	volume = {53},
	year = {2008}}

@article{mahony21,
	author = {Mahony, Robert and Trumpf, Jochen},
	journal = {IFAC-PapersOnLine},
	number = {9},
	pages = {253--260},
	publisher = {Elsevier},
	title = {Equivariant filter design for kinematic systems on lie groups},
	volume = {54},
	year = {2021}}

@book{chirikjian01,
	author = {G S Chirikjian and A B Kyatkin},
	publisher = {CRC Press},
	title = {Engineering Applications of Noncommutative Harmonic Analysis: With Emphasis on Rotation and Motion Groups},
	year = {2001}}

@book{chirikjian16,
	author = {G S Chirikjian and A B Kyatkin},
	publisher = {Dover Publications},
	title = {Harmonic Analysis for Engineers and Applied Scientists: Updated and Expanded Edition},
	year = {2016}}

@book{absil09,
	author = {P A Absil and R Mahony and R Sepulchre},
	publisher = {Princeton University Press},
	title = {Optimization on Matrix Manifolds},
	year = {2009}}

@inproceedings{anderson_iros15,
	address = {Hamburg, Germany},
	author = {S Anderson and T D Barfoot},
	booktitle = {Proceedings of the IEEE/RSJ International Conference on Intelligent Robots and Systems (IROS)},
	month = {28 September - 2 October},
	title = {Full STEAM Ahead: Exactly Sparse Gaussian Process Regression for Batch Continuous-Time Trajectory Estimation on SE(3)},
	year = {2015}}

@article{kaess08,
	author = {M Kaess and A Ranganathan and R Dellaert},
	journal = {IEEE TRO},
	number = {6},
	pages = {1365-1378},
	title = {{iSAM}: Incremental Smoothing and Mapping},
	volume = {24},
	year = {2008}}

@article{barfoot_tro14,
	author = {Timothy D Barfoot and Paul T Furgale},
	doi = {10.1109/TRO.2014.2298059},
	journal = {IEEE Transactions on Robotics},
	number = {3},
	pages = {679-693},
	title = {Associating Uncertainty with Three-Dimensional Poses for use in Estimation Problems},
	volume = {30},
	year = {2014}}

@article{tong_ijrr13b,
	author = {C H Tong and P T Furgale and T D Barfoot},
	doi = {10.1177/0278364913478672},
	journal = {International Journal of Robotics Research},
	number = {5},
	pages = {507-525},
	title = {Gaussian Process Gauss-Newton for Non-Parametric Simultaneous Localization and Mapping},
	volume = {32},
	year = {2013}}

@article{anderson_ar15,
	author = {S Anderson and T D Barfoot and C H Tong and S S\"{a}rkk\"{a}},
	doi = {10.1007/s10514-015-9455-y},
	journal = {Autonomous Robots, {\em special issue on ``Robotics Science and Systems''}},
	number = {3},
	pages = {221-238},
	title = {Batch Nonlinear Continuous-Time Trajectory Estimation as Exactly Sparse Gaussian Process Regression},
	url = {sbib/anderson_ar15.pdf},
	volume = {39},
	year = {2015}}

@inproceedings{barfoot_rss14,
	address = {Berkeley, USA},
	author = {T D Barfoot and C H Tong and S S\"{a}rkk\"{a}},
	booktitle = {Proceedings of Robotics: Science and Systems (RSS)},
	month = {12-16 July},
	title = {Batch Continuous-Time Trajectory Estimation as Exactly Sparse Gaussian Process Regression},
	year = {2014}}

@phdthesis{sarkka06,
	author = {S S\"{a}rkk\"{a}},
	school = {Helsinki University of Technology},
	title = {Recursive Bayesian Inference on Stochastic Differential Equations},
	year = {2006}}

@book{chirikjian09,
	address = {New York},
	author = {G S Chirikjian},
	publisher = {Birkhauser},
	title = {Stochastic Models, Information Theory, and {Lie} Groups: Classical Results and Geometric Methods},
	volume = {1-2},
	year = {2009}}

@inproceedings{bullo95,
	author = {F Bullo and R M Murray},
	booktitle = {Proceedings of the European Control Conference},
	pages = {1091--1097},
	title = {Proportional Derivative Control On The {Euclidean} Group},
	year = {1995}}

@techreport{deleuterio85,
	author = {G M T D'Eleuterio},
	institution = {Dynacon Enterprises Ltd.},
	month = {June},
	number = {SS-3},
	title = {Multibody Dynamics for Space Station Manipulators: Recursive Dynamics of Topological Chains},
	year = {1985}}

@inproceedings{long12,
	author = {A W Long and K C Wolfe and M J Mashner and G S Chirikjian},
	booktitle = {Proceedings of {Robotics: Science and Systems}},
	title = {The Banana Distribution is {Gaussian}: A Localization Study with Exponential Coordinates},
	year = {2012}}

@article{park95,
	author = {F C Park and J E Bobrow and S R Ploen},
	journal = {International Journal of Robotics Research},
	pages = {609-618},
	title = {A {Lie} Group Formulation of Robot Dynamics},
	volume = {14},
	year = {1995}}

@book{sastry99,
	address = {New York},
	author = {S Sastry},
	publisher = {Springer},
	title = {Nonlinear Systems: Analysis, Stability, and Control},
	year = {1999}}

@book{stillwell08,
	author = {J Stillwell},
	publisher = {Springer},
	title = {Naive {Lie} Theory},
	year = {2008}}

@article{wang06,
	author = {Y Wang and G S Chirikjian},
	journal = {{IEEE} Transactions on Robotics},
	number = {4},
	pages = {591-602},
	title = {Error Propagation on the {Euclidean} Group With Applications to Manipulator Kinematics},
	volume = {22},
	year = {2006}}

@article{wang08,
	author = {Y Wang and G S Chirikjian},
	journal = {International Journal of Robotics Research},
	number = {11},
	pages = {1258-1273},
	title = {Nonparametric Second-order Theory of Error Propagation on Motion Groups},
	volume = {27},
	year = {2008}}

@article{wolfe11,
	author = {K Wolfe and M Mashner and G Chirikjian},
	journal = {Journal of Algebraic Statistics},
	number = {1},
	pages = {75-97},
	title = {{Bayesian} Fusion on {Lie} Groups},
	volume = {2},
	year = {2011}}

@book{murray94,
	author = {R M Murray and Z Li and S Sastry},
	publisher = {CRC Press},
	title = {A Mathematical Introduction to Robotic Manipulation},
	year = {1994}}

@book{rasmussen06,
	address = {Cambridge, MA},
	author = {C E Rasmussen and C K I Williams},
	publisher = {MIT Press},
	title = {{G}aussian Processes for Machine Learning},
	year = {2006}}

@unpublished{nguyen_tro25,
	author = {T M Nguyen and Z Cao and K Li and W Talbot and T Jin and S Yuan and T D Barfoot and L Xie},
	note = {Submitted to the {IEEE Transactions on Robotics} on June 11, 2025. Manuscript \# 25-0851.},
	title = {A Third-Order Gaussian Process Trajectory Representation Framework with Closed-Form Kinematics for Continuous-Time Motion Estimation},
	year = {2025}}

@article{talbot_tro25,
	author = {W Talbot and J Nubert and T Tuna and C {Cadena Lerma} and F D\"uembgen and J {Tordesillas Torres} and T D Barfoot and M Hutter},
	journal = {IEEE Transactions on Robotics (T-RO), to appear},
	note = {(\href{https://arxiv.org/abs/2411.03951}{arXiv:2411.03951 [cs.RO]})},
	title = {Continuous-Time State Estimation Methods in Robotics: A Survey},
	year = {2025}}

@article{lilge_rob25,
	author = {S Lilge and T D Barfoot},
	journal = {Robotica, to appear},
	note = {(\href{https://arxiv.org/abs/2408.01333}{arXiv:2408.01333 [cs.RO]})},
	title = {Incorporating Control Inputs in Continuous-Time Gaussian Process State Estimation for Robotics},
	year = {2025}}

@inproceedings{teetaert_icra24,
	author = {S Teetaert and S Lilge and J Burgner-Kahrs and T D Barfoot},
	booktitle = {Proceedings of the {\em 40th Anniversary of the IEEE International Conference on Robotics and Automation (ICRA40)}},
	note = {(\href{https://arxiv.org/abs/2409.12302}{arXiv:2409.12302 [cs.RO]})},
	title = {Space-Time Continuum: Continuous Shape and Time State Estimation for Flexible Robots},
	year = {2024}}

@article{burnett_tro25,
	author = {K Burnett and A P Schoellig and T D Barfoot},
	doi = {10.1109/TRO.2024.3521856},
	journal = {IEEE Transactions on Robotics (T-RO)},
	note = {(\href{https://arxiv.org/abs/2402.06174}{arXiv:2402.06174 [cs.RO]}), presented at ICRA 2025.},
	pages = {1059-1076},
	title = {Continuous-Time Radar-Inertial and Lidar-Inertial Odometry using a Gaussian Process Motion Prior},
	url = {sbib/burnett_tro25.pdf},
	volume = {41},
	year = {2025}}

@unpublished{johnson_arxiv24,
	author = {J Johnson and J Mangelson and T D Barfoot and R Beard},
	note = {(\href{https://arxiv.org/abs/2402.00399}{arXiv:2402.00399 [cs.RO]})},
	title = {Continuous-time Trajectory Estimation: A Comparative Study Between Gaussian Process and Spline-based Approaches},
	year = {2024}}

@article{barfoot_ijrr25,
	author = {T D Barfoot and C Holmes and F D\"uembgen},
	doi = {10.1177/02783649241269337},
	journal = {International Journal of Robotics Research (IJRR)},
	note = {(\href{https://arxiv.org/abs/2308.12418}{arXiv:2308.12418 [cs.RO]})},
	number = {3},
	pages = {366-387},
	title = {Certifiably Optimal Rotation and Pose Estimation Based on the Cayley Map},
	url = {sbib/barfoot_ijrr25.pdf},
	volume = {44},
	year = {2025}}

@article{zhang_tro24,
	author = {H Zhang and C Chen and H Vallery and T D Barfoot},
	doi = {10.1109/TRO.2024.3443699},
	journal = {IEEE Transactions on Robotics (T-RO)},
	note = {(\href{https://arxiv.org/abs/2309.11134}{arXiv:2309.11134 [cs.RO]}), presented at ICRA 2025.},
	pages = {4003-4023},
	title = {GNSS/Multi-Sensor Fusion Using Continuous-Time Factor Graph Optimization for Robust Localization},
	url = {sbib/zhang_tro24.pdf},
	volume = {40},
	year = {2024}}

@article{duembgen_ral23,
	author = {F D\"uembgen and C Holmes and T D Barfoot},
	doi = {10.1109/LRA.2022.3233232},
	journal = {IEEE Robotics and Automation Letters (RAL)},
	note = {(\href{https://arxiv.org/abs/2209.04266}{arXiv:2209.04266 [cs.RO]}), presented at IROS 2023.},
	number = {2},
	pages = {1117-1124},
	title = {Safe and Smooth: Certified Continuous-Time Range-Only Localization},
	url = {sbib/duembgen_ral23.pdf},
	volume = {8},
	year = {2023}}

@article{lilge_tro25,
	author = {S Lilge and T D Barfoot and J Burgner-Kahrs},
	doi = {10.1109/TRO.2024.3521859},
	journal = {IEEE Transactions on Robotics (T-RO)},
	note = {(\href{https://arxiv.org/abs/2401.13540}{arXiv:2401.13540 [cs.RO]}), presented at ICRA 2025.},
	number = {905-925},
	title = {State Estimation for Continuum Multi-Robot Systems on SE(3)},
	url = {sbib/lilge_tro25.pdf},
	volume = {41},
	year = {2025}}

@book{barfoot_ser24,
	author = {T D Barfoot},
	doi = {10.1017/9781009299909},
	edition = {2nd},
	isbn = {9781009299893},
	publisher = {Cambridge University Press},
	title = {State Estimation for Robotics},
	url = {bib/barfoot_ser24.pdf},
	year = {2024}}

@inproceedings{anderson_icra14,
	address = {Hong Kong, China},
	author = {S Anderson and F Dellaert and T D Barfoot},
	booktitle = {Proceedings of the IEEE International Conference on Robotics and Automation (ICRA)},
	doi = {10.1109/ICRA.2014.6906884},
	month = {31 May - 7 June},
	pages = {373-380},
	title = {A Hierarchical Wavelet Decomposition for Continuous-Time SLAM},
	url = {sbib/anderson_icra14.pdf},
	year = {2014}}

@inproceedings{tong_crv12,
	address = {Toronto, Canada},
	author = {Chi Hay Tong and Paul T Furgale and Timothy D Barfoot},
	booktitle = {Proceedings of the 9th Conference on Computer and Robot Vision (CRV)},
	doi = {10.1109/CRV.2012.35},
	month = {28-30 May},
	pages = {206-213},
	title = {Gaussian Process Gauss-Newton: Non-Parametric State Estimation},
	url = {sbib/tong_crv12.pdf},
	year = {2012}}

@article{dellaertFactorGraphsRobot2017,
  title = {Factor {{Graphs}} for {{Robot Perception}}},
  author = {Dellaert, Frank and Kaess, Michael},
  year = {2017},
  month = aug,
  journal = {Foundations and Trends{\textregistered} in Robotics},
  volume = {6},
  number = {1-2},
  pages = {1--139},
  publisher = {Now Publishers, Inc.},
  issn = {1935-8253, 1935-8261},
  doi = {10.1561/2300000043},
  urldate = {2025-07-09},
  langid = {english}
}

@inproceedings{dongSparseGaussianProcesses2018,
  title = {Sparse {{Gaussian Processes}} on {{Matrix Lie Groups}}: {{A Unified Framework}} for {{Optimizing Continuous-Time Trajectories}}},
  shorttitle = {Sparse {{Gaussian Processes}} on {{Matrix Lie Groups}}},
  booktitle = {2018 {{IEEE International Conference}} on {{Robotics}} and {{Automation}} ({{ICRA}})},
  author = {Dong, Jing and Mukadam, Mustafa and Boots, Byron and Dellaert, Frank},
  year = {2018},
  month = may,
  pages = {6497--6504},
  issn = {2577-087X},
  doi = {10.1109/ICRA.2018.8461077},
  urldate = {2025-07-25}
}

@article{johnsonContinuousTimeTrajectoryEstimation2023,
  title = {Continuous-{{Time Trajectory Estimation}} for {{Differentially Flat Systems}}},
  author = {Johnson, Jacob C. and Mangelson, Joshua G. and Beard, Randal W.},
  year = {2023},
  month = jan,
  journal = {IEEE Robotics and Automation Letters},
  volume = {8},
  number = {1},
  pages = {145--151},
  issn = {2377-3766},
  doi = {10.1109/LRA.2022.3224364},
  urldate = {2025-07-27}
}

@article{kaessISAM2IncrementalSmoothing2012,
  title = {{{iSAM2}}: {{Incremental}} Smoothing and Mapping Using the {{Bayes}} Tree},
  shorttitle = {{{iSAM2}}},
  author = {Kaess, Michael and Johannsson, Hordur and Roberts, Richard and Ila, Viorela and Leonard, John J and Dellaert, Frank},
  year = {2012},
  month = feb,
  journal = {The International Journal of Robotics Research},
  volume = {31},
  number = {2},
  pages = {216--235},
  publisher = {SAGE Publications Ltd STM},
  issn = {0278-3649},
  doi = {10.1177/0278364911430419},
  urldate = {2025-07-25},
  langid = {english}
}

@article{legentilContinuousLatentState2023,
  title = {Continuous Latent State Preintegration for Inertial-Aided Systems},
  author = {Le Gentil, Cedric and {Vidal-Calleja}, Teresa},
  year = {2023},
  month = sep,
  journal = {The International Journal of Robotics Research},
  volume = {42},
  number = {10},
  pages = {874--900},
  publisher = {SAGE Publications Ltd STM},
  issn = {0278-3649},
  doi = {10.1177/02783649231199537},
  urldate = {2025-07-25},
  langid = {english}
}

@article{legentilGaussianProcessPreintegration2020,
  title = {Gaussian {{Process Preintegration}} for {{Inertial-Aided State Estimation}}},
  author = {Le Gentil, Cedric and {Vidal-Calleja}, Teresa and Huang, Shoudong},
  year = {2020},
  month = apr,
  journal = {IEEE Robotics and Automation Letters},
  volume = {5},
  number = {2},
  pages = {2108--2114},
  issn = {2377-3766},
  doi = {10.1109/LRA.2020.2970940},
  urldate = {2025-07-25}
}

@article{mukadamContinuoustimeGaussianProcess2018,
  title = {Continuous-Time {{Gaussian}} Process Motion Planning via Probabilistic Inference},
  author = {Mukadam, Mustafa and Dong, Jing and Yan, Xinyan and Dellaert, Frank and Boots, Byron},
  year = {2018},
  month = sep,
  journal = {The International Journal of Robotics Research},
  volume = {37},
  number = {11},
  pages = {1319--1340},
  publisher = {SAGE Publications Ltd STM},
  issn = {0278-3649},
  doi = {10.1177/0278364918790369},
  urldate = {2025-07-25},
  langid = {english}
}

@misc{yanIncrementalSparseGP2015,
  title = {Incremental {{Sparse GP Regression}} for {{Continuous-time Trajectory Estimation}} \& {{Mapping}}},
  author = {Yan, Xinyan and Indelman, Vadim and Boots, Byron},
  year = {2015},
  month = apr,
  number = {arXiv:1504.02696},
  eprint = {1504.02696},
  primaryclass = {cs},
  publisher = {arXiv},
  doi = {10.48550/arXiv.1504.02696},
  urldate = {2025-07-25},
  archiveprefix = {arXiv}
}

@article{yanIncrementalSparseGP2017,
  title = {Incremental Sparse {{GP}} Regression for Continuous-Time Trajectory Estimation and Mapping},
  author = {Yan, Xinyan and Indelman, Vadim and Boots, Byron},
  year = {2017},
  month = jan,
  journal = {Robotics and Autonomous Systems},
  volume = {87},
  pages = {120--132},
  issn = {0921-8890},
  doi = {10.1016/j.robot.2016.10.004},
  urldate = {2025-07-25}
}

@book{gallierDifferentialGeometryLie2020,
	address = {Cham},
	author = {Gallier, Jean and Quaintance, Jocelyn},
	copyright = {http://www.springer.com/tdm},
	doi = {10.1007/978-3-030-46040-2},
	isbn = {978-3-030-46039-6 978-3-030-46040-2},
	langid = {english},
	publisher = {Springer International Publishing},
	series = {Geometry and {{Computing}}},
	shorttitle = {Differential {{Geometry}} and {{Lie Groups}}},
	title = {Differential {{Geometry}} and {{Lie Groups}}: {{A Computational Perspective}}},
	urldate = {2025-02-26},
	volume = {12},
	year = {2020}}

@misc{solaMicroLieTheory2021,
	archiveprefix = {arXiv},
	author = {Sol{\`a}, Joan and Deray, Jeremie and Atchuthan, Dinesh},
	doi = {10.48550/arXiv.1812.01537},
	eprint = {1812.01537},
	month = dec,
	number = {arXiv:1812.01537},
	primaryclass = {cs},
	publisher = {arXiv},
	title = {A Micro {{Lie}} Theory for State Estimation in Robotics},
	urldate = {2025-02-13},
	year = {2021}}

\end{document}